%% file: 0_main.tex
\pdfoutput=1

\documentclass[11pt]{article}

\usepackage[final]{acl}
\usepackage{times}
\usepackage{latexsym}
\usepackage[T1]{fontenc}
\usepackage[utf8]{inputenc}
\usepackage{microtype}
\usepackage{inconsolata}
\usepackage{graphicx}
\usepackage{hyperref}
\usepackage{booktabs}
\usepackage{multirow}
\usepackage{multicol}
\usepackage{graphicx}
\usepackage{tabularx}
\usepackage{amsmath}
\usepackage{microtype}
\usepackage{latexsym}
\usepackage{inconsolata}
\usepackage{amssymb}
\usepackage{url}
\usepackage{placeins}
\usepackage{natbib}
\usepackage{balance}

\title{HalluCounter: Reference-free LLM Hallucination Detection in the Wild!}

\author{Ashok Urlana\textsuperscript{1,2} \enspace \enspace Gopichand Kanumolu\textsuperscript{2} \enspace \enspace Charaka Vinayak Kumar\textsuperscript{2} \\ \enspace \enspace \textbf{Bala Mallikarjunarao Garlapati\textsuperscript{2}} \enspace \enspace \textbf{Rahul Mishra\textsuperscript{1,3}}\\
IIIT Hyderabad\textsuperscript{1} \enspace \enspace \enspace \enspace \enspace \enspace
TCS Research, Hyderabad, India\textsuperscript{2} \enspace
University of Oslo, Norway\textsuperscript{3}\\
{\tt ashok.u@research.iiit.ac.in, rahul.mishra@iiit.ac.in}, \\ \texttt{\{ashok.urlana, gopichand.kanumolu, charaka.v, balamallikarjuna.g\}}@tcs.com
}

\begin{document}
\maketitle
\begin{abstract}
 Response consistency-based, reference-free hallucination detection (RFHD) methods do not depend on internal model states, such as generation probabilities or gradients, which Grey-box models typically rely on but are inaccessible in closed-source LLMs. However, their inability to capture query-response alignment patterns often results in lower detection accuracy. Additionally, the lack of large-scale benchmark datasets spanning diverse domains remains a challenge, as most existing datasets are limited in size and scope. To this end, we propose \textbf{\textit{HalluCounter}}, a novel reference-free hallucination detection method that utilizes both response-response and query-response consistency and alignment patterns. This enables the training of a classifier that detects hallucinations and provides a confidence score and an optimal response for user queries. Furthermore, we introduce \textbf{\textit{HalluCounterEval}}, a benchmark dataset comprising both synthetically generated and human-curated samples across multiple domains. Our method outperforms state-of-the-art approaches by a significant margin, achieving over 90\% average confidence in hallucination detection across datasets\footnote{Code and data are publicly available for research purposes: \url{https://github.com/rahulOm9/HalluCounter}}.

\end{abstract}

\input{1_Introduction}
\input{3_dataset_creation}
\input{4_methodology}
\input{5_experiments_results}

\input{6_discussion}
\input{2_Related_works}

\input{7_conclusions_and_limitations}
\bibliography{custom}
\newpage

\appendix
\input{8_appendix}

\FloatBarrier
\section{Datasheet for HaluCounterEval}\label{appendix:data_sheet}
\input{datasheet}



\end{document}

%% file: 1_Introduction.tex
\section{Introduction}
Reference-free hallucination detection (RFHD) is gaining significant traction in the research community \cite{manakul-etal-2023-selfcheckgpt, zhang-etal-2023-sac3, yehuda-etal-2024-interrogatellm}, as it obviates the need for reference texts or external knowledge bases (KBs) to identify potential hallucinations. This enhances the scalability and applicability of RFHD across a broader range of tasks and scenarios, which would otherwise be constrained by reference- or KB-dependent approaches \cite{hu2024refchecker, liu2024litcab}. In the literature, RFHD approaches can be broadly categorized into two major classes. The first category, known as black-box approaches, relies on analyzing multiple responses generated by LLMs to assess consistency and alignment among them, thereby detecting hallucinations in the output \cite{manakul-etal-2023-selfcheckgpt}.

On the other hand, grey-box models leverage internal states of the models, such as decoder generation probabilities \cite{farquhar2024detecting}, final-layer gradients \cite{ji2024llm, snyder2024early}, and entropy of the generated tokens \cite{farquhar2024detecting} to identify hallucinations. While grey-box models achieve higher detection accuracy than black-box models, they are computationally more demanding and cannot be applied to closed-source models due to restricted access to internal states. Conversely, black-box models, though computationally simpler, tend to perform less effectively \cite{deutsch-etal-2022-limitations}. Additionally, we observe a significant lack of suitable and sufficiently large benchmark datasets spanning multiple domains to facilitate the evaluation and development of future RFHD methods \cite{sahoo-etal-2024-comprehensive}.

In this paper, we propose HalluCounter, a novel method that enhances response-consistency-based approaches by incorporating both response-response and query-response interactions. By leveraging consistency and alignment scores, HalluCounter learns a robust hallucination detection classifier. Response consistency-based approaches aim to detect hallucination in LLMs by generating multiple responses for the same input query and analyzing the variation in these responses \cite{manakul-etal-2023-selfcheckgpt}. Significant inconsistencies or contradictions across the generated responses signal potential hallucinations. Unlike prior methods, HalluCounter does not evaluate hallucination at the level of individual responses; rather, it assesses the self-consistency of an LLM when generating multiple responses to the same query. The core objective is to determine whether the LLM exhibits a tendency to hallucinate for a given query, rather than making a binary decision about a single response. Our model not only achieves higher detection accuracy compared to popular baselines but also provides a confidence score indicating how certain it is about its decision.
Additionally, HalluCounter suggests the optimal response for users, regardless of whether the original generation contains hallucinations. Furthermore, we introduce a large-scale, multi-domain dataset for the RFHD task, comprising both synthetic and human-annotated samples. Unlike other existing datasets, this dataset poses significantly greater challenges for RFHD methods. It includes samples that demand domain knowledge across diverse fields, ranging from factual queries to those requiring reasoning and mathematical skills, which could be a good test bench for further RFHD explorations.

The key contributions of this work are: 1) We introduce HalluCounter, a novel approach for the RFHD task. 2) We present a large-scale, multi-domain benchmark dataset for RFHD, featuring both synthetic and human-annotated samples. 3) We conduct extensive experiments exploring various feature combinations, labeling strategies, classifiers, and LLMs across different sizes and families. 4) We perform a rigorous human evaluation of the model's selected optimal responses and carry out a thorough error analysis to uncover its potential limitations.

%% file: 3_dataset_creation.tex
\section{HalluCounterEval dataset creation}
\label{sec:data_creation}
This section describes the creation of the HalluCounterEval dataset, which consists of various synthetic and human-annotated datasets for training and testing.
\subsection{Raw data collection and processing}
HalluCounterEval consists of two different training datasets. To create the first one, we obtain the raw data from an American television game show `Jeopardy' \cite{Jeopardy} and filter the dataset, which is highly diverse by including question-answer pairs related to six major domains and 22 sub-domains as detailed in Table~\ref{tab:halugurad_data}. Moreover, the dataset includes factoid-based QA pairs, where many questions are not straightforward to answer. These questions often contain indirect hints, which increase their complexity and challenge the LLM's ability to handle ambiguity. The second dataset is the combination of multiple datasets obtained from Kaggle including Scientific QA \cite{ScientificQA}, MathQA \cite{MathQA}, Math QSA \cite{MathQSA}, and General Knowledge \cite{GK} QA pairs as shown in Table~\ref{tab:kaggle_data_stats}. In the Kaggle dataset, scientific and GK questions test the LLMs' ability to extract factual knowledge. Whereas, MathQA and MathQSA questions assess the LLMs' logical reasoning and familiarization capabilities with mathematical notations. Both datasets undergo rule-based filtration steps as detailed in Appendix~\ref{sec:jeopardy_filter} to maintain the high quality. In accordance with \citet{datasheet}’s recommendation, we include a data sheet in Appendix~\ref{appendix:data_sheet}.
\subsection{Training dataset creation}
\label{sec:train_datac_create}
The creation of training datasets consists of two stages 1) generation of sample responses, 
and 2) data labeling. 
\subsubsection{Sample responses generation}
We utilize six different LLMs, including TinyLLaMA-1.1B \cite{zhang2024tinyllama}, Phi-3.5-B-mini \cite{abdin2024phi}, Mistral-7B-instruct \cite{jiang2023mistral}, LLaMA-3-instruct 8B and 70B \cite{dubey2024llama}, and Gemma-7B-instruct \cite{team2024gemma} models to generate \textit{`k'} responses\footnote{LLM generated `responses' interchangeably refereed as `sample responses'} for each query by prompting each model \textit{`k'} times. Due to limited compute, we use the 8-bit quantized version of the LLaMA-3-instruct-70B model for the inference, whereas other models are non-quantized versions. Further, as depicted in Appendix~\ref{sec:train_test} Figure~\ref{fig:unique_responses}, we notice that TinyLLaMA-1.1B has the highest number of unique responses (lowest self-consistency) followed by Mistral-7B-instruct. All the prompts and corresponding inference configurations can be found in Appendix~\ref{sec:responses_generation}.  

\input{table_pilot_study_results}

\subsubsection{Data Labeling}
\label{sec:obtain_label}
Data labeling aims to classify each LLM-generated sample response as either accurate (0) or hallucinated (1). The labeling can be achieved either through an LLM as a judge approach or a search-based string-matching method.\\ 
\textbf{(1) LLM as a judge.} Prompt an LLM by providing the question, LLM response, and gold answer to classify whether the LLM response is accurate (0) or hallucinated (1).\\
\textbf{(2) Exact-match.} A search-based string-matching approach classifies an LLM's response as non-hallucinated if it matches the gold answer; otherwise, it is labeled as hallucinated.\\
\textbf{Pilot study.} To find the appropriate approach for the data labeling, we create a human-annotated dataset of 500 samples with the help of three expert annotators. To perform the annotation, we provide the question, gold answer, and LLM-generated response and ask the annotators to classify whether the LLM-generated response is hallucinated. \\
\textbf{Selection of best labeling strategy.} 
To find out the appropriate labeling strategy, we generate the labels by prompting GPT-4o mini \cite{achiam2023gpt} (closed source), LLaMA3-70B and Qwen2.5-32B \cite{yang2024qwen2} (open source), and string-based matching methods and compare the percentage of labels match with the human-annotated dataset. As illustrated in Table~\ref{tab:pilot_study_results}, all three LLM-based labeling strategies perform similarly, with only minor variations when compared to human-annotated labels. However, we choose the Qwen2.5-32B for the entire training dataset labeling to reduce the compute requirements and encourage reproducibility by utilizing open-source models. The corresponding prompt for the labeling method is mentioned in Appendix~\ref{sec:qwen_prompt_labeling} Table~\ref{tab:qwen_labeling_prompt}.

\subsection{Test datasets creation}
\label{sec:Test dataset creation}
The HalluCounterEval dataset consists of 16 test datasets. Out of these, 14 are synthetically generated and two are human-annotated test sets. To create these test sets, we leverage both LLM and human annotation strategies. \\
\textbf{Synthetic test sets.} To create each test set, we follow the similar steps detailed for the training dataset creation (see Section~\ref{sec:train_datac_create}). We obtain the test sets corresponding to Jeopardy and Kaggle datasets for TinyLLaMA-1.1B (\textit{TL-1.1B-Gen}), Phi-3.5-B-mini (\textit{PHI-3.5B-Gen}), Mistral-7B-instruct (\textit{MST-7B-Gen}), LLaMA-3-instruct 8B (\textit{LL-7B-Gen}) and 70B (\textit{LL-70B-Gen}), Gemma-7B-instruct (\textit{GM-7B-Gen}) and `ensemble' (\textit{ENSB-Gen}) models. The `ensemble' test set consists of an equal number of samples assigned to different LLMs to generate the sample responses. In the rest of the paper, we report all the results on the test sets with corresponding acronyms of each LLM. \\
\textbf{Human-annotated test set (HA-Test)} is a manually curated dataset consisting of 1,956 samples or queries, with 956 sourced from Jeopardy and 1,000 from Kaggle datasets. For each query, we generate 10 responses, resulting in a total dataset size of 19,560 query-response pairs. Similar to the `ensemble' test set, the HA-Test consists of LLM-generated responses from various LLMs. We classify the sample responses with the help of three expert annotators. Where, we provide a question, gold answer, and LLM response to the annotator and ask them to label it as either hallucinated (1) or non-hallucinated (0). We measure the Inter Annotator Agreement (IAA) between the annotators and obtain the Fleiss\footnote{\url{https://en.wikipedia.org/wiki/Fleiss\%27\_kappa}} kappa score of 0.83, which indicates an almost perfect agreement. 

%% file: table_pilot_study_results.tex
\begin{table}[t]
\centering
\resizebox{\columnwidth}{!}{
\begin{tabular}{@{}cccc@{}}
\toprule
\textbf{String matching} &\textbf{ Qwen2.5-32B }& \textbf{Llama3-70B} & \textbf{GPT-4o} \\ \midrule
69.4\%                     & 89.4\%         & 89.6\%       & 89.8\%        \\ \bottomrule
\end{tabular}}
\caption{Proportion of samples where the classification aligns with the human-annotated dataset.}
\label{tab:pilot_study_results}
\vspace{-5mm}
\end{table}

%% file: 4_methodology.tex
\section{Methodology}
\subsection{Task formulation}
We prompt a query \textit{Q} to an LLM and collect \textit{`k'} responses, denoted as \textit{R} = ${R_1, R_2, \dots, R_k}$, by querying the model \textit{`k'} times with the same prompt. The query and its corresponding \textit{`k'} responses are then processed by the proposed HalluCounter pipeline, which performs three key tasks: 1) determines whether LLM makes the hallucination for the given query, 2) provides a confidence score for the classifier's overall prediction, and 3) identifies the least hallucinated response among the \textit{`k'} responses, referred as the optimal response.

 \begin{figure*}[t]
  \centering
  \scalebox{.87}
   { \includegraphics[clip, trim=2cm 5.4cm 5.2cm 3.5cm, width=1\linewidth]{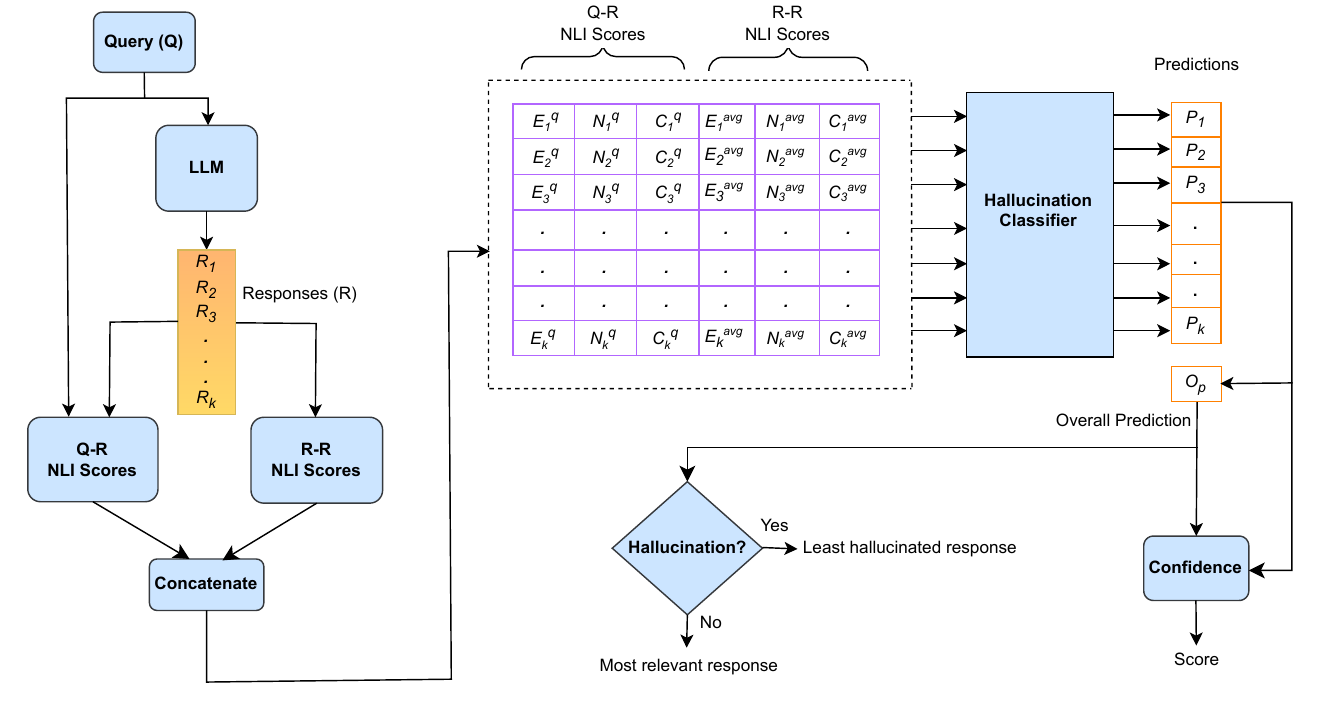}}
  \caption{\textbf{\textit{HalluCounter:}} A reference-free Hallucination Detection Pipeline for LLMs with three key components, 1) Extracting NLI features for query-response and response-response pairs, 2) A hallucination classifier that predicts hallucinations, and 3) Aggregating the final prediction, confidence score, and optimal response.}
  \label{fig:pipeline}
  \vspace{-5mm}
\end{figure*}

\subsection{HalluCounter Approach}
The HalluCounter pipeline consists of three stages: 1) Extracting the NLI features, 2) Classification of the responses, and 3) Optimal response generation, and confidence score calculation. The following is a detailed description of each stage.

\subsubsection{Extracting NLI features}
We extract the NLI features between the Query-Response (Q-R) and Response-Response (R-R) pairs using the DeBERTa-v3-large \citep{hedeberta} based cross-encoder model, fine-tuned on MNLI \cite{williams-etal-2018-broad}. We measure the NLI scores by concatenating the query with the LLM response or between the sample responses. The outputs from the NLI model are the logits associated with entailment, neutral, and contradiction.

\noindent \textbf{Query-Response NLI features.} To understand whether the generated response is relevant to the query or not, we obtain the NLI scores between the query and each response among all the \textit{`k'} responses. As shown in Figure~\ref{fig:pipeline}, the corresponding NLI scores indicated as: $\left( E_{i}^{q}, N_{i}^{q}, C_{i}^{q} \right) \quad \text{for} \quad i = 1, 2, \dots, k$. We adopted the use of Q-R NLI scores following prior research \cite{fortier-dubois-rosati-2023-using}, which highlights the effectiveness of contradiction-based reasoning in improving QA models. \\
\textbf{Response-Response NLI features.}
 To verify the consistency among the sample responses, each response in the \textit{R} is compared with other responses and obtains the corresponding NLI scores. We average the entailment, neutral, and contradiction features for each response. \text{For a response } R$_{i}$,
\begin{equation}
     \text{ Avg NLI scores} = 
    \begin{cases}
        \text{E}_{i}^{avg} = \frac{1}{k-1}\sum\limits_{i=1,j\neq i}^{k} e_{ij} \\[10pt]
        \text{N}_{i}^{avg} =  \frac{1}{k-1}\sum\limits_{i=1,j\neq i}^{k} n_{ij} \\[10pt]
        \text{C}_{i}^{avg} =  \frac{1}{k-1}\sum\limits_{i=1,j\neq i}^{k} c_{ij}
    \end{cases}
    \label{eqn:nli_mean}
\end{equation}
Where $e_{ij}$, $n_{ij}$, $c_{ij}$ are the entailment, neutral and contradiction scores between $i^{th}$ and $j^{th}$ responses.

\subsubsection{Hallucination detection classifier}
We build a classifier to classify whether the generated response contains hallucination or not. It takes the input as NLI feature values and generates binary output `1' for hallucination and `0' for non-hallucination. We built two different classifiers using statistical and BERT-based approaches.\\ 
\textbf{Statistical Method.} We utilize the ensemble of the Decision Trees, XGBoost, gradient-boosted Decision Trees (GBDT), and a voting classifier to design an ensemble classifier. \\
\textbf{BERT classifier.}
We use the bert-base-uncased \cite{devlin-etal-2019-bert} model to fine-tune the classifier by converting all the numerical features into textual features. Additional experimental details can be found in Appendix~\ref{sec:exp_setup_info}. Furthermore, our pipeline yields the following three key outcomes.

\noindent \textbf{1. Overall prediction:} Let the \( k \) predictions be denoted as \( p_1, p_2, \dots, p_k \), where each \( p_i \in \{0, 1\} \).
    We define the overall prediction \( \hat{y} \) as:
    \begin{equation}
    \hat{y} = \begin{cases} 
    1 & \text{if } \sum\limits_{i=1}^k p_i \geq \frac{k}{2} \\
    0 & \text{if } \sum\limits_{i=1}^k p_i < \frac{k}{2}
    \end{cases}
    \end{equation}

\noindent \textbf{2. Optimal response:} We select the optimal response based on the overall prediction ($\hat{y}$) of the classifier. If the overall prediction is hallucinated, we choose all sample responses categorized as hallucination and among them pick the sample with the lowest contradiction score, whereas if the over-
\begin{equation}
    \resizebox{\linewidth}{!}{$
    \text{R*} =
    \begin{cases}
    \arg\min\limits_{\underset{R_i \in R}{}} \left( \epsilon_1 \cdot \left( c_i^{\text{q}} \right) + \epsilon_2 \cdot \left( c_i^{\text{avg}} \right) \right) &  \text{$\hat{y}$ = 1} \\
    \arg\max\limits_{\underset{R_i \in R}{}} \left( \epsilon_1 \cdot \left( e_i^{\text{q}} \right) + \epsilon_2 \cdot \left(  e_i^{\text{avg}}   \right) \right) &  \text{$\hat{y}$ = 0}
    \end{cases}
    $}
    \label{eqn:optimal_response}
    \vspace{-2mm}
\end{equation}

\noindent all prediction is non-hallucinated, we select all the corresponding sample responses and among them pick the sample with the highest entailment score. This process ensures an optimal response to user queries. The optimal response \( R^* \) is selected using Equation~\ref{eqn:optimal_response}. Where \( R = [R_1, R_2, \dots, R_k] \) represents the set of responses, $\epsilon_1$ and $\epsilon_2$ values indicate the weightage given to the Q-R and R-R feature values. After experimenting with various combinations of $\epsilon_1$ and $\epsilon_2$ values, we set $\epsilon_1$ = 0.3 and $\epsilon_2$ = 0.7. \\
\textbf{3. Confidence score (CS): } The confidence score is measured using all \textit{`k'} responses predictions and overall prediction. Let's take the \textit{`k'} responses individual classifier predictions are $\{p_{1}$, $p_{2}$ \dots $p_{k}\}$ and $\hat{y}$ is the overall prediction for the given query, then the confidence score is measured using Equation~\ref{eqn:confidence_score}.
    \begin{equation}
    \text{CS} = 
    \begin{cases} 
     \frac{1}{k}\sum\limits_{i=1}^{k} p_{i} & \text{$\hat{y}$ = 1} \\ 
    1-\frac{1}{k}\sum\limits_{i=1}^{k} p_{i} & \text{$\hat{y}$ = 0}
    \end{cases}
    \label{eqn:confidence_score}
    \end{equation}

%% file: 5_experiments_results.tex
\section{Experiments and Results}
\label{sec:experiment_results}
This section presents the experimental results of the proposed pipeline and corresponding analysis. We report the F1-Score, AUC, and Balanced accuracy scores to evaluate the hallucination classifier performance. 
\subsection{NLI features combinations}
\label{sec:features_desc}
We obtain various combinations of NLI features to train different classifiers. In total, we obtain eight features for a given query, out of them 6 are numerical features (three from each query-response (Q-R) and response-response (R-R) pairs NLI scores) and two are textual features (`query' \& `LLM response'). Using these features, we built several classifiers by combining them as shown in Table~\ref{tab:features_combination}.  
\input{table_features_combinations}
\input{table_hallucounter_varying_samples}
\input{table_hallu_eval}

\input{table_pipeline_results}
\subsection{Jeopardy and Kaggle results analysis}
We conduct experiments on Jeopardy and Kaggle datasets, by training various classifiers using statistical and BERT-based models on the 16 test sets. All the combinations of the experiments conducted are listed in Table~\ref{tab:all_results_details}. As shown in Table~\ref{tab:jeopardy_results}, for the Jeopardy dataset, the BERT classifier trained on a combination of numerical and textual features (q-r+Q-R-R-R) outperforms all other models, except for the HA-Test. Whereas on HA-test the model trained using a statistical classifier with EC-EC feature combination performs better than others. Additionally, as detailed in Table~\ref{tab:jeopardy_category_wise}, we conduct experiments to evaluate the performance of the hallucination classifier across six sub-categories present in the Jeopardy dataset.


Similarly, we conduct experiments with the Kaggle test sets and listed the results in Table~\ref{tab:kaggle_results}. Given the variations, such as mathematical formulations, present in the Kaggle test sets, we notice that the classifier trained on EC-EC feature combination performs comparably or even surpasses the `q-r+Q-R-R-R' combination. Moreover, we report the results from all four datasets within the Kaggle dataset in Table~\ref{tab:kaggle_datasetwise}. Appendix~\ref{sec:halu_classifer_results} presents the hallucination classifier results for all the combinations listed in Table~\ref{tab:all_results_details} and  Appendix~\ref{sec:gpt4_testing} describes HalluCounter’s performance on responses generated by GPT-4o \cite{hurst2024gpt}.
We recommend using the ‘q-r+Q-R-R-R’ feature combination with a BERT classifier as a strong starting point when applying HalluCounter to new datasets. This combination has shown robust performance across multiple test sets, making it a reliable default choice.
\input{table_SOTA_comparison}
\input{table_error_classification}

\input{table_kaggle_results}
\subsection{Ablation study}
\textbf{Impact on the varying number of responses.} We experiment with different numbers of sample responses (k = 3, 5, 7, 10) and notice the variations in the pipeline's prediction confidence values and hallucination rates. As detailed in Table~\ref{tab:pipeline_results}, we find that as the number of sample responses increases, both the hallucination rate and the confidence of the hallucination classifier slightly decrease. However, despite changing the number of responses, our pipeline exhibits more than 90\% confidence across different test sets, which indicates that the proposed pipeline is independent of the number of responses and the best results can be obtained with three sample responses as well. Moreover, as shown in Table~\ref{tab:hallucounter_varying_k} the pipeline exhibits stable performance across different `k' values.  \\
\textbf{Performance on non-QA tasks.}
To verify the efficacy of HalluCounter on other than factoid QA datasets, we tested the HalluCounter on HaluEval \cite{li-etal-2023-halueval} dataset. Which consists of summarization, knowledge-grounded dialogue, and QA tasks. The HalluCounter performance on the HaluEval dataset are reported in Table \ref{tab:hallueval_experiments_results}.
\subsection{Comparison with state-of-the-art}
We compare our approach with two popularly known reference-free hallucination detection approaches in LLMs, which are SelfCheckGPT \cite{manakul-etal-2023-selfcheckgpt} and InterrogateLLM \cite{yehuda-etal-2024-interrogatellm}, and uncertainty-based approaches, namely Perplexity \cite{renout}, Length Normalized entropy \cite{malininuncertainty}, and Lexical similarity \cite{lin-etal-2022-towards}. Moreover, we also compared with three reference-based approaches HaloScope \cite{du2024haloscope}, SAPLMA \cite{cheninside} and Eigenscore \cite{azaria-mitchell-2023-internal}.
As detailed in Table~\ref{tab:cross_compare}, HalluCounter outperforms current state-of-the-art methods by a significant average margin of 10\% with SelfCheckGPT and 21\% with InterrogateLLM. Our study proves that consistency among only generated responses is insufficient to perform the RFHD task, the proposed approach outperforms state-of-the-art approaches by incorporating both response-response and query-response interactions. In contrast to existing works, our pipeline provides a confidence score and optimal response as well. Further details on the comparison study experimental setup can be found in 
Appendix~\ref{sec:compare_exp}.


\subsection{Human evaluation}
We conduct a human evaluation on 500 samples each from the Jeopardy and Kaggle datasets to assess whether the pipeline-selected response is optimal. These samples are taken from the Human annotated test set. For this analysis, we choose the optimal responses from the \textit{`k'} sample responses for each query. We instruct the expert evaluators to indicate whether they agree or disagree with the pipeline-selected optimal response, based on the classification label (hallucinated or non-hallucinated). In the HA-test, for the Jeopardy dataset, we achieve 82.4\% agreement, whereas for the Kaggle dataset, the agreement is 84\%. Moreover, on the LL-70B-Gen test set, we obtain 75.8\%, and 86\% scores for Jeopardy and Kaggle datasets.
\subsection{Error analysis}
We perform the error analysis to understand the effectiveness of the proposed HalluCounter approach. We manually verify 500 samples each from HA-Test and LL-70B-Gen. Each category error analysis details are outlined in Table~\ref{tab:error_analysis}. The following are the major error categories, where the proposed pipeline might exhibit sub-standard performance.\\
\textbf{1. Misclassification.} The HalluCounter pipeline makes incorrect predictions, due to \textit{a). Complete inconsistency} among the sample responses, which is against the core principle of the design of the HalluCounter approach. \textit{b). Partial inconsistency.} The number of incorrect responses is greater than correct responses in total sample responses, \textit{c). Pipeline inefficiency.} The HalluCounter pipeline might fail due to the inefficacy of one or more components including measuring NLI scores, classifier prediction, or optimal response selection .\\
\textbf{2. Answer denial.} \textit{a). Insufficient context.} LLMs refuse to answer the query either due to insufficient context or ambiguous information present in the query. \textit{b). Problematic context.} Presence of misleading, violent, or contradictory information in the query. The corresponding examples for all the error categories are illustrated in 
Appendix~\ref{sec:err_analysis} Table~\ref{tab:error_cases_examples}.

%% file: table_features_combinations.tex

\begin{table}[t]
\centering\footnotesize
\setlength{\tabcolsep}{1.5pt} 
\begin{tabular}{l|c c c|c c c|c c}
\toprule
 & \multicolumn{3}{c|}{\textbf{Q-R}} & \multicolumn{3}{c|}{\textbf{R-R}} & \multicolumn{2}{c}{\textbf{Text}} \\ \cmidrule{2-9}
\multirow{-2}{*}{\textbf{Combination}} & E & C & N & E & C & N & Query (q) & Response (r) \\ \midrule
C-C           &   & \checkmark &   &   & \checkmark &   &   &   \\
EC-EC         & \checkmark & \checkmark &   & \checkmark & \checkmark &   &   &   \\
Q-R           & \checkmark & \checkmark & \checkmark &   &   &   &   &   \\
R-R           &   &   &   & \checkmark & \checkmark & \checkmark &   &   \\
(Q-R)+(R-R)    & \checkmark & \checkmark & \checkmark & \checkmark & \checkmark & \checkmark &   &   \\
q-r+(Q-R)+(R-R) & \checkmark & \checkmark & \checkmark & \checkmark & \checkmark & \checkmark & \checkmark & \checkmark \\ \bottomrule
\end{tabular}
\caption{NLI features combinations; E, C, N denote Entailment, Contradiction, and Neutral features.}
\label{tab:features_combination}
\vspace{-6mm}
\end{table}

%% file: table_hallucounter_varying_samples.tex
\begin{table*}[]
\centering
\resizebox{\textwidth}{!}{%
\begin{tabular}{@{}clccccccccccccccccccccc@{}} \\
\toprule
\multicolumn{1}{l}{} &  & \multicolumn{3}{c}{\textbf{TL-1.1B-Gen}} & \multicolumn{3}{c}{\textbf{PHI-3.5B-Gen}} & \multicolumn{3}{c}{\textbf{MST-7B-Gen}} & \multicolumn{3}{c}{\textbf{LL-8B-Gen}} & \multicolumn{3}{c}{\textbf{GM-7B-Gen}} & \multicolumn{3}{c}{\textbf{LL-70B-Gen}} & \multicolumn{3}{c}{\textbf{ENSB-Gen}} \\ \midrule
\multicolumn{1}{l}{} &  & 3 & 5 & 10 & 3 & 5 & 10 & 3 & 5 & 10 & 3 & 5 & 10 & 3 & 5 & 10 & 3 & 5 & 10 & 3 & 5 & 10 \\ \cmidrule(lr){3-5}\cmidrule(lr){6-8}\cmidrule(lr){9-11}\cmidrule(lr){12-14}\cmidrule(lr){15-17}\cmidrule(lr){18-20}\cmidrule(lr){21-23}
\multirow{3}{*}{\textbf{Jeopardy}} & F1 & 0.75 & 0.75 & 0.75 & 0.71 & 0.71 & 0.71 & 0.68 & 0.68 & 0.68 & 0.82 & 0.82 & 0.81 & 0.63 & 0.63 & 0.62 & 0.54 & 0.54 & 0.54 & 0.74 & 0.74 & 0.73 \\
 & B-ACC & 0.93 & 0.93 & 0.93 & 0.75 & 0.75 & 0.75 & 0.82 & 0.82 & 0.82 & 0.80 & 0.79 & 0.79 & 0.67 & 0.67 & 0.67 & 0.44 & 0.44 & 0.44 & 0.84 & 0.85 & 0.84 \\
 & ROC & 0.74 & 0.74 & 0.75 & 0.78 & 0.78 & 0.79 & 0.75 & 0.76 & 0.76 & 0.89 & 0.88 & 0.88 & 0.70 & 0.69 & 0.70 & 0.60 & 0.60 & 0.60 & 0.83 & 0.83 & 0.83 \\ \midrule
\multirow{3}{*}{\textbf{Kaggle}} & F1 & 0.83 & 0.84 & 0.83 & 0.70 & 0.70 & 0.70 & 0.54 & 0.54 & 0.54 & 0.75 & 0.75 & 0.75 & 0.66 & 0.66 & 0.66 & 0.79 & 0.79 & 0.79 & 0.75 & 0.75 & 0.75 \\
 & B-ACC & 0.92 & 0.93 & 0.93 & 0.63 & 0.61 & 0.60 & 0.65 & 0.65 & 0.65 & 0.63 & 0.64 & 0.65 & 0.72 & 0.72 & 0.72 & 0.70 & 0.70 & 0.68 & 0.80 & 0.79 & 0.80 \\
 & ROC & 0.68 & 0.67 & 0.68 & 0.66 & 0.65 & 0.64 & 0.54 & 0.55 & 0.55 & 0.70 & 0.69 & 0.70 & 0.66 & 0.66 & 0.66 & 0.77 & 0.77 & 0.76 & 0.72 & 0.72 & 0.73 \\ \bottomrule
\end{tabular}}
\caption{HalluCounter performance with varying the number of sample responses.}
\label{tab:hallucounter_varying_k}
\vspace{-5mm}
\end{table*}

%% file: table_hallu_eval.tex
\begin{table}[t]
\centering\footnotesize
\begin{tabular}{@{}lccc@{}}
\toprule
 & \multicolumn{1}{l}{} & \textbf{Jeopardy} & \textbf{Kaggle} \\ \midrule
\multirow{3}{*}{HaluEval Datasets} & Summarization & 0.60 & 0.70 \\
 & QA & 0.77 & 0.78 \\
 & Dialogue & 0.93 & 0.9 \\ \bottomrule 
\end{tabular}
\caption{HalluCounter performance on HaluEval.}
\label{tab:hallueval_experiments_results}
\end{table}


%% file: table_pipeline_results.tex
\begin{table}[t]
\centering\small
\setlength{\tabcolsep}{0.3ex}
\begin{tabular}{@{}c|c|lllc|cccc@{}}
\toprule
\multirow{2}{*}{} & \multicolumn{1}{c|}{\multirow{2}{*}{\textbf{Test set}}} & \multicolumn{4}{c|}{\textbf{Hallucination rate}} & \multicolumn{4}{c}{\textbf{Confidence score}} \\ \cmidrule(l){3-10}
 & \multicolumn{1}{l|}{} & \multicolumn{1}{c}{K=3} & \multicolumn{1}{c}{K=5} & \multicolumn{1}{c}{K=7} & \multicolumn{1}{c|}{K=10} & \multicolumn{1}{c}{K=3} & \multicolumn{1}{c}{K=5} & \multicolumn{1}{c}{K=7} & K=10 \\ \midrule
\multirow{8}{*}{\rotatebox{90}{\textbf{Jeopardy}}} & TL-1.1B-Gen  & \multicolumn{1}{c}{86} & \multicolumn{1}{c}{88} & \multicolumn{1}{c}{88} & 87 & \multicolumn{1}{r}{91} & \multicolumn{1}{r}{89} & \multicolumn{1}{r}{88} & 88\\  
 & PHI-3.5B-Gen & \multicolumn{1}{c}{53} & \multicolumn{1}{c}{53} & \multicolumn{1}{c}{53} & 51 & \multicolumn{1}{r}{92} & \multicolumn{1}{r}{91} & \multicolumn{1}{r}{90} & 90 \\ 
 & LL-8B-Gen  & \multicolumn{1}{c}{29} & \multicolumn{1}{c}{28} & \multicolumn{1}{c}{28} & 26 & \multicolumn{1}{r}{94} & \multicolumn{1}{r}{93} & \multicolumn{1}{r}{93} & 93\\  
 & MST-7B-Gen  & \multicolumn{1}{c}{59} & \multicolumn{1}{c}{59} & \multicolumn{1}{c}{58} & 55 & \multicolumn{1}{r}{88} & \multicolumn{1}{r}{86} & \multicolumn{1}{r}{84} & 84\\  
 & GM-7B-Gen  & \multicolumn{1}{c}{38} & \multicolumn{1}{c}{37} & \multicolumn{1}{c}{37} & 36 & \multicolumn{1}{r}{95} & \multicolumn{1}{r}{94} & \multicolumn{1}{r}{93} & 93\\  
 & LL-70B-Gen  & \multicolumn{1}{c}{17} & \multicolumn{1}{c}{17} & \multicolumn{1}{c}{17} & 17 & \multicolumn{1}{r}{100} & \multicolumn{1}{r}{100} & \multicolumn{1}{r}{100} & 100 \\  
 & ENSB-Gen  & \multicolumn{1}{c}{53} & \multicolumn{1}{c}{53} & \multicolumn{1}{c}{53} & 51 & \multicolumn{1}{r}{91} & \multicolumn{1}{r}{90} & \multicolumn{1}{r}{89} & 88 \\  
 & HA-Test & \multicolumn{1}{c}{53} & \multicolumn{1}{c}{53} & \multicolumn{1}{c}{54} & 52 & \multicolumn{1}{r}{87} & \multicolumn{1}{r}{84} & \multicolumn{1}{r}{83} & 82 \\ \midrule
\multirow{8}{*}{\rotatebox{90}{\textbf{Kaggle}}} &  TL-1.1B-Gen  & \multicolumn{1}{c}{87} & \multicolumn{1}{c}{87} & \multicolumn{1}{c}{87} & 86 & \multicolumn{1}{r}{96} & \multicolumn{1}{r}{95} & \multicolumn{1}{r}{95} & 95\\ 
 & PHI-3.5B-Gen  & \multicolumn{1}{c}{67} & \multicolumn{1}{c}{67} & \multicolumn{1}{c}{67} & 66 & \multicolumn{1}{r}{96} & \multicolumn{1}{r}{95} & \multicolumn{1}{r}{95} & 95\\ 
 &  LL-8B-Gen  & \multicolumn{1}{c}{63} & \multicolumn{1}{c}{63} & \multicolumn{1}{c}{64} & 62 & \multicolumn{1}{r}{93} & \multicolumn{1}{r}{92} & \multicolumn{1}{r}{92} & 92 \\  
 & MST-7B-Gen  & \multicolumn{1}{c}{76} & \multicolumn{1}{c}{76} & \multicolumn{1}{c}{76} & 75  & \multicolumn{1}{r}{95} & \multicolumn{1}{r}{94} & \multicolumn{1}{r}{93} & 93\\ 
 & GM-7B-Gen  & \multicolumn{1}{c}{73} & \multicolumn{1}{c}{73} & \multicolumn{1}{c}{73} & 72 & \multicolumn{1}{r}{95} & \multicolumn{1}{r}{94} & \multicolumn{1}{r}{93} & 93\\  
 & LL-70B-Gen* & \multicolumn{1}{c}{68} & \multicolumn{1}{c}{67} & \multicolumn{1}{c}{67} & 66 & \multicolumn{1}{r}{95} & \multicolumn{1}{r}{94} & \multicolumn{1}{r}{93} & 93 \\ 
 & ENSB-Gen  & \multicolumn{1}{c}{53} & \multicolumn{1}{c}{53} & \multicolumn{1}{c}{53} & 51 & \multicolumn{1}{r}{91} & \multicolumn{1}{r}{90} & \multicolumn{1}{r}{89} & 88\\  
 &  HA-Test  & \multicolumn{1}{c}{65} & \multicolumn{1}{c}{67} & \multicolumn{1}{c}{68} & 66 & \multicolumn{1}{r}{88} & \multicolumn{1}{r}{85} & \multicolumn{1}{r}{84} & 84 \\ \bottomrule
\end{tabular}
\caption{HalluCounter pipeline results by varying number of sample responses (\textit{`K'}); The results of best-performing model for each test is reported. * denotes the quantized version. All the values are in percentages.}
\label{tab:pipeline_results}
\vspace{-6mm}
\end{table}

%% file: table_SOTA_comparison.tex


\begin{table}[]
\centering\footnotesize
\setlength{\tabcolsep}{0.5ex}
\begin{tabular}{@{}clcc@{}}
\toprule
\textbf{Type of method}                                            &           \textbf{Approach}                                     & \textbf{Jeopardy}                   & \textbf{Kaggle}                     \\ \midrule
\multicolumn{1}{c}{\multirow{2}{*}{Response-consistency}}     & \multicolumn{1}{l}{SelfCheckGPT}              & \multicolumn{1}{c|}{0.651} & \multicolumn{1}{c}{0.674} \\ 
\multicolumn{1}{c}{}                                    & \multicolumn{1}{l}{InterrogateLLM}            & \multicolumn{1}{c|}{0.427} & \multicolumn{1}{c}{0.671} \\ \midrule
\multicolumn{1}{c}{\multirow{3}{*}{Uncertainty-based}} & \multicolumn{1}{l}{Perplexity}                & \multicolumn{1}{c|}{0.487} & \multicolumn{1}{c}{0.678} \\ 
\multicolumn{1}{c}{}                                    & \multicolumn{1}{l}{LN-Entropy} & \multicolumn{1}{c|}{0.441} & \multicolumn{1}{c}{0.707} \\ 
\multicolumn{1}{c}{}                                    & \multicolumn{1}{l}{LexicalSimilarity}         & \multicolumn{1}{c|}{0.442} & \multicolumn{1}{c}{0.711} \\ \midrule
\multicolumn{1}{c}{\multirow{3}{*}{Training-based}}  & \multicolumn{1}{l}{HaloScope}                    & \multicolumn{1}{c|}{0.323} & \multicolumn{1}{c}{0.402} \\ 
\multicolumn{1}{c}{}                                & \multicolumn{1}{l}{EigenScore}                & \multicolumn{1}{c|}{0.437} & \multicolumn{1}{c}{0.658} \\ 
\multicolumn{1}{c}{}                                    & \multicolumn{1}{l}{SAPLMA}                    & \multicolumn{1}{c|}{0.668} & \multicolumn{1}{c}{0.716} \\ 

\midrule
\multicolumn{1}{l}{}                                      & \textbf{HalluCounter}                                   & \textbf{0.743}                      & \textbf{0.782}                      \\ \bottomrule
\end{tabular}
\caption{Comparison with state-of-the-art approaches, all the values are F1-scores.}
\label{tab:cross_compare}
\vspace{-5mm}
\end{table}

%% file: table_error_classification.tex
\begin{table}[t]
\centering\small
\setlength{\tabcolsep}{0.3ex}
\begin{tabular}{llccc|cc}
\toprule
\multicolumn{1}{l}{} &  & \multicolumn{3}{c|}{\textbf{Misclassification}} & \multicolumn{2}{c}{\textbf{Answer Denial}} \\ \midrule
\multicolumn{1}{l}{} &  & \multicolumn{1}{c}{C1} & \multicolumn{1}{c}{C2} & \multicolumn{1}{c|}{C3} & \multicolumn{1}{c}{C4} & \multicolumn{1}{c}{C5} \\ \cmidrule{2-5}\cmidrule{6-7}
\multirow{2}{*}{LL-70B-Gen} & Jeopardy & 21.4 & 0 & 2 & 0 & 0 \\
 & Kaggle & \phantom{0}5.2 & \phantom{0}6.2 & \phantom{0}2.8 & 0 & 0 \\ \cmidrule{2-7}
\multirow{2}{*}{HA-Test} & Jeopardy & \phantom{0}8.4 & \phantom{0}3.2 & \phantom{0}2.6 & \phantom{0}1.4 & 1 \\
 & Kaggle & 11.4 & \phantom{0}0.8 & \phantom{0}3.6 & \phantom{0}3.8 & 0 \\ \bottomrule
\end{tabular}
\caption{Error analysis of 500 samples for the following error categories, C1) Complete inconsistency, C2) Partial inconsistency, C3) Pipeline inefficiency, C4) Insufficient context, C5) Problematic context; Each value represents percentages of error instances. }
\label{tab:error_analysis}
\vspace{-6mm}
\end{table}

%% file: table_kaggle_results.tex
\begin{table*}
\centering
\resizebox{\textwidth}{!}{%
\begin{tabular}{@{}c|c|ccc|ccc|ccc|ccc|ccc|ccc@{}}
\toprule
 &  & \multicolumn{3}{c}{\textbf{QR}} & \multicolumn{3}{c}{\textbf{RR}} & \multicolumn{3}{c}{\textbf{EC-EC}} & \multicolumn{3}{c}{\textbf{CC}} & \multicolumn{3}{c}{\textbf{QR-RR}} & \multicolumn{3}{c}{\textbf{q-r+Q-R+R-R}} \\ \midrule
\textbf{Test Data} & \textbf{Classifier}  & \multicolumn{1}{c}{\textbf{F1}} & \multicolumn{1}{c}{\textbf{AUC}} & \textbf{B-ACC} & \multicolumn{1}{c}{\textbf{F1}} & \multicolumn{1}{c}{\textbf{AUC}} & \textbf{B-ACC} & \multicolumn{1}{c}{\textbf{F1}} & \multicolumn{1}{c}{\textbf{AUC}} & \textbf{B-ACC} & \multicolumn{1}{c}{\textbf{F1}} & \multicolumn{1}{c}{\textbf{AUC}} & \textbf{B-ACC} & \multicolumn{1}{c}{\textbf{F1}} & \multicolumn{1}{c}{\textbf{AUC}} & \textbf{B-ACC} & \multicolumn{1}{c}{\textbf{F1}} & \multicolumn{1}{c}{\textbf{AUC}} & \textbf{B-ACC} \\ \midrule

\multirow{2}{*}{TL-1.1B-Gen} & Statistical & \multicolumn{1}{c}{0.71} & \multicolumn{1}{c}{0.60} & 0.88 & \multicolumn{1}{c}{0.80} & \multicolumn{1}{c}{0.61} & 0.92 & \multicolumn{1}{c}{0.82} & \multicolumn{1}{c}{0.68} & 0.93 & \multicolumn{1}{c}{0.73} & \multicolumn{1}{c}{0.62} & 0.90 & \multicolumn{1}{c}{0.83} & \multicolumn{1}{c}{0.68} & 0.93 & \multicolumn{1}{c}{-} & \multicolumn{1}{c}{-} & - \\
 & BERT & \multicolumn{1}{c}{0.82} & \multicolumn{1}{c}{0.60} & 0.88 & \multicolumn{1}{c}{0.63} & \multicolumn{1}{c}{0.61} & 0.88 & \multicolumn{1}{c}{0.85} & \multicolumn{1}{c}{0.70} & 0.94 & \multicolumn{1}{c}{0.74} & \multicolumn{1}{c}{0.64} & 0.91 & \multicolumn{1}{c}{0.85} & \multicolumn{1}{c}{0.70} & 0.94 & \multicolumn{1}{c}{\textbf{0.86}} & \multicolumn{1}{c}{0.76} & 0.94 \\ \midrule

\multirow{2}{*}{PHI-3.5B-Gen} & Statistical & \multicolumn{1}{c}{0.58} & \multicolumn{1}{c}{0.50} & 0.49 & \multicolumn{1}{c}{0.66} & \multicolumn{1}{c}{0.63} & 0.60 & \multicolumn{1}{c}{0.68} & \multicolumn{1}{c}{0.62} & 0.59 & \multicolumn{1}{c}{0.61} & \multicolumn{1}{c}{0.54} & 0.51 & \multicolumn{1}{c}{0.70} & \multicolumn{1}{c}{0.64} & 0.60 & \multicolumn{1}{c}{-} & \multicolumn{1}{c}{-} & - \\
 & BERT  & \multicolumn{1}{c}{0.68} & \multicolumn{1}{c}{0.65} & 0.62 & \multicolumn{1}{c}{0.66} & \multicolumn{1}{c}{0.52} & 0.50 & \multicolumn{1}{c}{0.70} & \multicolumn{1}{c}{0.65} & 0.61 & \multicolumn{1}{c}{0.66} & \multicolumn{1}{c}{0.55} & 0.51 & \multicolumn{1}{c}{0.71} & \multicolumn{1}{c}{0.65} & 0.62 & \multicolumn{1}{c}{\textbf{0.77}} & \multicolumn{1}{c}{0.71} & 0.65 \\ \midrule

\multirow{2}{*}{LL-8B-Gen} & Statistical & \multicolumn{1}{c}{0.56} & \multicolumn{1}{c}{0.53} & 0.51 & \multicolumn{1}{c}{0.73} & \multicolumn{1}{c}{0.69} & 0.64 & \multicolumn{1}{c}{0.75} & \multicolumn{1}{c}{0.70} & 0.65 & \multicolumn{1}{c}{0.63} & \multicolumn{1}{c}{0.60} & 0.56 & \multicolumn{1}{c}{0.75} & \multicolumn{1}{c}{0.70} & 0.65 & \multicolumn{1}{c}{-} & \multicolumn{1}{c}{-} & - \\ 
 & BERT& \multicolumn{1}{c}{0.76} & \multicolumn{1}{c}{0.72} & 0.66 & \multicolumn{1}{c}{0.65} & \multicolumn{1}{c}{0.56} & 0.52 & \multicolumn{1}{c}{\textbf{0.77}} & \multicolumn{1}{c}{0.73} & 0.67 & \multicolumn{1}{c}{0.72} & \multicolumn{1}{c}{0.65} & 0.61 & \multicolumn{1}{c}{\textbf{0.77}} & \multicolumn{1}{c}{0.72} & 0.66 & \multicolumn{1}{c}{\textbf{0.77}} & \multicolumn{1}{c}{0.75} & 0.69 \\ \midrule

\multirow{2}{*}{MST-7B-Gen} & Statistical& \multicolumn{1}{c}{0.53} & \multicolumn{1}{c}{0.53} & 0.66 & \multicolumn{1}{c}{0.56} & \multicolumn{1}{c}{0.49} & 0.64 & \multicolumn{1}{c}{0.54} & \multicolumn{1}{c}{0.47} & 0.62 & \multicolumn{1}{c}{0.54} & \multicolumn{1}{c}{0.55} & 0.66 & \multicolumn{1}{c}{0.54} & \multicolumn{1}{c}{0.55} & 0.65 & \multicolumn{1}{c}{-} & \multicolumn{1}{c}{-} & - \\ 
 & BERT & \multicolumn{1}{c}{0.55} & \multicolumn{1}{c}{0.52} & 0.64 & \multicolumn{1}{c}{0.53} & \multicolumn{1}{c}{0.51} & 0.64 & \multicolumn{1}{c}{0.53} & \multicolumn{1}{c}{0.54} & 0.65 & \multicolumn{1}{c}{0.53} & \multicolumn{1}{c}{0.45} & 0.61 & \multicolumn{1}{c}{0.53} & \multicolumn{1}{c}{0.51} & 0.63 & \multicolumn{1}{c}{0.56} & \multicolumn{1}{c}{0.68} & 0.74 \\ \midrule

\multirow{2}{*}{GM-7B-Gen} & Statistical  & \multicolumn{1}{c}{0.60} & \multicolumn{1}{c}{0.53} & 0.62 & \multicolumn{1}{c}{\textbf{0.67}} & \multicolumn{1}{c}{0.67} & 0.72 & \multicolumn{1}{c}{0.66} & \multicolumn{1}{c}{0.66} & 0.71 & \multicolumn{1}{c}{0.63} & \multicolumn{1}{c}{0.60} & 0.67 & \multicolumn{1}{c}{\textbf{0.67}} & \multicolumn{1}{c}{0.67} & 0.72 & \multicolumn{1}{c}{-} & \multicolumn{1}{c}{-} & - \\ 
 & BERT  & \multicolumn{1}{c}{\textbf{0.67}} & \multicolumn{1}{c}{0.68} & 0.73 & \multicolumn{1}{c}{0.68} & \multicolumn{1}{c}{0.55} & 0.63 & \multicolumn{1}{c}{0.64} & \multicolumn{1}{c}{0.68} & 0.73 & \multicolumn{1}{c}{\textbf{0.67}} & \multicolumn{1}{c}{0.62} & 0.68 & \multicolumn{1}{c}{0.66} & \multicolumn{1}{c}{0.67} & 0.71 & \multicolumn{1}{c}{0.65} & \multicolumn{1}{c}{0.70} & 0.75 \\ \midrule

 \multirow{2}{*}{LL-70B-Gen} & Statistical & \multicolumn{1}{c}{0.55} & \multicolumn{1}{c}{0.49} & 0.48 & \multicolumn{1}{c}{0.79} & \multicolumn{1}{c}{0.77} & 0.71 & \multicolumn{1}{c}{0.80} & \multicolumn{1}{c}{0.78} & 0.72 & \multicolumn{1}{c}{0.61} & \multicolumn{1}{c}{0.60} & 0.58 & \multicolumn{1}{c}{0.79} & \multicolumn{1}{c}{0.76} & 0.68 & \multicolumn{1}{c}{-} & \multicolumn{1}{c}{-} & - \\
 & BERT & \multicolumn{1}{c}{0.83} & \multicolumn{1}{c}{0.80} & 0.73 & \multicolumn{1}{c}{0.65} & \multicolumn{1}{c}{0.55} & 0.52 & \multicolumn{1}{c}{\textbf{0.84}} & \multicolumn{1}{c}{0.81} & 0.74 & \multicolumn{1}{c}{0.72} & \multicolumn{1}{c}{0.65} & 0.60 & \multicolumn{1}{c}{0.82} & \multicolumn{1}{c}{0.80} & 0.72 & \multicolumn{1}{c}{0.80} & \multicolumn{1}{c}{0.80} & 0.73 \\ \midrule

\multirow{2}{*}{ENSB-Gen} & Statistical & \multicolumn{1}{c}{0.60} & \multicolumn{1}{c}{0.53} & 0.65 & \multicolumn{1}{c}{0.73} & \multicolumn{1}{c}{0.72} & 0.80 & \multicolumn{1}{c}{0.76} & \multicolumn{1}{c}{0.72} & 0.80 & \multicolumn{1}{c}{0.66} & \multicolumn{1}{c}{0.60} & 0.72 & \multicolumn{1}{c}{0.75} & \multicolumn{1}{c}{0.73} & 0.81 & \multicolumn{1}{c}{-} & \multicolumn{1}{c}{-} & - \\ 
 & BERT & \multicolumn{1}{c}{0.77} & \multicolumn{1}{c}{0.74} & 0.78 & \multicolumn{1}{c}{0.64} & \multicolumn{1}{c}{0.54} & 0.65 & \multicolumn{1}{c}{0.78} & \multicolumn{1}{c}{0.76} & 0.82 & \multicolumn{1}{c}{0.69} & \multicolumn{1}{c}{0.63} & 0.73 & \multicolumn{1}{c}{0.79} & \multicolumn{1}{c}{0.75} & 0.82 & \multicolumn{1}{c}{\textbf{0.80}} & \multicolumn{1}{c}{0.83} & 0.86 \\ \midrule

\multirow{2}{*}{HA-Test} & Statistical& \multicolumn{1}{c}{0.65} & \multicolumn{1}{c}{0.51} & 0.70 & \multicolumn{1}{c}{0.76} & \multicolumn{1}{c}{0.66} & 0.82 & \multicolumn{1}{c}{\textbf{0.78}} & \multicolumn{1}{c}{0.69} & 0.82 & \multicolumn{1}{c}{0.70} & \multicolumn{1}{c}{0.59} & 0.77 & \multicolumn{1}{c}{0.77} & \multicolumn{1}{c}{0.70} & 0.82 & \multicolumn{1}{c}{-} & \multicolumn{1}{c}{-} & - \\ 
 & BERT  & \multicolumn{1}{c}{0.23} & \multicolumn{1}{c}{0.50} & 0.68 & \multicolumn{1}{c}{0.59} & \multicolumn{1}{c}{0.50} & 0.68 & \multicolumn{1}{c}{0.23} & \multicolumn{1}{c}{0.50} & 0.68 & \multicolumn{1}{c}{0.59} & \multicolumn{1}{c}{0.50} & 0.68 & \multicolumn{1}{c}{0.23} & \multicolumn{1}{c}{0.50} & 0.68 & \multicolumn{1}{c}{0.68} & \multicolumn{1}{c}{0.76} & 0.81 \\ \bottomrule
\end{tabular}
}
\caption{Hallucination classifier results on various test sets from Kaggle dataset, \textbf{AUC:} Area Under Curve, \textbf{B-ACC:} Balanced Accuracy. All the values are the average scores of four Kaggle datasets, with the best result in \textbf{bold}.}
\label{tab:kaggle_results}
\vspace{-5mm}
\end{table*}

%% file: 6_discussion.tex
\section{Discussion and Insights}
\begin{figure*}[t]
    \centering
    \includegraphics[width=1\linewidth]{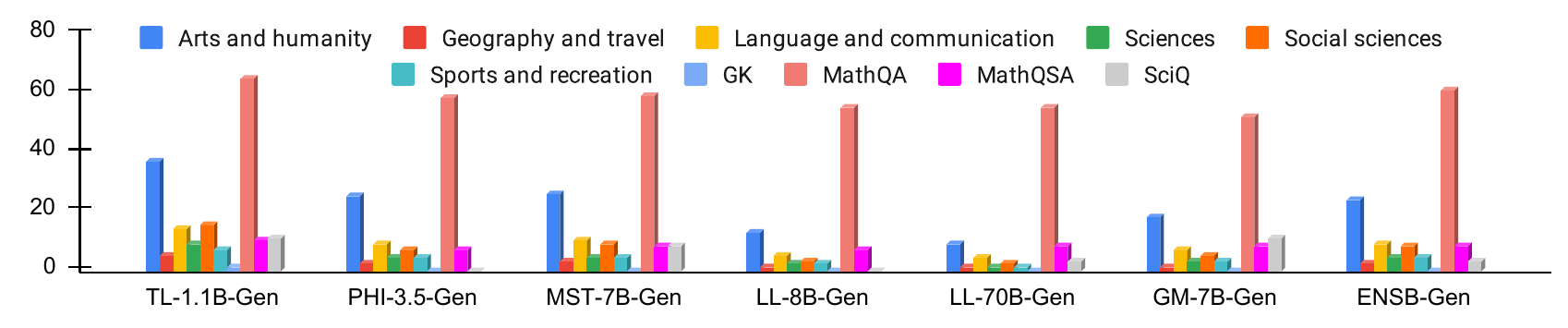}
    \caption{ Hallucination rates across different sub-domains in various test sets of the Jeopardy and Kaggle datasets.}
    \label{fig:halu_rates_jeopardy}
    \vspace{-6mm}
\end{figure*}
\textbf{Performance across various domains.} As shown in Table~\ref{tab:pipeline_results}, all LLMs exhibit a higher tendency to hallucinate on the Kaggle test sets compared to the Jeopardy test sets. Specifically, Figure~\ref{fig:halu_rates_jeopardy} reveals that LLMs experience the highest hallucination rates on questions related to ``MathQA'', ``arts and humanity'', followed by ``language and communication'', with the lowest rates occurring in the ``GK'' and ``Geography and travel'' categories. It is evident from our study that, the majority of LLMs face significant challenges with queries demanding mathematical reasoning \cite{srivatsa-kochmar-2024-makes, ahn-etal-2024-large} and scientific factual knowledge \cite{yang-zhao-2024-llms}. 

\noindent \textbf{High resiliency.} The confidence score in HalluCounter reflects the level of resiliency in determining whether a response is hallucinated. As presented in Table~\ref{tab:pipeline_results}, despite the slight variations in the hallucination rates with varying numbers of sample responses, the proposed pipeline consistently achieves an average confidence score above 90\% across both the Jeopardy and Kaggle test sets. From this result, it is evident that the performance of the HalluCounter pipeline remains largely unchanged regardless of the number of sample responses.\\

\noindent \textbf{LLMs hallucination rate.} To assess which LLMs are highly prone to hallucination, we compare overall prediction with the actual label. As shown in Table~\ref{tab:pipeline_results}, we find that for the Jeopardy dataset TinyLLaMA-1.1B and Mistral-7B models are more likely to generate hallucinated responses, and LLaMA-3-70B produces the least percentage of hallucinations. Whereas in the case of Kaggle datasets TinyLLaMA-1.1B, Mistral-7B, and Gemma-7B models are prone to higher hallucination. The models that failed on the Jeopardy dataset lack logical reasoning capabilities because most of the Jeopardy dataset consists of hint-based general knowledge questions. \\

\noindent \textbf{NLI model robustness.} We notice that often the NLI model assigns high scores to longer LLM response sequences and unseen premise-hypothesis pairs \cite{yang2024improving}, which leads to high entailment and contradiction scores. In such cases, the classifier might exhibit mediocre performance.\\
\noindent \textbf{Assessing the ambiguity.} Since most of the Jeopardy dataset questions are hint-based, there is a possibility of providing a biased answer to an ambiguous question that could have multiple correct answers \cite{park-kim-2025-llms}. In such cases, the HalluCounter pipeline might struggle to classify it as either accurate or hallucinated. Similarly, in a few instances, the labeling model Qwen2.5-32B fails to perform accurate semantic matching.

%% file: 2_Related_works.tex
\section{Background on Hallucination detection}
Hallucinations in LLMs remain an enduring challenge across text, image, audio, and video \citep{sahoo2024unveiling, li-etal-2024-reference}, and detecting them is crucial, especially when no external reference or ground truth is available.

Self-consistency approaches gained a lot of attention in detecting the factual correctness in the LLM-generated responses. Approaches such as SelfCheckGPT \citep{manakul-etal-2023-selfcheckgpt}, which relies on the principle of self-consistency among the stochastically generated responses and detects the hallucination based on whether the generated responses support the original answer. $SAC^{3}$ \cite{zhang-etal-2023-sac3} detect hallucination by analyzing cross-model consistency and cross-rephrased queries. InterrogateLLM \citep{yehuda-etal-2024-interrogatellm}, detects hallucination by asking the reverse question and verifies whether the original question can be generated. LogicCheckGPT \cite{wu-etal-2024-logical}, asks LLMs questions with logical correlations to detect hallucination. SELF-FAMILIARITY \cite{luo-etal-2024-zero-resource} focuses on evaluating the model's familiarity with the concepts present in the instruction. 

Several approaches leverage LLM's internal representations to detect hallucination, by training a classifier using the LLM's hidden representations \cite{azaria-mitchell-2023-internal}, weighting LLMs' expertise \cite{wei2024measuring}, by calculating the probability of each token in the given text \cite{liu-etal-2022-token}, measuring the semantic consistency across various generations in embedding space \cite{cheninside}. Additionally, uncertainty-based estimation approaches based on aleatoric and epistemic uncertainty have been studied to detect hallucination in auto-regressive generation \cite{xiao-wang-2021-hallucination, malininuncertainty}. However, these approaches are limited to white-box models.

We draw inspiration from the SelfCheckGPT, which uses the normalized scores of entailment and contradiction NLI scores between the responses to detect the hallucinations. In contrast, our approach leverages query-response and response-response consistency and alignment patterns to train a hallucination detection classifier. Additionally, unlike existing methods, our pipeline provides the least hallucinated response among all the responses along with overall prediction and the corresponding confidence score.

%% file: 7_conclusions_and_limitations.tex
\section{Conclusion}
In this work, we propose HalluCounter, a novel method for RFHD in LLMs. This method improves response consistency-based hallucination detection methods and generates confidence scores and optimal responses along with hallucination detection. We introduce a large-scale HalluCounterEval dataset, which consists of a large set of synthetic and human-annotated samples across diverse domains. Through extensive experiments and ablations, we evaluate various NLI feature combinations, classifiers, and labeling strategies. Additionally, we offer a detailed error analysis, key insights, and takeaways from our method and benchmark dataset.

\section{Limitations}
This paper proposes a novel reference-free hallucination detection pipeline, despite the best efforts, our paper still has several limitations. (1) Synthetic datasets creation: To create synthetic train and test sets, we experiment with zero-shot prompting only, and to increase the quality of the datasets further studies can experiment with few-shot and Chain-of-thought prompting strategies as well. (2) Cross-encoder module sensitivity towards longer sequences: The classifier heavily relies on the cross-encoder module to obtain NLI logit values, however the cross-encode module is prone to provide high entailment values for longer sequences, which might lead to inaccurate classifier prediction. (3) Inconsistency among sample responses: Our approach works on the principle of self-consistency among the sample responses, we face challenges if all the responses are hallucinated in that case our approach may exhibit mediocre performance. (4) Computational complexity: Despite HalluCounter's superior performance compared to state-of-the-art approaches, it is quite computationally heavy, which could be addressed in future work to be made more efficient.

\section{Ethics Statement}
In this work, we utilize only the publicly available datasets. We make all the synthetic and human-annotated datasets public to encourage reproducibility. Moreover, by tackling the issue of hallucinations in LLMs, this work points out that undetected hallucinations could lead to misinformation.

%% file: 8_appendix.tex
\section{HalluCounterEval dataset filtration details}
\label{sec:jeopardy_filter}
The datasets present in the HaluCounterEval (Jeopardy and Kaggle) undergo rule-based filtering stages to ensure quality and consistency before being split into training and test sets. The following filtration steps are common to all the training and testing datasets.
\begin{itemize}
    \item Initial Dataset: The raw dataset consists of question-answer pairs collected from their respective sources.
    \item Removal of URLs: Questions containing URLs in the text are filtered out.
    \item Exclusion of ``Fill-in-the-Blank`` Questions: Questions with dashes (representing blanks) are excluded from the dataset.
    \item Elimination of Short Questions: Questions with fewer than five words are removed to maintain sufficient context.
\end{itemize}
\begin{figure}[b]
    \centering
    \includegraphics[width=0.50\textwidth]{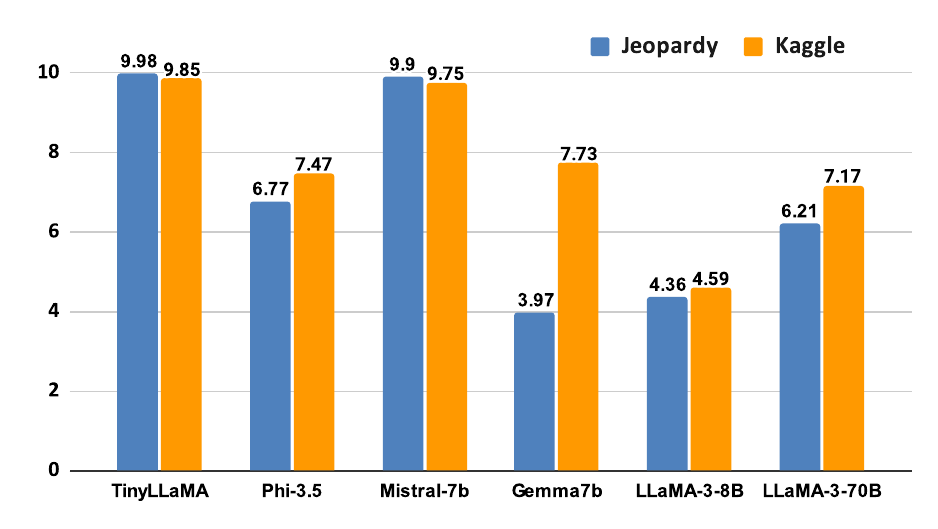}
    \caption{Number of unique responses generated by each LLM out of 10 responses for Jeopardy and Kaggle datasets. The lower the number represents the higher consistency.}
    \label{fig:unique_responses}
\end{figure}
\input{table_jeopardy_data_stats}
\input{table_kaggle_dataset_stats}
\input{appendix_tables/table_model_sources}

\section{Train and test dataset details}
\label{sec:train_test}
The training and testing dataset statistics of Jeopardy and Kaggle are detailed in Table~\ref{tab:halugurad_data} and \ref{tab:kaggle_data_stats}. All the values are in Table~\ref{tab:halugurad_data} and \ref{tab:kaggle_data_stats} corresponding to total number of unique queries. We generate 10 samples per each query and obtain 10 times of the total unique samples for the purposes of training and testing. Moreover, the jeopardy dataset comprises of 6 major categories and 22 sub-categories of various domains of data. Whereas, the Kaggle dataset consists of four different datasets including scientific, general knowledge, and mathematical domain factoid question-answer pairs. Further, as shown in Figure~\ref{fig:unique_responses}, TinyLLaMA-1.1B has the highest number of unique responses followed by Mistral-7B model.

\input{table_qwen_labeling_prompt}
\input{appendix_tables/jeopardy_hallucination_rates}
\input{appendix_tables/table_error_cases_example}
\input{table_gpt4_inference}
\input{table_cross-comparison}

\input{appendix_tables/jeopardy_confidence_scores}

\section{More results for the Hallucination classifier}
\label{sec:halu_classifer_results}
We perform a series of experiments across multiple test sets, using different classifiers and labeling strategies for both the Jeopardy and Kaggle datasets. 
 \subsection{Results on Jeopardy dataset}
We built various classifiers to detect hallucination in LLMs. For all the best-performing models, hallucination classifier results are detailed in Table~\ref{tab:jeopardy_results}. Moreover, we report the results of the statistical approach-based hallucination classifier trained on the Jeopardy dataset with labels obtained from the exact-match approach in Table~\ref{tab:string_jeopardy_ml}, LLM-based approach in Table~\ref{tab:string_jeopardy_bert}. Similarly, the BERT classifier is trained on the Jeopardy dataset with labels obtained from the exact-match approach in Table~\ref{tab:string_jeopardy_bert}, LLM-based approach in Table~\ref{tab:qwen_jeopardy_bert}. Additionally, we report each category-wise result for the Jeopardy dataset in Table~\ref{tab:jeopardy_category_wise}. 
\input{table_jeopardy_results}
\subsection{Results on Kaggle dataset}
We report the results of the statistical approach-based hallucination classifier trained on the Kaggle dataset with labels obtained from the exact-match approach in Table~\ref{tab:string_ml_kaggle}, LLM-based approach in Table~\ref{tab:string_kaggle_bert}. Similarly, the BERT classifier is trained on the Kaggle dataset with labels obtained from the exact-match approach in Table~\ref{tab:qwen_kaggle_ml}, LLM-based approach in Table~\ref{tab:qwen_kaggle_bert}. Additionally, we test the efficiency of the classifier on four different Kaggle datasets, and the corresponding results are mentioned in Table~\ref{tab:kaggle_datasetwise}.
\subsection{Experiments with additional features}
We additionally include two token-based features for training the classifier: the length of the LLM-generated response and the number of punctuation marks it contains. Incorporating these features alongside the NLI-based features yields a modest improvement in overall classifier accuracy. Experimental results on the HA-Test dataset are presented in Table~\ref{tab:results_token_counts}.
\input{appendix_tables/results_token_counts_experiment}

\input{table_jeopardy_category_wise}

\input{table_kaggle_datasetwise}
\input{table_generate_responses}

\section{Sample responses generation}
\label{sec:responses_generation}
As mentioned in Table~\ref{tab:responses_prompt}, we use the same prompt \textit{`k'} times to generate \textit{`k'} responses each time to avoid the mismatch in the total number of sample responses for each query. We did the inference with various LLMs by using the same prompt. While generating the data for training, we set the \textit{`k'} value to 10.
\subsection{LLM inference configuration details}
We did the inference with various small and large language models. Across all the models we use the max\_new\_tokens=32, top\_k=50, top\_p=0.95, and temperature=1. Additionally, we did the necessary response parsing to obtain only the relevant information related to the given query.

\section{Labeling using Qwen2.5-32B Model}
\label{sec:qwen_prompt_labeling}
We perform the labeling using the Qwen2.5-32B \cite{yang2024qwen2} to classify whether each LLM response is hallucinated or non-hallucinated. We used the prompt mentioned in Table~\ref{tab:qwen_labeling_prompt} to perform the labeling. 
\section{Comparison experiments details}
\label{sec:compare_exp}
We compare our approach with two popularly known reference-free hallucination detection approaches, which are SelfCheckGPT \cite{manakul-etal-2023-selfcheckgpt} and InterrogateLLM \cite{yehuda-etal-2024-interrogatellm}.\\
\textbf{SelfCheckGPT.} To compare with the SelfCheckGPT approach, we utilize the prompt variant approach, where by providing the context, sentence and instruct the Qwen2.5-32B \cite{yang2024qwen2} LLM to whether the sentence is supported by the context or not. The final inconsistency score is computed by averaging the sentence scores.   \\
\textbf{InterrogateLLM.} To compare with the InterrogateLLM approach, first, we create a few-shot prompt with question and answer pairs. In the forward pass, we generate an answer to each question and in the back-ward pass obtain the 10 questions to the same answer by modifying the few-shot prompt. In the end, by measuring the average cosine similarity between the original question and generated questions, we classify the question with more than 0.91 threshold as non-hallucinated. In the forward and backward process, we utilize the LLaMA3-8B model for inference.\\

\section{Generalization experiments}
To verify the generalizability of the HalluCounter approach, we train the HalluCounter on Jeopardy, test on Kaggle, and perform the vice-versa experiments, and the corresponding results are detailed in Table~\ref{tab:cross_comparison}.
\subsection{Hallucounter performance with varying number of sample responses}
We conduct experiments to analyze the performance of HalluCounter while varying the number of sample responses obtained from the LLM and the corresponding results are outlined in Table~\ref{tab:hallucounter_varying_k}. From the results, it is evident that despite varying the K values, there is no significant variation in the accuracies across various tests for both the Jeopardy and Kaggle datasets. This indicates that our proposed HalluCounter pipeline is stable across different K values.

\input{table_all_results_analysis}

\section{Experimental setup}
\label{sec:exp_setup_info}
We conduct all experiments using two Nvidia GeForce RTX A6000 (48GB) GPUs. We do not perform the hyperparameter search. The maximum sequence length for classifier training with various feature combinations is set to 200, except for the `q-r+Q-R-R-R', where it is set to 512. All other configurations follow the default settings of the Hugging Face trainer\footnote{\url{https://huggingface.co/docs/transformers/main_classes/trainer}}. The huggingface models used in the experiments along with their sources are detailed in Table~\ref{tab:model_sources}.

\subsection{Conversion of numerical to textual features}
The following template is used to convert the numerical features into textual features for training the classifier. The template takes into account the question, response, and several scores related to query-response and response-response entailment, neutrality, and contradiction.
\begin{itemize}
    \item \textbf{Question}: The given question is the text input represented by \texttt{Question}.
    \item \textbf{Response}: The given response from the model is represented by \texttt{Response}.
    \item \textbf{Query-Response Entailment Score}: The numerical score indicating the entailment score obtained between the query to response, represented by \texttt{feature\_1}.
    \item \textbf{Query-Response Neutral Score}: The numerical score representing the neutral score obtained between the query to response, represented by \texttt{feature\_2}.
    \item \textbf{Query-Response Contradiction Score}: The numerical score representing the contradiction score obtained between the query to response, represented by \texttt{feature\_3}.
    \item \textbf{Response-Response Entailment Score}: The numerical score indicating the entailment score obtained between the response to response, represented by \texttt{feature\_4}.
    \item \textbf{Response-Response Neutral Score}: The numerical score representing the neutral score obtained between the response to response, represented by \texttt{feature\_5}.
    \item \textbf{Response-Response Contradiction Score}: The numerical score representing the contradiction score obtained between the response to response, represented by \texttt{feature\_6}.
\end{itemize}
\noindent This conversion process generates a structured textual feature that combines the question, response, and scores in the following format:
\begin{quote}
    ``The given question is \texttt{\{Question\}} and the corresponding answer is \texttt{\{Response\}}, and they got the query-response entailment score: \texttt{\{feature\_1\}}, neutral score: \texttt{\{feature\_2\}}, and contradiction score: \texttt{\{feature\_3\}}. And they got the response-response entailment score: \texttt{\{feature\_4\}}, neutral score: \texttt{\{feature\_5\}}, contradiction score: \texttt{\{feature\_6\}}."
\end{quote}
This textual feature is used as input for the classifier.
\input{appendix_tables/table_string_jeopardy_ML_appendix}
\input{appendix_tables/table_string_kaggle_ML_appendix}

\section{Error analysis examples}
\label{sec:err_analysis}
We observe various error cases, where our HalluCounter pipeline fails to do the accurate classification and optimal response selection. The corresponding examples are detailed in Table~\ref{tab:error_cases_examples}. 

\section{HalluCounter performance on GPT4}
\label{sec:gpt4_testing}
To understand the efficiency of the HalluCounter pipeline on closed-source models, we ran our pipeline on the samples generated using the GPT4o-mini \cite{achiam2023gpt} LLM. We utilized the queries from the Human annotated dataset and generated 10 responses to each query and obtained the corresponding NLI scores. As shown in Table~\ref{tab:gpt4_inference_scores}, the GPT4o-mini model exhibits 13.5\% hallucination rate on Jeopardy and 22.6\% on the Kaggle dataset queries. Moreover, our HalluCounter pipeline exhibits more than 80\% prediction confidence.
\section{Category-wise hallucination rates and confidence scores}
We perform the category-wise results analysis to understand the category-wise hallucination rates for all test sets corresponding to the Jeopardy and Kaggle datasets. All the hallucination rates details are mentioned in Table~\ref{tab:jeopardy_hallucination_rates}
and corresponding confidence scores are listed in Table~\ref{tab:jeopardy_confidence_scores}.

\input{appendix_tables/table_qwen_jeopardy_ML_appendix}
\input{appendix_tables/table_qwen_kaggle_ML_appendix}
\input{appendix_tables/table_string_jeopardy_BERT_appendix}

\input{appendix_tables/table_string_kaggle_BERT_appendix}
\input{appendix_tables/table_qwen_jeopardy_BERT_appendix}
\input{appendix_tables/table_qwen_kaggle_BERT_appendix}

%% file: table_jeopardy_data_stats.tex
\begin{table}[t]
\centering
\resizebox{\columnwidth}{!}{
\begin{tabular}{@{}l|l|l|l}
\toprule
\textbf{Main category}                       & \textbf{Sub-category} & \textbf{Train} & \textbf{Test} \\ \midrule
\multirow{9}{*}{Arts and Humanities} & Authors        &  843   &  94 \\ \cmidrule(l){2-4} 
                                     & Books           &  997   &  111 \\ \cmidrule(l){2-4} 
                                     & Culture       &   300  &  33 \\ \cmidrule(l){2-4} 
                                     & Literature   &  1370   &  152 \\ \cmidrule(l){2-4} 
                                     & Movies        &   1426  &  159  \\ \cmidrule(l){2-4} 
                                     & Music      &  2581   & 287  \\ \cmidrule(l){2-4} 
                                     & TV      & 2272    &  253  \\ \midrule
\multirow{4}{*}{Geography and travel}                & Geography    &   1245   & 138\\ \cmidrule(l){2-4}
                                     & Rivers       & 320   & 35\\ \cmidrule(l){2-4}
                                     & Travel       & 535 	& 60 \\ \midrule
\multirow{2}{*}{Language and communication}           & Language     & 526 	& 58 \\ \cmidrule(l){2-4}
                                     & Words    & 3424 & 380   \\ \midrule
\multirow{3}{*}{Sciences}                             &  Animals   &550 & 61  \\ \cmidrule(l){2-4}
                                     & Physics    &189 &	21 \\ \cmidrule(l){2-4}
                                     &   Science  & 1819 & 202 \\ \midrule
\multirow{6}{*}{Social sciences}     & Education  &137 	&15 \\ \cmidrule(l){2-4} 
                                     & History  &3245 	&361 \\ \cmidrule(l){2-4} 
                                     & Law    &233	&26 \\ \cmidrule(l){2-4} 
                                     & Politics &259 	&29 \\ \cmidrule(l){2-4} 
                                     & Presidents   &547 	&61  \\ \midrule

\multirow{2}{*}{Sports and recreation}                & Awards   &335 	&37 \\ \cmidrule(l){2-4}
                                     & Sports   &1512 	&168\\ \midrule
                                     & \textbf{Total} & \textbf{24665} & \textbf{2741} \\
                                     \bottomrule
\end{tabular}%
}
\caption{Jeopardy dataset statistics.}
\label{tab:halugurad_data}
\end{table}

%% file: table_kaggle_dataset_stats.tex
\begin{table}[t]
\centering
\resizebox{\columnwidth}{!}{
\begin{tabular}{@{}cccccc@{}}
\toprule
      & MathQA & MathQSA & SciQ & GK & \textbf{Total}\\ \midrule
Train &     32980   &   4956      & 12102      &  657 & \textbf{50695} \\ \midrule
Test &     3665   &    550     &    1345   &   73  & \textbf{5633}\\ \bottomrule
\end{tabular}}
\caption{Kaggle dataset statistics.}
\label{tab:kaggle_data_stats}
\end{table}

%% file: appendix_tables/table_model_sources.tex
\begin{table}
\centering
\resizebox{\columnwidth}{!}{%
\begin{tabular}{l|l}
\toprule
\textbf{Model} & \multicolumn{1}{c}{\textbf{Source}}                                              \\ \midrule
TinyLlama-1.1B & \url{https://huggingface.co/TinyLlama/TinyLlama-1.1B-Chat-v1.0} \\ 
Gemma-7B       & \url{https://huggingface.co/google/gemma-7b-it}                 \\ 
Mistral-7B     & \url{https://huggingface.co/mistralai/Mistral-7B-Instruct-v0.1} \\ 
Phi-3.5B       & \url{https://huggingface.co/microsoft/Phi-3.5-mini-instruct}    \\ 
Llama-8B       & \url{https://huggingface.co/meta-llama/Llama-3.1-8B-Instruct}   \\ 
Llama-70B      & \url{https://huggingface.co/Groq/Llama-3-Groq-70B-Tool-Use}     \\ 
Qwen-32B       & \url{https://huggingface.co/Qwen/Qwen2.5-32B-Instruct}          \\ \bottomrule
\end{tabular}%
}
\caption{Source of Huggingface models.}
\label{tab:model_sources}
\end{table}

%% file: table_qwen_labeling_prompt.tex
\begin{table*}[ht]
\centering
\resizebox{\textwidth}{!}{%
\begin{tabular}{c|l}
\toprule
\textbf{Role} & \textbf{Content} \\
\midrule
System & You are Qwen, created by Alibaba Cloud. You are a helpful assistant. \\ \hline
\multirow{11}{*}{User} & You are a helpful assistant tasked with evaluating whether a model-generated response is hallucinated or not. \\
      & Here is the context: \\
      & Question: \texttt{\{question\}} \\
      & Correct Answer: \texttt{\{gold\_answer\}} \\
      & Model Response: \texttt{\{llm\_response\}} \\
      & \\
      & Your task is as follows: \\
      & 1. Check if the correct answer or its meaningful variations (e.g., initials, abbreviations, synonyms) appear in the model response. \\
      & 2. If the correct answer (or a variation) is present, even partially, and the essence of correctness is captured, label it as '0' (not hallucinated). \\
      & 3. If the correct answer or meaningful variations are completely absent or contradicted, label it as '1' (hallucinated). \\
      & 4. Provide only the label (1 or 0) as your output. Do not include any additional information. \\
\bottomrule
\end{tabular}}
\caption{Prompt for classifying whether LLM generated response is hallucinated or not}
\label{tab:qwen_labeling_prompt}
\end{table*}

%% file: appendix_tables/jeopardy_hallucination_rates.tex
\begin{table*}
\resizebox{\textwidth}{!}{
\begin{tabular}{ccccccccc}
\toprule
\textbf{Dataset }&\textbf{ Category}                     &\textbf{ ENSB-Gen} &\textbf{ GM-7B-Gen }& \textbf{LL-70B-Gen} &\textbf{ LL-8B-Gen}  & \textbf{MST-7B-Gen }& \textbf{PHI-3.5B-Gen} &\textbf{ TL-1.1B-Gen} \\ \midrule 
\multirow{6}{*}{Jeopardy}& Arts and humanity          & \phantom{0}24       & \phantom{0}18    & 9           & \phantom{0}13         & \phantom{0}26      & \phantom{0}25   & \phantom{0}37        \\ 
& Geography and travel       & 2        & 1     & 1           & 1          & 3       & 2    & 5         \\ 
& Language and communication & 9        & 7     & 4           & 5          & \phantom{0}10      & 9    & \phantom{0}14        \\ 
& Sciences                     & 4        & 3     & 1           & 2          & 4       & 4    & 9         \\ 
& Social sciences             & 8        & 5     & 2           & 3          & 9       & 7    & \phantom{0}15        \\ 
& Sports and recreation      & 4        & 3     & 1           & 2          & 4       & 4    & 7       \\
\midrule
\multirow{4}{*}{Kaggle}& GK       & 0        & 0         & 0                  & 0          & 0       & 0    & 1         \\ 
& MathQA   & \phantom{0}61       & \phantom{0}52        & \phantom{0}55                 & \phantom{0}55         & \phantom{0}59      & \phantom{0}58   & \phantom{0}65        \\ 
& MathQSA  & 8        & 8         & 8                  & 7          & 8       & 7    & \phantom{0}10        \\ 
& SciQ     & 3        & \phantom{0}11        & 3                  & 0          & 8       & 0    & \phantom{0}11        \\
\bottomrule
\end{tabular}
}
\caption{Hallucination rate for each category in Jeopardy and Kaggle datasets across various test sets generated by LLMs; all the values are in percentages.}
\label{tab:jeopardy_hallucination_rates}
\end{table*}

%% file: appendix_tables/table_error_cases_example.tex
\begin{table*}
\centering\small
\begin{tabularx}{\textwidth}{lX}
\toprule
\multicolumn{1}{c}{\textbf{Error category}} & \multicolumn{1}{c}{\textbf{Examples}} \\
\midrule
Complete inconsistency &  \begin{tabular}[c]{@{}l@{}} \textbf{Question:} A record from years ago that's still worth listening to is an oldie but this \\ 
\textbf{Correct answer:} Goodie \\
\textbf{Sample responses:} [`Goldfinger.', `Goldfinger', `Goldfinger.', `Gold.', `gold.',\\ `goldmine.', `Goldfinger.', `Goldfish Crackers', `gold.', `Goldfinger.'] \\ 
\textbf{Optimal response:} Goldfinger \\ \textbf{Classification:} Non-hallcuinated \end{tabular}  \\
\midrule

Partial inconsistency & \begin{tabular}[c]{@{}l@{}} \textbf{Question:} Type of machine you can use to send \& receive letters or photos over the telephone line \\
\textbf{Correct answer:} A fax \\
\textbf{Sample responses:} [`Modem.', `Modem.', `Modem (Modulator-Demodulator).', `Modem.',\\ `Modem.', `Modem.', `Facsimile machine or- Fax machine.', `Fax machine.', \\ `Modem.', `Modem.'] \\
\textbf{Optimal response:} Modem.\\
\textbf{Classification:} Non-hallucinated \end{tabular}   \\
\midrule

Pipeline failure & \begin{tabular}[c]{@{}l@{}} \textbf{Question:} What is the innermost layer of the sun called? \\ 
\textbf{Correct answer:} The core \\
\textbf{Sample responses:} [`The core.', `Core.', `Core.', `The core.', `Core.', `Core.', `The inner most \\ layer of the sun is called the core.', `The core.', `The inner most layer of the sun is called \\ the Core.', `The core.'] \\ 
\textbf{Optimal response:} The core \\ 
\textbf{Classification:} Hallucinated \end{tabular} \\
\midrule

Insufficient context & \begin{tabular}[c]{@{}l@{}} \textbf{Question:} Could go in an ``nct" category.  Period \\
\textbf{Correct answer:} succinct \\
\textbf{Sample responses:} ['The answer is: Not applicable to the context.', 'Okay.', 'The category you \\ requested is not relevant to the given question, so I am unable to provide an answer.', 'Alright.', \\'Certainly.', 'Alright.', 'Okay.', 'Insufficient context provided for a response.', 'Insufficient\\ information provided to categorize.', 'Okay.'] \\
\textbf{Optimal response:} The answer is: Not applicable to the context. \\
\textbf{Classification:} Hallucinated
\end{tabular}  \\ 
\midrule

Problematic context & \begin{tabular}[c]{@{}l@{}} \textbf{Question:} "I didn't kill my wife!" \\
\textbf{Correct answer:} The Fugitive \\
\textbf{Sample responses:} [`I am unable to provide responses that suggest innocence or guilt in criminal \\ matters.', `I am unable to provide responses that suggest or deny involvement in criminal  \\ activities.', `I am unable to provide responses that suggest or imply criminal actions or deception.', \\ `I am unable to provide responses that may contain misleading or potentially incriminating \\ information.', `I am unable to provide responses that may contain misleading or contradictory \\ information.', `I am unable to provide responses that suggest innocence in criminal activities or \\potentially misleading information.', `I am unable to provide responses that may be misleading or \\ potentially conceal criminal activity.', `I am unable to provide responses that suggest or imply \\criminal activity or harmful actions.', `I am unable to provide responses that may provide\\ misleading or potentially incriminating information.', `I am unable to provide subjective \\ information or opinions, including personal claims of innocence.'] \\
\textbf{Optimal response:} I am unable to provide responses that may contain misleading or contradictory \\ information.\\
\textbf{Classification:} Hallucinated\end{tabular} \\
\bottomrule
\end{tabularx}
\caption{Examples for different error categories.}
\label{tab:error_cases_examples}
\end{table*}

%% file: table_gpt4_inference.tex
\begin{table}
\centering\small
\begin{tabular}{@{}lcc@{}}
\toprule
 & \multicolumn{2}{c}{\textbf{HA-Test}} \\ \cmidrule{2-3} 
 & Jeopardy & Kaggle \\  \midrule
Hallucination rate & \phantom{0}13.5 & \phantom{0}22.6 \\
Confidence score & 90 & 83 \\ \bottomrule
\end{tabular}%
\caption{HalluCounter performance on GPT-4o-mini generated sample responses; all the values are in percentages.}
\label{tab:gpt4_inference_scores}
\end{table}

%% file: table_cross-comparison.tex
\begin{table}[]
\centering\small
\begin{tabular}{@{}lllll@{}}
\toprule
{ \textbf{Trained}} & { \textbf{Tested}} & { \textbf{F1-Score}} & { \textbf{B-ACC}} & { \textbf{AUC}} \\ \midrule
{ } & { Jeopardy} & { 0.73} & { 0.86} & { 0.82} \\
\multirow{-2}{*}{{ \textbf{Jeopardy}}} & { Kaggle} & { 0.77} & { 0.61} & { 0.80} \\ \midrule
{ } & { Kaggle} & { 0.66} & { 0.82} & { 0.76} \\
\multirow{-2}{*}{{ \textbf{Kaggle}}} & { Jeopardy} & { 0.68} & { 0.82} & { 0.79} \\ \bottomrule
\end{tabular}
\caption{Cross comparison experiments results.}
\label{tab:cross_comparison}
\end{table}

%% file: appendix_tables/jeopardy_confidence_scores.tex
\begin{table*}
\resizebox{\textwidth}{!}{
\begin{tabular}{ccccccccc}
\toprule
\textbf{Dataset }&\textbf{ Category}                     &\textbf{ ENSB-Gen} &\textbf{ GM-7B-Gen }& \textbf{LL-70B-Gen} &\textbf{ LL-8B-Gen}  & \textbf{MST-7B-Gen }& \textbf{PHI-3.5B-Gen} &\textbf{ TL-1.1B-Gen} \\ \midrule 
\multirow{6}{*}{Jeopardy}& Arts and humanity          & 88       & 92    & 100         & 92         & 85      & 90   & 90        \\ 
& Geography and travel       & 88       & 96    & 100         & 95         & 81      & 90   & 79        \\ 
& Language and communication & 87       & 92    & 100         & 92         & 85      & 89   & 91        \\ 
& Sciences                     & 89       & 94    & 100         & 93         & 83      & 89   & 84        \\ 
& Social sciences             & 89       & 94    & 100         & 94         & 81      & 90   & 85        \\ 
& Sports and recreation      & 89       & 94    & 100         & 92         & 83      & 90   & 89        \\ \midrule 
\multirow{4}{*}{Kaggle}& GK       & 93      & 98        & 93\phantom{0}                  & 96         & 89      & 96   & 85        \\ 
& MathQA   & 96       & 92        & 94\phantom{0}                 & 90         & 96      & 95   & \phantom{0}100       \\ 
& MathQSA  & 93       & 96        & 92\phantom{0}                 & 89         & 91      & 87   & 98        \\ 
& SciQ     & 91       & 93        & 89\phantom{0}                 & 96         & 87      & 97   & 80        \\ \bottomrule
\end{tabular}
}
\caption{Confidence Score for each category in Jeopardy and Kaggle datasets across various test sets generated by LLMs. All the values are in percentages.}
\label{tab:jeopardy_confidence_scores}
\end{table*}

%% file: table_jeopardy_results.tex
\begin{table*}
\centering
\resizebox{\textwidth}{!}{%
\begin{tabular}{@{}c|c|c|ccc|ccc|ccc|ccc|ccc|ccc@{}}
\toprule
 &  &  & \multicolumn{3}{c|}{\textbf{QR}} & \multicolumn{3}{c|}{\textbf{RR}} & \multicolumn{3}{c|}{\textbf{EC-EC}} & \multicolumn{3}{c|}{\textbf{CC}} & \multicolumn{3}{c|}{\textbf{QR-RR}} & \multicolumn{3}{c}{\textbf{q-r+Q-R+R-R}} \\ \midrule
\textbf{Test Data} & \textbf{Classifier} & \textbf{Labeling} & \multicolumn{1}{c}{\textbf{F1}} & \multicolumn{1}{c}{\textbf{AUC}} & \textbf{B-ACC} & \multicolumn{1}{c}{\textbf{F1}} & \multicolumn{1}{c}{\textbf{AUC}} & \textbf{B-ACC} & \multicolumn{1}{c}{\textbf{F1}} & \multicolumn{1}{c}{\textbf{AUC}} & \textbf{B-ACC} & \multicolumn{1}{c}{\textbf{F1}} & \multicolumn{1}{c}{\textbf{AUC}} & \textbf{B-ACC} & \multicolumn{1}{c}{\textbf{F1}} & \multicolumn{1}{c}{\textbf{AUC}} & \textbf{B-ACC} & \multicolumn{1}{c}{\textbf{F1}} & \multicolumn{1}{c}{\textbf{AUC}} & \textbf{B-ACC} \\ \midrule
\multirow{4}{*}{TL-1.1B-Gen} & \multirow{2}{*}{Statistical} & Exact-match & \multicolumn{1}{c}{0.68} & \multicolumn{1}{c}{0.58} & 0.90 & \multicolumn{1}{c}{0.74} & \multicolumn{1}{c}{0.57} & 0.90 & \multicolumn{1}{c}{0.76} & \multicolumn{1}{c}{0.64} & 0.92 & \multicolumn{1}{c}{0.69} & \multicolumn{1}{c}{0.59} & 0.91 & \multicolumn{1}{c}{0.76} & \multicolumn{1}{c}{0.63} & 0.91 & \multicolumn{1}{c}{-} & \multicolumn{1}{c}{-} & - \\ 
 &  & LLM-based & \multicolumn{1}{c}{0.65} & \multicolumn{1}{c}{0.67} & 0.90 & \multicolumn{1}{c}{0.73} & \multicolumn{1}{c}{0.68} & 0.90 & \multicolumn{1}{c}{0.75} & \multicolumn{1}{c}{0.75} & 0.93 & \multicolumn{1}{c}{0.64} & \multicolumn{1}{c}{0.62} & 0.89 & \multicolumn{1}{c}{0.75} & \multicolumn{1}{c}{0.75} & 0.93 & \multicolumn{1}{c}{-} & \multicolumn{1}{c}{-} & - \\ \cmidrule(l){3-21} 
 & \multirow{2}{*}{BERT} & Exact-match & \multicolumn{1}{c}{0.81} & \multicolumn{1}{c}{0.59} & 0.90 & \multicolumn{1}{c}{0.64} & \multicolumn{1}{c}{0.62} & 0.92 & \multicolumn{1}{c}{0.82} & \multicolumn{1}{c}{0.67} & 0.92 & \multicolumn{1}{c}{0.70} & \multicolumn{1}{c}{0.61} & 0.91 & \multicolumn{1}{c}{0.81} & \multicolumn{1}{c}{0.66} & 0.92 & \multicolumn{1}{c}{\textbf{0.88}} & \multicolumn{1}{c}{0.90} & 0.99 \\ 
 &  & LLM-based & \multicolumn{1}{c}{0.77} & \multicolumn{1}{c}{0.70} & 0.91 & \multicolumn{1}{c}{0.55} & \multicolumn{1}{c}{0.67} & 0.91 & \multicolumn{1}{c}{0.79} & \multicolumn{1}{c}{0.78} & 0.94 & \multicolumn{1}{c}{0.53} & \multicolumn{1}{c}{0.61} & 0.89 & \multicolumn{1}{c}{0.78} & \multicolumn{1}{c}{0.78} & 0.94 & \multicolumn{1}{c}{0.82} & \multicolumn{1}{c}{0.84} & 0.96 \\ \midrule

\multirow{4}{*}{PHI-3.5B-Gen} & \multirow{2}{*}{Statistical} & Exact-match & \multicolumn{1}{c}{0.55} & \multicolumn{1}{c}{0.59} & 0.69 & \multicolumn{1}{c}{0.63} & \multicolumn{1}{c}{0.69} & 0.77 & \multicolumn{1}{c}{0.65} & \multicolumn{1}{c}{0.71} & 0.79 & \multicolumn{1}{c}{0.58} & \multicolumn{1}{c}{0.61} & 0.72 & \multicolumn{1}{c}{0.65} & \multicolumn{1}{c}{0.71} & 0.79 & \multicolumn{1}{c}{-} & \multicolumn{1}{c}{-} & - \\ 
 &  & LLM-based & \multicolumn{1}{c}{0.53} & \multicolumn{1}{c}{0.58} & 0.56 & \multicolumn{1}{c}{0.69} & \multicolumn{1}{c}{0.77} & 0.74 & \multicolumn{1}{c}{0.71} & \multicolumn{1}{c}{0.79} & 0.76 & \multicolumn{1}{c}{0.58} & \multicolumn{1}{c}{0.63} & 0.61 & \multicolumn{1}{c}{0.71} & \multicolumn{1}{c}{0.79} & 0.75 & \multicolumn{1}{c}{-} & \multicolumn{1}{c}{-} & - \\ \cmidrule(l){3-21} 
 & \multirow{2}{*}{BERT} & Exact-match & \multicolumn{1}{c}{0.81} & \multicolumn{1}{c}{0.59} & 0.90 & \multicolumn{1}{c}{0.64} & \multicolumn{1}{c}{0.62} & 0.92 & \multicolumn{1}{c}{0.82} & \multicolumn{1}{c}{0.67} & 0.92 & \multicolumn{1}{c}{0.70} & \multicolumn{1}{c}{0.61} & 0.91 & \multicolumn{1}{c}{0.81} & \multicolumn{1}{c}{0.66} & 0.92 & \multicolumn{1}{c}{\textbf{0.88}} & \multicolumn{1}{c}{0.90} & 0.99 \\ 
 &  & LLM-based & \multicolumn{1}{c}{0.72} & \multicolumn{1}{c}{0.80} & 0.77 & \multicolumn{1}{c}{0.56} & \multicolumn{1}{c}{0.63} & 0.58 & \multicolumn{1}{c}{0.71} & \multicolumn{1}{c}{0.81} & 0.78 & \multicolumn{1}{c}{0.63} & \multicolumn{1}{c}{0.68} & 0.66 & \multicolumn{1}{c}{0.71} & \multicolumn{1}{c}{0.81} & 0.78 & \multicolumn{1}{c}{0.79} & \multicolumn{1}{c}{0.86} & 0.84 \\ \midrule

\multirow{4}{*}{LL-8B-Gen} & \multirow{2}{*}{Statistical} & Exact-match & \multicolumn{1}{c}{0.51} & \multicolumn{1}{c}{0.57} & 0.57 & \multicolumn{1}{c}{0.68} & \multicolumn{1}{c}{0.76} & 0.76 & \multicolumn{1}{c}{0.69} & \multicolumn{1}{c}{0.76} & 0.76 & \multicolumn{1}{c}{0.65} & \multicolumn{1}{c}{0.69} & 0.70 & \multicolumn{1}{c}{0.68} & \multicolumn{1}{c}{0.75} & 0.75 & \multicolumn{1}{c}{-} & \multicolumn{1}{c}{-} & - \\ 
 &  & LLM-based & \multicolumn{1}{c}{0.55} & \multicolumn{1}{c}{0.64} & 0.47 & \multicolumn{1}{c}{0.81} & \multicolumn{1}{c}{0.88} & 0.78 & \multicolumn{1}{c}{0.82} & \multicolumn{1}{c}{0.88} & 0.78 & \multicolumn{1}{c}{0.75} & \multicolumn{1}{c}{0.80} & 0.68 & \multicolumn{1}{c}{0.81} & \multicolumn{1}{c}{0.88} & 0.79 & \multicolumn{1}{c}{-} & \multicolumn{1}{c}{-} & - \\ \cmidrule(l){3-21} 
 & \multirow{2}{*}{BERT} & Exact-match & \multicolumn{1}{c}{0.69} & \multicolumn{1}{c}{0.78} & 0.79 & \multicolumn{1}{c}{0.58} & \multicolumn{1}{c}{0.62} & 0.61 & \multicolumn{1}{c}{0.68} & \multicolumn{1}{c}{0.79} & 0.79 & \multicolumn{1}{c}{0.66} & \multicolumn{1}{c}{0.74} & 0.74 & \multicolumn{1}{c}{0.69} & \multicolumn{1}{c}{0.79} & 0.79 & \multicolumn{1}{c}{0.73} & \multicolumn{1}{c}{0.82} & 0.82 \\ 
 &  & LLM-based & \multicolumn{1}{c}{0.82} & \multicolumn{1}{c}{0.90} & 0.81 & \multicolumn{1}{c}{0.69} & \multicolumn{1}{c}{0.69} & 0.51 & \multicolumn{1}{c}{0.82} & \multicolumn{1}{c}{0.90} & 0.81 & \multicolumn{1}{c}{0.78} & \multicolumn{1}{c}{0.84} & 0.73 & \multicolumn{1}{c}{0.82} & \multicolumn{1}{c}{0.90} & 0.81 & \multicolumn{1}{c}{\textbf{0.84}} & \multicolumn{1}{c}{0.90} & 0.83 \\ \midrule

\multirow{4}{*}{MST-7B-Gen} & \multirow{2}{*}{Statistical} & Exact-match & \multicolumn{1}{c}{0.58} & \multicolumn{1}{c}{0.58} & 0.76 & \multicolumn{1}{c}{0.63} & \multicolumn{1}{c}{0.66} & 0.80 & \multicolumn{1}{c}{0.65} & \multicolumn{1}{c}{0.68} & 0.82 & \multicolumn{1}{c}{0.59} & \multicolumn{1}{c}{0.59} & 0.77 & \multicolumn{1}{c}{0.65} & \multicolumn{1}{c}{0.68} & 0.82 & \multicolumn{1}{c}{-} & \multicolumn{1}{c}{-} & - \\ 
 &  & LLM-based & \multicolumn{1}{c}{0.58} & \multicolumn{1}{c}{0.63} & 0.74 & \multicolumn{1}{c}{0.66} & \multicolumn{1}{c}{0.73} & 0.78 & \multicolumn{1}{c}{0.69} & \multicolumn{1}{c}{0.76} & 0.82 & \multicolumn{1}{c}{0.57} & \multicolumn{1}{c}{0.62} & 0.73 & \multicolumn{1}{c}{0.68} & \multicolumn{1}{c}{0.76} & 0.82 & \multicolumn{1}{c}{-} & \multicolumn{1}{c}{-} & - \\ \cmidrule(l){3-21} 
 & \multirow{2}{*}{BERT} & Exact-match & \multicolumn{1}{c}{0.70} & \multicolumn{1}{c}{0.69} & 0.82 & \multicolumn{1}{c}{0.57} & \multicolumn{1}{c}{0.58} & 0.76 & \multicolumn{1}{c}{0.7} & \multicolumn{1}{c}{0.72} & 0.83 & \multicolumn{1}{c}{0.64} & \multicolumn{1}{c}{0.62} & 0.79 & \multicolumn{1}{c}{0.69} & \multicolumn{1}{c}{0.70} & 0.82 & \multicolumn{1}{c}{0.81} & \multicolumn{1}{c}{0.89} & 0.95 \\ 
 &  & LLM-based & \multicolumn{1}{c}{0.70} & \multicolumn{1}{c}{0.76} & 0.80 & \multicolumn{1}{c}{0.56} & \multicolumn{1}{c}{0.65} & 0.75 & \multicolumn{1}{c}{0.72} & \multicolumn{1}{c}{0.80} & 0.84 & \multicolumn{1}{c}{0.53} & \multicolumn{1}{c}{0.64} & 0.75 & \multicolumn{1}{c}{0.71} & \multicolumn{1}{c}{0.79} & 0.84 & \multicolumn{1}{c}{\textbf{0.81}} & \multicolumn{1}{c}{0.89} & 0.93 \\ \midrule

\multirow{4}{*}{GM-7B-Gen} & \multirow{2}{*}{Statistical} & Exact-match & \multicolumn{1}{c}{0.56} & \multicolumn{1}{c}{0.48} & 0.68 & \multicolumn{1}{c}{0.59} & \multicolumn{1}{c}{0.63} & 0.77 & \multicolumn{1}{c}{0.58} & \multicolumn{1}{c}{0.61} & 0.75 & \multicolumn{1}{c}{0.59} & \multicolumn{1}{c}{0.57} & 0.74 & \multicolumn{1}{c}{0.55} & \multicolumn{1}{c}{0.61} & 0.74 & \multicolumn{1}{c}{-} & \multicolumn{1}{c}{-} & - \\ 
 &  & LLM-based & \multicolumn{1}{c}{0.54} & \multicolumn{1}{c}{0.59} & 0.57 & \multicolumn{1}{c}{0.62} & \multicolumn{1}{c}{0.69} & 0.66 & \multicolumn{1}{c}{0.63} & \multicolumn{1}{c}{0.70} & 0.67 & \multicolumn{1}{c}{0.63} & \multicolumn{1}{c}{0.66} & 0.64 & \multicolumn{1}{c}{0.62} & \multicolumn{1}{c}{0.70} & 0.67 & \multicolumn{1}{c}{-} & \multicolumn{1}{c}{-} & - \\ \cmidrule(l){3-21} 
 & \multirow{2}{*}{BERT} & Exact-match & \multicolumn{1}{c}{0.68} & \multicolumn{1}{c}{0.78} & 0.79 & \multicolumn{1}{c}{0.59} & \multicolumn{1}{c}{0.62} & 0.61 & \multicolumn{1}{c}{0.68} & \multicolumn{1}{c}{0.79} & 0.79 & \multicolumn{1}{c}{0.66} & \multicolumn{1}{c}{0.74} & 0.74 & \multicolumn{1}{c}{0.69} & \multicolumn{1}{c}{0.79} & 0.79 & \multicolumn{1}{c}{0.68} & \multicolumn{1}{c}{0.72} & 0.65 \\ 
 &  & LLM-based & \multicolumn{1}{c}{0.61} & \multicolumn{1}{c}{0.70} & 0.67 & \multicolumn{1}{c}{0.56} & \multicolumn{1}{c}{0.62} & 0.60 & \multicolumn{1}{c}{0.61} & \multicolumn{1}{c}{0.71} & 0.68 & \multicolumn{1}{c}{0.60} & \multicolumn{1}{c}{0.68} & 0.65 & \multicolumn{1}{c}{0.61} & \multicolumn{1}{c}{0.71} & 0.68 & \multicolumn{1}{c}{\textbf{0.70}} & \multicolumn{1}{c}{0.78} & 0.76 \\ \midrule

\multirow{4}{*}{LL-70B-Gen} & \multirow{2}{*}{Statistical} & Exact-match & \multicolumn{1}{c}{0.51} & \multicolumn{1}{c}{0.59} & 0.61 & \multicolumn{1}{c}{0.45} & \multicolumn{1}{c}{0.56} & 0.53 & \multicolumn{1}{c}{0.45} & \multicolumn{1}{c}{0.57} & 0.56 & \multicolumn{1}{c}{0.53} & \multicolumn{1}{c}{0.55} & 0.56 & \multicolumn{1}{c}{0.47} & \multicolumn{1}{c}{0.58} & 0.55 & \multicolumn{1}{c}{-} & \multicolumn{1}{c}{-} & - \\
 &  & LLM-based & \multicolumn{1}{c}{0.52} & \multicolumn{1}{c}{0.61} & 0.49 & \multicolumn{1}{c}{0.53} & \multicolumn{1}{c}{0.54} & 0.38 & \multicolumn{1}{c}{0.53} & \multicolumn{1}{c}{0.60} & 0.43 & \multicolumn{1}{c}{0.62} & \multicolumn{1}{c}{0.58} & 0.47 & \multicolumn{1}{c}{0.54} & \multicolumn{1}{c}{0.60} & 0.44 & \multicolumn{1}{c}{-} & \multicolumn{1}{c}{-} & - \\ \cmidrule(l){3-21} 
 & \multirow{2}{*}{BERT} & Exact-match & \multicolumn{1}{c}{0.34} & \multicolumn{1}{c}{0.52} & 0.53 & \multicolumn{1}{c}{0.59} & \multicolumn{1}{c}{0.62} & 0.63 & \multicolumn{1}{c}{0.37} & \multicolumn{1}{c}{0.57} & 0.58 & \multicolumn{1}{c}{0.46} & \multicolumn{1}{c}{0.55} & 0.58 & \multicolumn{1}{c}{0.36} & \multicolumn{1}{c}{0.55} & 0.57 & \multicolumn{1}{c}{0.71} & \multicolumn{1}{c}{0.80} & 0.83 \\ 
 &  & LLM-based & \multicolumn{1}{c}{0.52} & \multicolumn{1}{c}{0.48} & 0.34 & \multicolumn{1}{c}{0.66} & \multicolumn{1}{c}{0.67} & 0.54 & \multicolumn{1}{c}{0.52} & \multicolumn{1}{c}{0.56} & 0.40 & \multicolumn{1}{c}{0.6} & \multicolumn{1}{c}{0.58} & 0.48 & \multicolumn{1}{c}{0.52} & \multicolumn{1}{c}{0.53} & 0.38 & \multicolumn{1}{c}{\textbf{0.72}} & \multicolumn{1}{c}{0.78} & 0.71 \\ \midrule

\multirow{4}{*}{ENSB-Gen} & \multirow{2}{*}{Statistical} & Exact-match & \multicolumn{1}{c}{0.58} & \multicolumn{1}{c}{0.60} & 0.76 & \multicolumn{1}{c}{0.67} & \multicolumn{1}{c}{0.75} & 0.84 & \multicolumn{1}{c}{0.68} & \multicolumn{1}{c}{0.76} & 0.85 & \multicolumn{1}{c}{0.62} & \multicolumn{1}{c}{0.66} & 0.79 & \multicolumn{1}{c}{0.62} & \multicolumn{1}{c}{0.66} & 0.79 & \multicolumn{1}{c}{-} & \multicolumn{1}{c}{-} & - \\ 
 &  & LLM-based & \multicolumn{1}{c}{0.57} & \multicolumn{1}{c}{0.63} & 0.69 & \multicolumn{1}{c}{0.72} & \multicolumn{1}{c}{0.81} & 0.82 & \multicolumn{1}{c}{0.73} & \multicolumn{1}{c}{0.83} & 0.85 & \multicolumn{1}{c}{0.63} & \multicolumn{1}{c}{0.69} & 0.74 & \multicolumn{1}{c}{0.73} & \multicolumn{1}{c}{0.83} & 0.84 & \multicolumn{1}{c}{-} & \multicolumn{1}{c}{-} & - \\ \cmidrule(l){3-21} 
 & \multirow{2}{*}{BERT} & Exact-match & \multicolumn{1}{c}{0.73} & \multicolumn{1}{c}{0.78} & 0.86 & \multicolumn{1}{c}{0.59} & \multicolumn{1}{c}{0.64} & 0.77 & \multicolumn{1}{c}{0.73} & \multicolumn{1}{c}{0.80} & 0.87 & \multicolumn{1}{c}{0.66} & \multicolumn{1}{c}{0.72} & 0.82 & \multicolumn{1}{c}{0.74} & \multicolumn{1}{c}{0.79} & 0.87 & \multicolumn{1}{c}{0.79} & \multicolumn{1}{c}{0.89} & 0.94 \\  
 &  & LLM-based & \multicolumn{1}{c}{0.76} & \multicolumn{1}{c}{0.84} & 0.85 & \multicolumn{1}{c}{0.58} & \multicolumn{1}{c}{0.66} & 0.71 & \multicolumn{1}{c}{0.76} & \multicolumn{1}{c}{0.86} & 0.87 & \multicolumn{1}{c}{0.61} & \multicolumn{1}{c}{0.74} & 0.77 & \multicolumn{1}{c}{0.76} & \multicolumn{1}{c}{0.85} & 0.86 & \multicolumn{1}{c}{\textbf{0.82}} & \multicolumn{1}{c}{0.90} & 0.92 \\ \midrule

\multirow{4}{*}{HA-Test} & \multirow{2}{*}{Statistical} & Exact-match & \multicolumn{1}{c}{0.56} & \multicolumn{1}{c}{0.64} & 0.71 & \multicolumn{1}{c}{0.71} & \multicolumn{1}{c}{0.80} & 0.82 & \multicolumn{1}{c}{\textbf{0.74}} & \multicolumn{1}{c}{0.82} & 0.83 & \multicolumn{1}{c}{0.63} & \multicolumn{1}{c}{0.69} & 0.74 & \multicolumn{1}{c}{0.74} & \multicolumn{1}{c}{0.82} & 0.83 & \multicolumn{1}{c}{-} & \multicolumn{1}{c}{-} & - \\  
 &  & LLM-based & \multicolumn{1}{c}{0.56} & \multicolumn{1}{c}{0.63} & 0.69 & \multicolumn{1}{c}{0.73} & \multicolumn{1}{c}{0.83} & 0.83 & \multicolumn{1}{c}{\textbf{0.74}} & \multicolumn{1}{c}{0.84} & 0.84 & \multicolumn{1}{c}{0.64} & \multicolumn{1}{c}{0.71} & 0.75 & \multicolumn{1}{c}{\textbf{0.74}} & \multicolumn{1}{c}{0.84} & 0.84 & \multicolumn{1}{c}{-} & \multicolumn{1}{c}{-} & - \\ \cmidrule(l){3-21} 
 & \multirow{2}{*}{BERT} & Exact-match & \multicolumn{1}{c}{0.28} & \multicolumn{1}{c}{0.50} & 0.55 & \multicolumn{1}{c}{0.40} & \multicolumn{1}{c}{0.50} & 0.55 & \multicolumn{1}{c}{0.40} & \multicolumn{1}{c}{0.50} & 0.55 & \multicolumn{1}{c}{0.40} & \multicolumn{1}{c}{0.50} & 0.55 & \multicolumn{1}{c}{0.40} & \multicolumn{1}{c}{0.50} & 0.55 & \multicolumn{1}{c}{0.73} & \multicolumn{1}{c}{0.82} & 0.86 \\ 
 &  & LLM-based & \multicolumn{1}{c}{0.32} & \multicolumn{1}{c}{0.50} & 0.52 & \multicolumn{1}{c}{0.35} & \multicolumn{1}{c}{0.50} & 0.52 & \multicolumn{1}{c}{0.32} & \multicolumn{1}{c}{0.50} & 0.52 & \multicolumn{1}{c}{0.35} & \multicolumn{1}{c}{0.50} & 0.52 & \multicolumn{1}{c}{0.35} & \multicolumn{1}{c}{0.50} & 0.52 & \multicolumn{1}{c}{0.64} & \multicolumn{1}{c}{0.82} & 0.83 \\ \bottomrule
\end{tabular}
}
\caption{Hallucination classifier results on various test sets created using the Jeopardy dataset samples, \textbf{AUC:} Area Under Curve, \textbf{B-Acc:} Balanced Accuracy. The best result highlighted in \textbf{bold}.}
\label{tab:jeopardy_results}
\end{table*}

%% file: appendix_tables/results_token_counts_experiment.tex
\begin{table*}[h!]
\centering
\resizebox{\columnwidth}{!}{%
\begin{tabular}{c|c|c|c|c}
\hline
\textbf{Classifier}       & \textbf{Feature Combination} & \textbf{F1} & \textbf{AUC} & \textbf{B-ACC} \\ \hline
\multirow{2}{*}{Jeopardy} & EC-EC+TokenCounts             & 0.74        & 0.84         & 0.84           \\ \cline{2-5} 
                          & QR-RR+TokenCounts             & 0.74        & 0.85         & 0.85           \\ \hline
\multirow{2}{*}{Kaggle}   & EC-EC+TokenCounts             & 0.79        & 0.73         & 0.84           \\ \cline{2-5} 
                          & QR-RR+TokenCounts             & 0.80        & 0.74         & 0.83           \\ \hline
\end{tabular}
}
\caption{Classifier results with combination of NLI features, TokenCounts (total tokens, and special tokens count)}
\label{tab:results_token_counts}
\end{table*}

%% file: table_jeopardy_category_wise.tex
\begin{table*}
\centering
\resizebox{\textwidth}{!}{%
\begin{tabular}{@{}cl|cccc|cccc|cccc|cccc @{}}
\toprule
\multirow{2}{*}{\textbf{Test set}} & \multirow{2}{*}{\textbf{Sub-category}} & \multicolumn{4}{c}{\textbf{(Statistical, Exact-match)}} & \multicolumn{4}{c}{\textbf{(Statistical, LLM-based)}} & \multicolumn{4}{c}{\textbf{(BERT, Exact-match)}} & \multicolumn{4}{c}{\textbf{(BERT, LLM-based)}} \\ \cmidrule{3-18}
 &  & \multicolumn{1}{l}{\textbf{ACC}} & \multicolumn{1}{l}{\textbf{F1}} & \multicolumn{1}{l}{\textbf{AUC}} & \multicolumn{1}{l}{\textbf{B-ACC}} & \multicolumn{1}{l}{\textbf{ACC}} & \multicolumn{1}{l}{\textbf{F1}} & \multicolumn{1}{l}{\textbf{AUC}} & \multicolumn{1}{l}{\textbf{B-ACC}} & \multicolumn{1}{l}{\textbf{ACC}} & \multicolumn{1}{l}{\textbf{F1}} & \multicolumn{1}{l}{\textbf{AUC}} & \multicolumn{1}{l}{\textbf{B-ACC}} & \multicolumn{1}{l}{\textbf{ACC}} & \multicolumn{1}{l}{\textbf{F1}} & \multicolumn{1}{l}{\textbf{AUC}} & \multicolumn{1}{l}{\textbf{B-ACC}} \\ \midrule
 & Arts and humanity & 0.77 & 0.80 & 0.62 & 0.93 & 0.76 & 0.78 & 0.74 & 0.94 & 0.90 & \textbf{0.90} & 0.89 & 0.99 & 0.86 & 0.85 & 0.85 & 0.97 \\
 & Geography and travel & 0.70 & 0.70 & 0.65 & 0.85 & 0.71 & 0.71 & 0.76 & 0.90 & 0.83 & \textbf{0.84} & 0.93 & 0.98 & 0.78 & 0.79 & 0.84 & 0.94 \\
 & Language and communication & 0.70 & 0.76 & 0.61 & 0.94 & 0.70 & 0.74 & 0.70 & 0.94 & 0.90 & \textbf{0.90} & 0.87 & 0.98 & 0.86 & 0.85 & 0.82 & 0.97 \\
 & Sciences & 0.74 & 0.76 & 0.65 & 0.92 & 0.72 & 0.74 & 0.75 & 0.93 & 0.84 & \textbf{0.86} & 0.89 & 0.98 & 0.79 & 0.79 & 0.83 & 0.95 \\
 & Social sciences & 0.75 & 0.77 & 0.63 & 0.92 & 0.75 & 0.76 & 0.76 & 0.92 & 0.87 & \textbf{0.88} & 0.92 & 0.99 & 0.82 & 0.82 & 0.86 & 0.96 \\
\multirow{-6}{*}{\textbf{TL-1.1B-Gen}} & Sports and recreation & 0.77 & 0.79 & 0.64 & 0.92 & 0.78 & 0.79 & 0.77 & 0.94 & 0.90& \textbf{0.90} & 0.92 & 0.99 & 0.85 & 0.84 & 0.84 & 0.96 \\  \midrule
 & Arts and humanity & 0.69 & 0.70& 0.73 & 0.89 & 0.69 & 0.70& 0.78 & 0.86 & 0.80& \textbf{0.81} & 0.88 & 0.96 & 0.80& 0.80& 0.88 & 0.92 \\
 & Geography and travel & 0.63 & 0.63 & 0.69 & 0.66 & 0.75 & 0.74 & 0.81 & 0.64 & 0.75 & 0.75 & 0.86 & 0.86 & 0.82 & \textbf{0.82} & 0.87 & 0.76 \\
 & Language and communication & 0.61 & 0.62 & 0.69 & 0.81 & 0.64 & 0.63 & 0.75 & 0.76 & 0.73 & \textbf{0.74} & 0.83 & 0.91 & 0.73 & 0.73 & 0.82 & 0.83 \\
 & Sciences & 0.64 & 0.64 & 0.71 & 0.73 & 0.74 & 0.74 & 0.82 & 0.69 & 0.72 & 0.72 & 0.81 & 0.81 & 0.84 & \textbf{0.83} & 0.89 & 0.81 \\
 & Social sciences & 0.64 & 0.64 & 0.71 & 0.77 & 0.75 & 0.75 & 0.81 & 0.74 & 0.75 & 0.75 & 0.85 & 0.89 & 0.81 & \textbf{0.80} & 0.89 & 0.83 \\
\multirow{-6}{*}{\textbf{PHI-3.5B-Gen}} & Sports and recreation & 0.68 & 0.69 & 0.74 & 0.87 & 0.68 & 0.68 & 0.75 & 0.82 & 0.81 & \textbf{0.81} & 0.91 & 0.96 & 0.75 & 0.75 & 0.83 & 0.87 \\  \midrule
 & Arts and humanity & 0.71 & 0.71 & 0.78 & 0.82 & 0.81 & 0.81 & 0.88 & 0.84 & 0.75 & 0.75 & 0.82 & 0.84 & 0.84 & \textbf{0.84} & 0.90& 0.88 \\
 & Geography and travel & 0.75 & 0.73 & 0.75 & 0.64 & 0.89 & 0.88 & 0.92 & 0.72 & 0.80& 0.78 & 0.86 & 0.80& 0.91 & \textbf{0.90} & 0.92 & 0.78 \\
 & Language and communication & 0.66 & 0.66 & 0.76 & 0.82 & 0.74 & 0.73 & 0.84 & 0.81 & 0.71 & 0.72 & 0.82 & 0.86 & 0.77 & \textbf{0.76} & 0.85 & 0.84 \\
 & Sciences & 0.69 & 0.68 & 0.74 & 0.73 & 0.83 & 0.83 & 0.89 & 0.77 & 0.70& 0.69 & 0.79 & 0.77 & 0.85 & \textbf{0.85} & 0.89 & 0.81 \\
 & Social sciences & 0.69 & 0.67 & 0.74 & 0.74 & 0.85 & 0.85 & 0.91 & 0.79 & 0.71 & 0.70& 0.81 & 0.80& 0.88 & \textbf{0.87} & 0.92 & 0.83 \\
\multirow{-6}{*}{\textbf{LL-8B-Gen}} & Sports and recreation & 0.71 & 0.70& 0.76 & 0.81 & 0.80& 0.80& 0.88 & 0.81 & 0.74 & 0.74 & 0.84 & 0.86 & 0.83 & \textbf{0.83} & 0.90& 0.86 \\  \midrule
 & Arts and humanity & 0.68 & 0.69 & 0.68 & 0.88 & 0.69 & 0.70& 0.74 & 0.86 & 0.83 & 0.84 & 0.90& 0.97 & 0.82 & \textbf{0.82} & 0.90& 0.95 \\
 & Geography and travel & 0.63 & 0.62 & 0.68 & 0.70& 0.71 & 0.71 & 0.79 & 0.77 & 0.79 & 0.79 & 0.91 & 0.94 & 0.80& \textbf{0.80} & 0.89 & 0.88 \\
 & Language and communication & 0.61 & 0.64 & 0.65 & 0.85 & 0.63 & 0.65 & 0.72 & 0.83 & 0.79 & \textbf{0.80} & 0.87 & 0.95 & 0.79 & 0.79 & 0.87 & 0.93 \\
 & Sciences & 0.65 & 0.65 & 0.71 & 0.80& 0.69 & 0.69 & 0.78 & 0.81 & 0.79 & 0.79 & 0.89 & 0.94 & 0.82 & \textbf{0.82} & 0.90& 0.92 \\
 & Social sciences & 0.64 & 0.64 & 0.67 & 0.80& 0.69 & 0.69 & 0.77 & 0.81 & 0.79 & 0.80& 0.89 & 0.95 & 0.82 & \textbf{0.82} & 0.91 & 0.93 \\
\multirow{-6}{*}{\textbf{MST-7B-Gen}} & Sports and recreation & 0.66 & 0.68 & 0.70& 0.87 & 0.69 & 0.69 & 0.77 & 0.86 & 0.82 & 0.82 & 0.90& 0.97 & 0.81 & \textbf{0.81} & 0.89 & 0.94 \\  \midrule
 & Arts and humanity & 0.71 & 0.71 & 0.78 & 0.82 & 0.60& 0.60& 0.70& 0.79 & 0.74 & \textbf{0.74} & 0.81 & 0.84 & 0.69 & 0.70& 0.77 & 0.86 \\
 & Geography and travel & 0.75 & \textbf{0.73} & 0.75 & 0.64 & 0.70& 0.66 & 0.68 & 0.55 & 0.74 & 0.71 & 0.76 & 0.67 & 0.73 & 0.70& 0.76 & 0.66 \\
 & Language and communication & 0.66 & 0.66 & 0.76 & 0.82 & 0.59 & 0.59 & 0.69 & 0.76 & 0.70& \textbf{0.71} & 0.73 & 0.74 & 0.66 & 0.67 & 0.76 & 0.83 \\
 & Sciences & 0.69 & 0.68 & 0.74 & 0.73 & 0.68 & 0.66 & 0.71 & 0.57 & 0.70& 0.68 & 0.76 & 0.77 & 0.72 & \textbf{0.71} & 0.79 & 0.70\\
 & Social sciences & 0.69 & 0.67 & 0.74 & 0.74 & 0.68 & 0.65 & 0.69 & 0.64 & 0.69 & 0.66 & 0.76 & 0.76 & 0.73 & \textbf{0.72} & 0.79 & 0.75 \\
\multirow{-6}{*}{\textbf{GM-7B-Gen}} & Sports and recreation & 0.70& \textbf{0.70} & 0.77 & 0.82 & 0.63 & 0.62 & 0.70& 0.69 & 0.70& 0.69 & 0.79 & 0.83 & 0.68 & 0.68 & 0.78 & 0.78 \\  \midrule
 & Arts and humanity & 0.53 & 0.52 & 0.56 & 0.67 & 0.57 & 0.54 & 0.59 & 0.56 & 0.71 & \textbf{0.70} & 0.81 & 0.88 & 0.70& 0.68 & 0.81 & 0.83 \\
 & Geography and travel & 0.64 & 0.57 & 0.59 & 0.53 & 0.78 & 0.69 & 0.68 & 0.36 & 0.76 & 0.75 & 0.85 & 0.83 & 0.80& \textbf{0.76} & 0.72 & 0.54 \\
 & Language and communication & 0.52 & 0.51 & 0.58 & 0.65 & 0.58 & 0.55 & 0.51 & 0.49 & 0.68 & \textbf{0.69} & 0.75 & 0.82 & 0.69 & 0.67 & 0.72 & 0.68 \\
 & Sciences & 0.59 & 0.53 & 0.57 & 0.51 & 0.67 & 0.64 & 0.54 & 0.38 & 0.74 & 0.73 & 0.80& 0.81 & 0.79 & \textbf{0.75} & 0.80& 0.69 \\
 & Social sciences & 0.55 & 0.54 & 0.55 & 0.51 & 0.67 & 0.65 & 0.60& 0.39 & 0.73 & 0.71 & 0.81 & 0.81 & 0.81 & \textbf{0.77} & 0.85 & 0.76 \\
\multirow{-6}{*}{\textbf{LL-70B-Gen}} & Sports and recreation & 0.54 & 0.54 & 0.61 & 0.65 & 0.69 & 0.67 & 0.62 & 0.62 & 0.70& \textbf{0.69} & 0.80& 0.83 & 0.73 & \textbf{0.69} & 0.78 & 0.77 \\  \midrule
 & Arts and humanity & 0.71 & 0.72 & 0.78 & 0.90& 0.74 & 0.75 & 0.83 & 0.89 & 0.83 & \textbf{0.84} & 0.90& 0.96 & 0.83 & 0.83 & 0.90& 0.94 \\
 & Geography and travel & 0.67 & 0.67 & 0.73 & 0.74 & 0.77 & 0.77 & 0.86 & 0.79 & 0.79 & 0.78 & 0.91 & 0.92 & 0.83 & \textbf{0.82} & 0.92 & 0.88 \\
 & Language and communication & 0.63 & 0.65 & 0.73 & 0.87 & 0.66 & 0.66 & 0.78 & 0.85 & 0.78 & \textbf{0.79} & 0.86 & 0.94 & 0.77 & 0.77 & 0.86 & 0.91 \\
 & Sciences & 0.69 & 0.70& 0.77 & 0.83 & 0.75 & 0.75 & 0.85 & 0.83 & 0.76 & 0.76 & 0.87 & 0.92 & 0.83 & \textbf{0.83} & 0.92 & 0.91 \\
 & Social sciences & 0.67 & 0.67 & 0.75 & 0.84 & 0.76 & 0.76 & 0.85 & 0.84 & 0.78 & 0.78 & 0.89 & 0.94 & 0.84 & \textbf{0.84} & 0.93 & 0.93 \\
\multirow{-6}{*}{\textbf{ENSB-Gen}} & Sports and recreation & 0.70& 0.71 & 0.78 & 0.90& 0.72 & 0.73 & 0.82 & 0.87 & 0.80& 0.81 & 0.90& 0.96 & 0.81 & \textbf{0.81} & 0.88 & 0.93 \\  \midrule
 & Arts and humanity & 0.76 & \textbf{0.76} & 0.82 & 0.88 & 0.74 & 0.75 & 0.82 & 0.88 & 0.68 & 0.68 & 0.79 & 0.90& 0.62 & 0.61 & 0.81 & 0.90\\
 & Geography and travel & 0.76 & \textbf{0.76} & 0.84 & 0.76 & 0.77 & \textbf{0.76} & 0.84 & 0.75 & 0.77 & \textbf{0.76} & 0.84 & 0.82 & 0.73 & 0.70& 0.85 & 0.82 \\
 & Language and communication & 0.69 & 0.70& 0.79 & 0.85 & 0.69 & 0.70& 0.81 & 0.87 & 0.72 & \textbf{0.73} & 0.80& 0.87 & 0.65 & 0.65 & 0.76 & 0.85 \\
 & Sciences & 0.78 & \textbf{0.78} & 0.86 & 0.84 & 0.76 & 0.76 & 0.87 & 0.84 & 0.72 & 0.72 & 0.81 & 0.83 & 0.67 & 0.64 & 0.82 & 0.83 \\
 & Social sciences & 0.73 & 0.73 & 0.82 & 0.79 & 0.77 & \textbf{0.77} & 0.86 & 0.84 & 0.76 & 0.75 & 0.83 & 0.85 & 0.71 & 0.69 & 0.87 & 0.89 \\
\multirow{-6}{*}{\textbf{HA-Test}} & Sports and recreation & 0.73 & \textbf{0.74} & 0.83 & 0.86 & 0.73 & 0.73 & 0.82 & 0.85 & 0.72 & 0.71 & 0.83 & 0.87 & 0.63 & 0.62 & 0.76 & 0.83 \\ \bottomrule
\end{tabular}}
\caption{Category wise results on Jeopardy test sets; \textbf{(Statistical, Exact-match)} - Statistical classifier trained on Exact-match based labels, \textbf{(Statistical, LLM-based)} - Statistical classifier trained on LLM-based labels, \textbf{(BERT, Exact-match)} - BERT classifier trained on Exact-match based labels, \textbf{(BERT, LLM-based)} - BERT classifier trained on LLM-based labels; we report the best classifier combination results for each LLM. The best result highlighted in \textbf{bold}.}
\label{tab:jeopardy_category_wise}
\end{table*}

%% file: table_kaggle_datasetwise.tex
\begin{table*}
\centering\footnotesize
\resizebox{\textwidth}{!}{%
\begin{tabular}{@{}cl|cccc|cccc|cccc|cccc@{}}
\toprule
\multirow{2}{*}{\textbf{Test set}} & \multirow{2}{*}{\textbf{Sub-category}} & \multicolumn{4}{c}{\textbf{(Statistical, Exact-match)}} & \multicolumn{4}{c}{\textbf{(Statistical, LLM-based)}} & \multicolumn{4}{c}{\textbf{(BERT, Exact-match)}} & \multicolumn{4}{c}{\textbf{(BERT, LLM-based)}} \\ \cmidrule{3-18}
 &  & \multicolumn{1}{l}{\textbf{ACC}} & \multicolumn{1}{l}{\textbf{F1}} & \multicolumn{1}{l}{\textbf{AUC}} & \multicolumn{1}{l}{\textbf{B-ACC}} & \multicolumn{1}{l}{\textbf{ACC}} & \multicolumn{1}{l}{\textbf{F1}} & \multicolumn{1}{l}{\textbf{AUC}} & \multicolumn{1}{l}{\textbf{B-ACC}} & \multicolumn{1}{l}{\textbf{ACC}} & \multicolumn{1}{l}{\textbf{F1}} & \multicolumn{1}{l}{\textbf{AUC}} & \multicolumn{1}{l}{\textbf{B-ACC}} & \multicolumn{1}{l}{\textbf{ACC}} & \multicolumn{1}{l}{\textbf{F1}} & \multicolumn{1}{l}{\textbf{AUC}} & \multicolumn{1}{l}{\textbf{B-ACC}} \\ \midrule
\multirow{4}{*}{\textbf{TL-1.1B-Gen}} & GK & 0.60& 0.61 & 0.71 & 0.83 & 0.78 & 0.77 & 0.82 & 0.87 & 0.75 & 0.73 & 0.74 & 0.83 & 0.79 & \textbf{0.77} & 0.86 & 0.90 \\
 & MathQA & 0.73 & 0.80 & 0.53 & 0.95 & 0.92 & 0.95 & 0.66 & 1\phantom{0}\phantom{0} & 0.88 & 0.90 & 0.83 & 0.99 & 0.99 & \textbf{0.99} & 0.74 & 1\phantom{0}\phantom{0} \\
 & MathQSA & 0.71 & 0.76 & 0.56 & 0.93 & 0.88 & 0.92 & 0.55 & 0.99 & 0.79 & 0.82 & 0.54 & 0.93 & 0.97 & \textbf{0.97} & 0.69 & 0.99 \\
 & SciQ & 0.59 & 0.63 & 0.62 & 0.86 & 0.71 & 0.70& 0.68 & 0.84 & 0.73 & \textbf{0.74} & 0.65 & 0.87 & 0.71 & 0.72 & 0.73 & 0.87 \\ \midrule
\multirow{4}{*}{\textbf{PHI-3.5B-Gen}} & GK & 0.71 & 0.70& 0.69 & 0.46 & 0.75 & 0.76 & 0.64 & 0.34 & 0.74 & 0.70& 0.66 & 0.50& 0.85 & \textbf{0.82} & 0.64 & 0.37 \\
 & MathQA & 0.63 & 0.73 & 0.53 & 0.95 & 0.65 & 0.67 & 0.56 & 0.83 & 0.89 & \textbf{0.91}& 0.89 & 0.99 & 0.82 & 0.80& 0.75 & 0.91 \\
 & MathQSA & 0.61 & 0.69 & 0.50& 0.91 & 0.63 & 0.65 & 0.64 & 0.82 & 0.73 & \textbf{0.78} & 0.70& 0.96 & 0.74 & 0.73 & 0.72 & 0.86 \\
 & SciQ & 0.57 & 0.57 & 0.58 & 0.48 & 0.69 & 0.71 & 0.67 & 0.38 & 0.61 & 0.59 & 0.64 & 0.54 & 0.75 & \textbf{0.75} & 0.68 & 0.39 \\ \midrule
\multirow{4}{*}{\textbf{LL-8B-Gen}} & GK & 0.73 & 0.71 & 0.66 & 0.48 & 0.82 & \textbf{0.82} & 0.71 & 0.38 & 0.73 & 0.68 & 0.70& 0.49 & 0.82 & \textbf{0.82} & 0.68 & 0.38 \\
 & MathQA & 0.67 & 0.73 & 0.62 & 0.93 & 0.77 & 0.76 & 0.67 & 0.88 & 0.78 & \textbf{0.82} & 0.81 & 0.97 & 0.82 & 0.81 & 0.78 & 0.92 \\
 & MathQSA & 0.62 & 0.64 & 0.66 & 0.85 & 0.71 & 0.70& 0.69 & 0.82 & 0.73 & 0.74 & 0.70& 0.86 & 0.76 & \textbf{0.76} & 0.77 & 0.86 \\
 & SciQ & 0.62 & 0.61 & 0.70& 0.65 & 0.73 & 0.72 & 0.72 & 0.51 & 0.62 & 0.58 & 0.74 & 0.70& 0.77 & \textbf{0.75} & 0.76 & 0.55 \\ \midrule
\multirow{4}{*}{\textbf{MST-7B-Gen}} & GK & 0.46 & 0.46 & 0.41 & 0.37 & 0.47 & 0.49 & 0.46 & 0.32 & 0.62 & \textbf{0.61} & 0.62 & 0.56 & 0.35 & 0.21 & 0.66 & 0.52 \\
 & MathQA & 0.93 & \textbf{0.91} & 0.52 & 0.95 & 0.93 & 0.90 & 0.55 & 0.94 & 0.90 & \textbf{0.91} & 0.86 & 0.99 & 0.93 & 0.90 & 0.76 & 0.98 \\
 & MathQSA & 0.89 & 0.86 & 0.53 & 0.91 & 0.91 & 0.87 & 0.55 & 0.93 & 0.82 & 0.84 & 0.75 & 0.96 & 0.91 & \textbf{0.87} & 0.68 & 0.95 \\
 & SciQ & 0.50& 0.50& 0.49 & 0.55 & 0.51 & 0.51 & 0.49 & 0.37 & 0.62 & \textbf{0.62} & 0.70& 0.74 & 0.39 & 0.25 & 0.62 & 0.50\\ \midrule
\multirow{4}{*}{\textbf{GM-7B-Gen}} & GK & 0.67 & 0.63 & 0.66 & 0.61 & 0.72 & \textbf{0.68} & 0.65 & 0.52 & 0.65 & 0.59 & 0.68 & 0.66 & 0.72 & 0.66 & 0.65 & 0.54 \\
 & MathQA & 0.60& 0.68 & 0.48 & 0.90 & 0.66 & 0.72 & 0.58 & 0.92 & 0.70& 0.76 & 0.72 & 0.96 & 0.73 & \textbf{0.77} & 0.51 & 0.89 \\
 & MathQSA & 0.60& 0.64 & 0.51 & 0.84 & 0.66 & 0.68 & 0.52 & 0.83 & 0.59 & 0.64 & 0.67 & 0.89 & 0.74 & \textbf{0.73} & 0.54 & 0.83 \\
 & SciQ & 0.58 & 0.56 & 0.65 & 0.67 & 0.68 & 0.66 & 0.66 & 0.51 & 0.50& 0.49 & 0.50& 0.57 & 0.71 & \textbf{0.68} & 0.68 & 0.53 \\ \midrule
\multirow{4}{*}{\textbf{LL-70B-Gen}} & GK & 0.72 & 0.72 & 0.70& 0.51 & 0.88 & \textbf{0.88} & 0.84 & 0.57 & 0.73 & 0.68 & 0.76 & 0.58 & 0.89 & \textbf{0.88} & 0.83 & 0.56 \\
 & MathQA & 0.63 & 0.70& 0.67 & 0.93 & 0.82 & 0.81 & 0.74 & 0.91 & 0.84 & \textbf{0.85} & 0.73 & 0.94 & 0.86 & \textbf{0.85} & 0.78 & 0.92 \\
 & MathQSA & 0.62 & 0.65 & 0.71 & 0.87 & 0.78 & 0.76 & 0.79 & 0.88 & 0.78 & 0.78 & 0.77 & 0.89 & 0.83 & \textbf{0.82} & 0.84 & 0.90 \\
 & SciQ & 0.63 & 0.62 & 0.68 & 0.63 & 0.74 & 0.75 & 0.76 & 0.51 & 0.65 & 0.61 & 0.75 & 0.71 & 0.80& \textbf{0.79} & 0.77 & 0.56 \\ \midrule
\multirow{4}{*}{\textbf{ENSB-Gen}} & GK & 0.65 & 0.63 & 0.73 & 0.69 & 0.77 & 0.76 & 0.76 & 0.69 & 0.73 & 0.71 & 0.77 & 0.74 & 0.79 & \textbf{0.79} & 0.78 & 0.71 \\
 & MathQA & 0.68 & 0.76 & 0.57 & 0.95 & 0.78 & 0.80& 0.70& 0.94 & 0.85 & \textbf{0.88} & 0.85 & 0.99 & 0.89 & \textbf{0.88} & 0.84 & 0.97 \\
 & MathQSA & 0.63 & 0.70& 0.55 & 0.91 & 0.77 & 0.79 & 0.68 & 0.92 & 0.70& 0.75 & 0.72 & 0.95 & 0.85 & \textbf{0.85} & 0.81 & 0.96 \\
 & SciQ & 0.58 & 0.58 & 0.68 & 0.71 & 0.70& 0.70& 0.76 & 0.67 & 0.66 & 0.65 & 0.74 & 0.75 & 0.74 & \textbf{0.74} & 0.80& 0.68 \\ \midrule
\multirow{4}{*}{\textbf{HA-Test}} & GK & 0.70& 0.70& 0.75 & 0.61 & 0.77 & \textbf{0.77} & 0.78 & 0.65 & 0.69 & 0.59 & 0.74 & 0.60& 0.71 & 0.63 & 0.71 & 0.55 \\
 & MathQA & 0.71 & 0.80& 0.66 & 0.98 & 0.80& 0.87 & 0.67 & 0.98 & 0.97 & \textbf{0.95} & 0.50& 0.97 & 0.97 & \textbf{0.95} & 0.50& 0.97 \\
 & MathQSA & 0.65 & 0.76 & 0.63 & 0.97 & 0.79 & 0.84 & 0.61 & 0.97 & 0.96 & \textbf{0.93} & 0.50& 0.96 & 0.96 & \textbf{0.93} & 0.50& 0.96 \\
 & SciQ & 0.62 & 0.61 & 0.70& 0.65 & \textbf{0.65} & 0.65 & 0.71 & 0.67 & 0.56 & 0.43 & 0.74 & 0.72 & 0.56 & 0.44 & 0.71 & 0.70\\ \bottomrule
\end{tabular}}
\caption{Dataset-wise results on Kaggle test sets; \textbf{(Statistical, Exact-match)} - Statistical classifier trained on Exact-match based labels, \textbf{(Statistical, LLM-based)} - Statistical classifier trained on LLM-based labels, \textbf{(BERT, Exact-match)} - BERT classifier trained on Exact-match based labels, \textbf{(BERT, LLM-based)} - BERT classifier trained on LLM-based labels; we report the best classifier combination results for each LLM. The best result highlighted in \textbf{bold}.}
\label{tab:kaggle_datasetwise}
\end{table*}

%% file: table_generate_responses.tex
\begin{table*}
\centering
\resizebox{\textwidth}{!}{%

\begin{tabular}{c|l}
\toprule
\textbf{Role} & \textbf{Content} \\
\midrule
System & You are a helpful AI assistant. Provide the answer to the question, do not provide any extra information. \\
\midrule
User & \texttt{\{question\}} \\
\bottomrule
\end{tabular}}
\caption{Prompt for response generation to a query, we used the same prompt for all the different LLMs inference}
\label{tab:responses_prompt}
\end{table*}

%% file: table_all_results_analysis.tex
\begin{table*}
\centering\small
\resizebox{\textwidth}{!}{%
\begin{tabular}{@{}l|lll|lll@{}}
\toprule
 & \multicolumn{3}{c|}{\textbf{Jeopardy}} & \multicolumn{3}{c}{\textbf{Kaggle}} \\ 
\multicolumn{1}{l|}{\textbf{Test set}} & \textbf{Labeling strategy} & \textbf{Feature combination} & \textbf{Classifier} & \textbf{Labeling strategy} & \textbf{Feature combination} & \textbf{Classifier} \\ \cmidrule{1-7}
\textbf{TL-1.1B-Gen} & Exact-match & q-r+(Q-R)+(R-R) & BERT & LLM-based & q-r+(Q-R)+(R-R) & BERT \\
\textbf{PHI-3.5B-Gen} & Exact-match & q-r+(Q-R)+(R-R) & BERT & LLM-based & q-r+(Q-R)+(R-R) & BERT \\
\textbf{LL-8B-Gen} & LLM-based & q-r+(Q-R)+(R-R) & BERT & Exact-match & q-r+(Q-R)+(R-R) & BERT \\
\textbf{MST-7B-Gen} & LLM-based & q-r+(Q-R)+(R-R) & BERT & Exact-match & q-r+(Q-R)+(R-R) & BERT \\
\textbf{GM-7B-Gen} & LLM-based & q-r+(Q-R)+(R-R) & BERT & LLM-based & (Q-R) + (R-R) & BERT \\
\textbf{LL-70B-Gen} & LLM-based & q-r+(Q-R)+(R-R) & BERT & LLM-based & q-r+(Q-R)+(R-R) & BERT \\
\textbf{ENSB-Gen} & LLM-based & q-r+(Q-R)+(R-R) & BERT & LLM-based & q-r+(Q-R)+(R-R) & BERT \\
\textbf{HA-Test} & Human-annotated & EC-EC & Statistical & Human-annotated & EC-EC & Statistical \\ \bottomrule
\end{tabular}%
}
\caption{Best feature combination for each test set, including the associated classifier and labeling strategy.}
\label{tab:all_results_details}
\vspace{-5mm}
\end{table*}

%% file: appendix_tables/table_string_jeopardy_ML_appendix.tex
\begin{table*}
\centering\footnotesize
\resizebox{\textwidth}{!}{%
\begin{tabular}{@{}ll|ccc|ccc|ccc|ccc|ccc @{}} \\ \toprule
\multirow{2}{*}{\textbf{Test set}} & \multirow{2}{*}{\textbf{Sub-category}} & \multicolumn{3}{c}{{\textbf{QR}}} & \multicolumn{3}{c}{{\textbf{RR}}} & \multicolumn{3}{c}{{\textbf{EC-EC}}} & \multicolumn{3}{c}{{\textbf{C-C}}} & \multicolumn{3}{c}{{\textbf{QR+RR}}} \\ \cmidrule{3-17}
 &  & { \textbf{F1}} & { \textbf{AUC}} & { \textbf{B-ACC}} & { \textbf{F1}} & { \textbf{AUC}} & { \textbf{B-ACC}} & { \textbf{F1}} & { \textbf{AUC}} & { \textbf{B-ACC}} & { \textbf{F1}} & { \textbf{AUC}} & { \textbf{B-ACC}} & { \textbf{F1}} & { \textbf{AUC}} & { \textbf{B-ACC}} \\ \midrule
 & {Arts and humanity} & { 0.71} & { 0.59} & { 0.93} & { 0.78} & { 0.57} & { 0.92} & { 0.79} & { 0.63} & { 0.94} & { 0.71} & { 0.58} & { 0.93} & { \textbf{0.80}} & { 0.62} & { 0.93} \\
 & {Geography and travel} & { 0.62} & 0.60 & { 0.83} & { 0.65} & { 0.57} & { 0.82} & \textbf{0.70} & { 0.66} & { 0.86} & { 0.64} & 0.60 & { 0.84} & \textbf{0.70} & { 0.65} & { 0.85} \\
 & {Language and communication} & { 0.71} & { 0.58} & { 0.93} & { 0.75} & { 0.56} & { 0.92} & { \textbf{0.76}} & 0.60 & { 0.94} & 0.70 & 0.60 & { 0.94} & { \textbf{0.76}} & { 0.61} & { 0.94} \\
 & {Sciences} & { 0.66} & { 0.58} & 0.90 & { 0.72} & 0.60 & 0.90 & { 0.74} & { 0.65} & { 0.92} & { 0.68} & 0.60 & 0.90 & { \textbf{0.76}} & { 0.65} & { 0.92} \\
 & {Social Sciences} & 0.70 & { 0.59} & { 0.91} & { 0.76} & { 0.58} & 0.90 & { \textbf{0.77}} & { 0.65} & { 0.92} & { 0.69} & { 0.59} & { 0.91} & { \textbf{0.77}} & { 0.63} & { 0.92} \\
\multirow{-6}{*}{{ \textbf{TL-1.1B-Gen}}} & {Sports and recreation} & 0.70 & { 0.57} & { 0.91} & { 0.76} & { 0.57} & { 0.91} & { 0.78} & { 0.64} & { 0.92} & 0.70 & { 0.59} & { 0.92} & { \textbf{0.79}} & { 0.64} & { 0.92} \\ \cmidrule{2-17}
 & { \textbf{Average}} & { 0.68} & { 0.59} & 0.90 & { 0.74} & { 0.58} & 0.90 & { 0.76} & { 0.64} & { 0.92} & { 0.69} & { 0.59} & { 0.91} & { \textbf{0.76}} & { 0.63} & { 0.91} \\ \midrule
 & {Arts and humanity} & { 0.61} & { 0.59} & { 0.82} & { 0.66} & { 0.69} & { 0.87} & \textbf{0.70} & { 0.73} & { 0.89} & { 0.64} & { 0.64} & { 0.85} & \textbf{0.70} & { 0.73} & { 0.89} \\
 & {Geography and travel} & { 0.51} & { 0.57} & { 0.54} & { \textbf{0.63}} & { 0.69} & { 0.66} & { \textbf{0.63}} & { 0.68} & { 0.66} & { 0.54} & { 0.58} & { 0.57} & { 0.62} & { 0.69} & { 0.66} \\
 & {Language and communication} & { 0.56} & { 0.57} & { 0.74} & { 0.59} & { 0.66} & 0.80 & { \textbf{0.62}} & { 0.69} & { 0.81} & { 0.55} & { 0.58} & { 0.75} & 0.60 & { 0.67} & 0.80 \\
 & {Sciences} & { 0.51} & { 0.58} & { 0.59} & { \textbf{0.64}} & { 0.71} & 0.70 & { \textbf{0.64}} & { 0.71} & { 0.71} & { 0.57} & { 0.64} & { 0.63} & { \textbf{0.64}} & { 0.71} & { 0.73} \\
 & {Social sciences} & { 0.51} & { 0.56} & { 0.65} & { 0.62} & 0.70 & { 0.75} & { \textbf{0.64}} & { 0.71} & { 0.77} & { 0.56} & { 0.61} & { 0.69} & { 0.63} & { 0.71} & { 0.77} \\
\multirow{-6}{*}{{ \textbf{PHI-3.5B-Gen}}} & {Sports and recreation} & 0.60 & { 0.64} & { 0.83} & { 0.66} & 0.70 & { 0.85} & { 0.69} & { 0.74} & { 0.87} & 0.60 & { 0.63} & { 0.82} & \textbf{0.70} & { 0.76} & { 0.88} \\ \cmidrule{2-17}
 & { \textbf{Average}} & { 0.55} & { 0.59} & 0.70 & { 0.63} & { 0.69} & { 0.77} & { \textbf{0.65}} & { 0.71} & { 0.79} & { 0.58} & { 0.61} & { 0.72} & { \textbf{0.65}} & { 0.71} & { 0.79} \\ \midrule
 & {Arts and humanity} & { 0.52} & { 0.57} & { 0.59} & { 0.69} & { 0.78} & { 0.81} & { \textbf{0.71}} & { 0.78} & { 0.82} & { 0.66} & { 0.72} & { 0.76} & { \textbf{0.71}} & { 0.78} & { 0.82} \\
 & {Geography and travel} & { 0.48} & { 0.56} & { 0.41} & { 0.72} & { 0.75} & { 0.64} & { \textbf{0.73}} & { 0.75} & { 0.64} & { 0.69} & { 0.68} & { 0.57} & { 0.69} & { 0.73} & { 0.61} \\
 & {Language and communication} & { 0.52} & { 0.58} & { 0.68} & { \textbf{0.66}} & { 0.76} & { 0.82} & { 0.64} & { 0.77} & { 0.82} & { 0.64} & { 0.71} & { 0.78} & { \textbf{0.66}} & { 0.76} & { 0.82} \\
 & {Sciences} & { 0.52} & { 0.59} & { 0.55} & { \textbf{0.68}} & { 0.74} & { 0.73} & { 0.67} & { 0.75} & { 0.73} & { 0.64} & { 0.67} & { 0.65} & { 0.67} & { 0.74} & { 0.72} \\
 & {Social sciences} & { 0.51} & { 0.56} & { 0.56} & { 0.65} & { 0.76} & { 0.74} & { \textbf{0.67}} & { 0.74} & { 0.74} & { 0.63} & { 0.68} & { 0.67} & { 0.66} & { 0.74} & { 0.73} \\
\multirow{-6}{*}{{ \textbf{LL-8B-Gen}}} & {Sports and recreation} & { 0.52} & { 0.59} & { 0.65} & { 0.66} & { 0.77} & 0.80 & 0.70 & { 0.76} & { 0.81} & { 0.62} & 0.70 & { 0.75} & \textbf{0.70} & { 0.77} & { 0.82} \\  \cmidrule{2-17}
 & { \textbf{Average}} & { 0.51} & { 0.58} & { 0.57} & { 0.68} & { 0.76} & { 0.76} & { \textbf{0.69}} & { 0.76} & { 0.76} & { 0.65} & { 0.69} & 0.70 & { 0.68} & { 0.75} & { 0.75} \\ \midrule
 & {Arts and humanity} & { 0.61} & { 0.56} & { 0.82} & { 0.67} & { 0.65} & { 0.86} & { \textbf{0.69}} & { 0.67} & { 0.87} & { 0.63} & { 0.59} & { 0.84} & { \textbf{0.69}} & { 0.68} & { 0.88} \\
 & {Geography and travel} & { 0.54} & { 0.59} & { 0.64} & { 0.59} & { 0.65} & { 0.68} & { 0.61} & { 0.68} & { 0.71} & { 0.55} & 0.60 & { 0.65} & { \textbf{0.62}} & { 0.68} & 0.70 \\
 & {Language and communication} & 0.60 & { 0.55} & 0.80 & { 0.63} & { 0.64} & { 0.85} & { \textbf{0.64}} & { 0.66} & { 0.85} & 0.60 & { 0.56} & { 0.81} & { \textbf{0.64}} & { 0.65} & { 0.85} \\
 & {Sciences} & { 0.58} & 0.60 & { 0.72} & { 0.59} & { 0.66} & { 0.77} & { 0.64} & 0.70 & { 0.79} & { 0.56} & { 0.58} & { 0.73} & {\textbf{ 0.65}} & { 0.71} & 0.80 \\
 & {Social sciences} & { 0.59} & { 0.58} & { 0.75} & { 0.63} & { 0.65} & { 0.79} & { 0.63} & { 0.67} & 0.80 & 0.60 & 0.60 & { 0.78} & { \textbf{0.64}} & { 0.67} & 0.80 \\
\multirow{-6}{*}{{ \textbf{MST-7B-Gen}}} & {Sports and recreation} & { 0.59} & { 0.58} & { 0.82} & { \textbf{0.68}} & { 0.68} & { 0.86} & { 0.67} & { 0.69} & { 0.87} & { 0.61} & 0.60 & { 0.83} & { \textbf{0.68}} & 0.70 & { 0.87} \\ \cmidrule{2-17}
 & { \textbf{Average}} & { 0.59} & { 0.58} & { 0.76} & { 0.63} & { 0.66} & 0.80 & { \textbf{0.65}} & { 0.68} & { 0.82} & { 0.59} & { 0.59} & { 0.77} & { \textbf{0.65}} & { 0.68} & { 0.82} \\ \midrule
 & {Arts and humanity} & { 0.52} & { 0.57} & { 0.59} & { 0.69} & { 0.78} & { 0.81} & { \textbf{0.71}} & { 0.78} & { 0.82} & { 0.66} & { 0.72} & { 0.76} & { \textbf{0.71}} & { 0.78} & { 0.82} \\
 & {Geography and travel} & { 0.48} & { 0.56} & { 0.41} & { 0.72} & { 0.75} & { 0.64} & { \textbf{0.73}} & { 0.75} & { 0.64} & { 0.69} & { 0.68} & { 0.57} & { 0.69} & { 0.73} & { 0.61} \\
 & {Language and communication} & { 0.52} & { 0.58} & { 0.68} & { \textbf{0.66}} & { 0.76} & { 0.82} & { 0.64} & { 0.77} & { 0.82} & { 0.64} & { 0.71} & { 0.78} & { \textbf{0.66}} & { 0.76} & { 0.82} \\
 & {Sciences} & { 0.52} & { 0.59} & { 0.55} & { 0.68} & { 0.74} & { 0.73} & { \textbf{0.67}} & { 0.75} & { 0.73} & { 0.64} & { 0.67} & { 0.65} & { \textbf{0.67}} & { 0.74} & { 0.72} \\
 & {Social sciences} & { 0.51} & { 0.56} & { 0.56} & { 0.65} & { 0.76} & { 0.74} & { \textbf{0.67}} & { 0.74} & { 0.74} & { 0.63} & { 0.68} & { 0.67} & { 0.66} & { 0.74} & { 0.73} \\
\multirow{-6}{*}{{ \textbf{GM-7B-Gen}}} & {Sports and recreation} & { 0.52} & { 0.59} & { 0.65} & { 0.66} & { 0.77} & 0.80 & \textbf{0.70} & { 0.76} & { 0.81} & { 0.62} & 0.70 & { 0.75} & \textbf{0.70} & { 0.77} & { 0.82} \\ \cmidrule{2-17}
 & { \textbf{Average}} & { 0.51} & { 0.58} & { 0.57} & { 0.68} & { 0.76} & { 0.76} & { \textbf{0.69}} & { 0.76} & { 0.76} & { 0.65} & { 0.69} & 0.70 & { 0.68} & { 0.75} & { 0.75} \\ \midrule
 & {Arts and humanity} & { \textbf{0.52}} & { 0.56} & { 0.67} & { 0.34} & { 0.54} & { 0.59} & { 0.38} & { 0.53} & { 0.64} & { 0.51} & { 0.56} & { 0.65} & { 0.44} & { 0.56} & { 0.64} \\
 & {Geography and travel} & { 0.47} & { 0.57} & { 0.51} & { 0.54} & { 0.58} & { 0.45} & { \textbf{0.57}} & { 0.59} & { 0.53} & { 0.56} & { 0.53} & { 0.44} & { 0.52} & { 0.58} & { 0.45} \\
 & {Language and communication} & { \textbf{0.51}} & { 0.58} & { 0.65} & 0.40 & { 0.54} & { 0.62} & { 0.41} & { 0.56} & { 0.62} & { 0.49} & { 0.53} & { 0.62} & { 0.45} & { 0.59} & { 0.64} \\
 & {Sciences} & { 0.49} & { 0.62} & { 0.62} & { \textbf{0.53}} & { 0.57} & { 0.51} & { 0.46} & { 0.58} & { 0.49} & { 0.51} & { 0.55} & { 0.51} & 0.50 & { 0.59} & { 0.54} \\
 & {Social sciences} & { 0.51} & { 0.59} & { 0.54} & { 0.49} & { 0.57} & 0.50 & { 0.49} & { 0.58} & { 0.51} & { \textbf{0.54}} & { 0.55} & { 0.51} & 0.50 & { 0.57} & { 0.49} \\
\multirow{-6}{*}{{ \textbf{LL-70B-Gen}}} & {Sports and recreation} & { \textbf{0.54}} & { 0.61} & { 0.65} & { 0.39} & { 0.55} & { 0.52} & 0.40 & { 0.57} & { 0.57} & { \textbf{0.54}} & { 0.56} & { 0.64} & { 0.42} & { 0.59} & { 0.57} \\ \cmidrule{2-17}
 & { \textbf{Average}} & { 0.51} & { 0.59} & { 0.61} & { 0.45} & { 0.56} & { 0.53} & { 0.45} & { 0.57} & { 0.56} & { \textbf{0.53}} & { 0.55} & { 0.56} & { 0.47} & { 0.58} & { 0.56} \\ \midrule
 & {Arts and humanity} & 0.60 & { 0.58} & { 0.79} & { 0.71} & { 0.76} & { 0.88} & { \textbf{0.72}} & { 0.78} & 0.90 & { 0.64} & { 0.67} & { 0.84} & { 0.64} & { 0.67} & { 0.84} \\
 & {Geography and travel} & { 0.55} & { 0.61} & { 0.64} & { \textbf{0.67}} & { 0.73} & { 0.74} & { 0.66} & { 0.73} & { 0.74} & { 0.61} & { 0.67} & { 0.69} & { 0.61} & { 0.67} & { 0.69} \\
 & {Language and communication} & 0.60 & { 0.58} & 0.80 & { \textbf{0.65}} & { 0.73} & { 0.87} & { \textbf{0.65}} & { 0.72} & { 0.87} & { 0.61} & { 0.62} & { 0.82} & { 0.61} & { 0.62} & { 0.82} \\
 & {Sciences} & { 0.57} & { 0.62} & { 0.73} & { 0.65} & { 0.75} & { 0.82} & \textbf{0.70} & { 0.77} & { 0.83} & 0.60 & { 0.66} & { 0.76} & 0.60 & { 0.66} & { 0.76} \\
 & {Social sciences} & { 0.54} & { 0.57} & { 0.73} & { \textbf{0.67}} & { 0.76} & { 0.84} & { \textbf{0.67}} & { 0.75} & { 0.84} & 0.60 & { 0.64} & { 0.78} & 0.60 & { 0.64} & { 0.78} \\
\multirow{-6}{*}{{ \textbf{ENSB-Gen}}} & {Sports and recreation} & 0.60 & { 0.64} & { 0.84} & { 0.69} & { 0.75} & { 0.88} & { \textbf{0.71}} & { 0.78} & 0.90 & { 0.65} & { 0.69} & { 0.86} & { 0.65} & { 0.69} & { 0.86} \\ \cmidrule{2-17}
 & { \textbf{Average}} & { 0.58} & 0.60 & { 0.76} & { 0.67} & { 0.75} & { 0.84} & { \textbf{0.69}} & { 0.76} & { 0.85} & { 0.62} & { 0.66} & { 0.79} & { 0.62} & { 0.66} & { 0.79} \\ \midrule
 & {Arts and humanity} & { 0.58} & 0.60 & { 0.76} & { 0.73} & 0.80 & { 0.87} & { 0.75} & { 0.82} & { 0.89} & { 0.64} & { 0.69} & { 0.82} & { \textbf{0.76}} & { 0.82} & { 0.88} \\
 & {Geography and travel} & { 0.55} & { 0.66} & 0.60 & { 0.74} & 0.80 & { 0.73} & { 0.72} & { 0.82} & { 0.74} & { 0.69} & { 0.76} & { 0.67} & { \textbf{0.76}} & { 0.84} & { 0.76} \\
 & {Language and communication} & { 0.56} & 0.60 & { 0.73} & { 0.67} & { 0.79} & { 0.86} & \textbf{0.70} & { 0.79} & { 0.85} & { 0.61} & { 0.64} & { 0.76} & \textbf{0.70} & { 0.78} & { 0.85} \\
 & {Sciences} & { 0.58} & { 0.67} & { 0.71} & { 0.71} & 0.80 & 0.80 & { \textbf{0.78}} & { 0.86} & { 0.84} & { 0.62} & { 0.69} & { 0.72} & { 0.74} & { 0.83} & { 0.82} \\
 & {Social sciences} & { 0.53} & { 0.62} & { 0.66} & { 0.71} & { 0.81} & 0.80 & { 0.72} & { 0.81} & { 0.81} & { 0.61} & { 0.65} & { 0.69} & { \textbf{0.73}} & { 0.82} & { 0.79} \\
\multirow{-6}{*}{{ \begin{tabular}[c]{@{}l@{}}\textbf{HA-Test}\end{tabular}}} & {Sports and recreation} & { 0.57} & { 0.68} & { 0.77} & { 0.71} & 0.80 & { 0.83} & { \textbf{0.74}} & { 0.83} & { 0.86} & { 0.63} & { 0.71} & { 0.78} & { 0.72} & { 0.81} & { 0.85} \\ \cmidrule{2-17}
 & { \textbf{Average}} & { 0.56} & { 0.64} & { 0.71} & { 0.71} & 0.80 & { 0.82} & { \textbf{0.74}} & { 0.82} & { 0.83} & { 0.63} & { 0.69} & { 0.74} & { \textbf{0.74}} & { 0.82} & { 0.83} \\ \toprule
\end{tabular}}
\caption{Hallucination detection with statistical classifier results for various models trained on labels obtained from Exact-match based approach on Jeopardy test sets. The best result highlighted in \textbf{bold}.}
\label{tab:string_jeopardy_ml}
\end{table*}

%% file: appendix_tables/table_string_kaggle_ML_appendix.tex
\begin{table*}
\centering\footnotesize
\resizebox{\textwidth}{!}{%
\begin{tabular}{@{}ll|ccc|ccc|ccc|ccc|ccc @{}} \\ \toprule
\multirow{2}{*}{\textbf{Test set}} & \multirow{2}{*}{\textbf{Sub-category}} & \multicolumn{3}{c}{{ \textbf{QR}}} & \multicolumn{3}{c}{{ \textbf{RR}}} & \multicolumn{3}{c}{{ \textbf{EC-EC}}} & \multicolumn{3}{c}{{ \textbf{C-C}}} & \multicolumn{3}{c}{{ \textbf{QR+RR}}} \\ \cmidrule{3-17}
&  & { \textbf{F1}} & { \textbf{AUC}} & { \textbf{B-ACC}} & { \textbf{F1}} & { \textbf{AUC}} & { \textbf{B-ACC}} & { \textbf{F1}} & { \textbf{AUC}} & { \textbf{B-ACC}} & { \textbf{F1}} & { \textbf{AUC}} & { \textbf{B-ACC}} & { \textbf{F1}} & { \textbf{AUC}} & { \textbf{B-ACC}} \\ \midrule
{ } & { GK} & 0.50& { 0.54} & { 0.71} & { 0.59} & { 0.69} & { 0.82} & { \textbf{0.61}} & { 0.71} & { 0.83} & { 0.54} & { 0.62} & { 0.76} & 0.60& { 0.72} & { 0.82} \\
{ } & { MathQA} & { 0.75} & { 0.52} & { 0.94} & { 0.74} & { 0.52} & { 0.95} & 0.80& { 0.53} & { 0.95} & { 0.74} & { 0.54} & { 0.95} & { \textbf{0.81}} & { 0.54} & { 0.95} \\
{ } & { MathQSA} & 0.70& { 0.54} & { 0.92} & { 0.71} & { 0.52} & { 0.92} & { 0.75} & { 0.55} & { 0.93} & { 0.68} & { 0.52} & { 0.92} & { \textbf{0.76}} & { 0.56} & { 0.93} \\
{ } & { SciQ} & { 0.55} & { 0.54} & { 0.82} & { \textbf{0.63}} & { 0.62} & { 0.86} & 0.60& { 0.63} & { 0.86} & { 0.56} & { 0.56} & { 0.84} & { 0.62} & { 0.64} & { 0.87} \\ \cmidrule{2-17}
\multirow{-5}{*}{{ \textbf{TL-1.1B-Gen}}} & { \textbf{Average}} & { 0.63} & { 0.54} & { 0.85} & { 0.67} & { 0.59} & { 0.89} & { 0.69} & { 0.61} & { 0.89} & { 0.63} & { 0.56} & { 0.87} & \textbf{0.70} & { 0.62} & { 0.89} \\ \midrule
{ } & { GK} & { 0.53} & { 0.49} & { 0.28} & { 0.64} & { 0.61} & 0.4 & { 0.69} & 0.70& { 0.42} & { 0.57} & { 0.54} & { 0.35} & \textbf{0.70} & { 0.69} & { 0.46} \\
{ } & { MathQA} & { \textbf{0.73}} & { 0.53} & { 0.95} & { 0.63} & { 0.54} & { 0.95} & { 0.69} & { 0.57} & { 0.95} & { 0.71} & { 0.55} & { 0.95} & 0.70& { 0.57} & { 0.95} \\
{ } & { MathQSA} & { \textbf{0.69}} & 0.50& { 0.91} & { 0.65} & { 0.55} & { 0.92} & { 0.65} & { 0.51} & { 0.91} & { 0.68} & 0.50& { 0.91} & { 0.64} & { 0.52} & { 0.91} \\
{ } & { SciQ} & 0.50& { 0.47} & { 0.39} & { 0.56} & 0.60& 0.50& { \textbf{0.57}} & { 0.58} & { 0.48} & 0.50& 0.50& { 0.42} & { 0.56} & { 0.57} & { 0.47} \\ \cmidrule{2-17}
\multirow{-5}{*}{{\textbf{PHI-3.5B-Gen}}} & { \textbf{Average}} & { 0.61} & 0.50& { 0.63} & { 0.62} & { 0.58} & { 0.69} & { \textbf{0.65}} & { 0.59} & { 0.69} & { 0.62} & { 0.52} & { 0.66} & { \textbf{0.65}} & { 0.59} & 0.70\\ \midrule
{ } & { GK} & { 0.54} & { 0.57} & { 0.33} & { 0.66} & { 0.67} & { 0.45} & 0.70& { 0.74} & { 0.52} & { 0.63} & 0.60& { 0.42} & { \textbf{0.71}} & { 0.66} & { 0.48} \\
{ } & { MathQA} & 0.70& { 0.54} & { 0.91} & { 0.69} & { 0.61} & { 0.93} & { \textbf{0.73}} & { 0.62} & { 0.93} & { 0.69} & { 0.57} & { 0.92} & { \textbf{0.73}} & { 0.62} & { 0.93} \\
{ } & { MathQSA} & 0.60& { 0.53} & { 0.78} & 0.60& { 0.64} & { 0.84} & { \textbf{0.64}} & { 0.66} & { 0.85} & { 0.59} & { 0.56} & { 0.79} & 0.60& { 0.63} & { 0.83} \\
{ } & { SciQ} & { 0.51} & 0.50& { 0.47} & { \textbf{0.61}} & 0.70& { 0.65} & { 0.59} & { 0.66} & { 0.62} & { 0.57} & 0.60& { 0.55} & { 0.59} & { 0.67} & { 0.62} \\ \cmidrule{2-17}
\multirow{-5}{*}{{ \textbf{LL-8B-Gen}}} & { \textbf{Average}} & { 0.59} & { 0.54} & { 0.62} & { 0.64} & { 0.66} & { 0.72} & { \textbf{0.67}} & { 0.67} & { 0.73} & { 0.62} & { 0.58} & { 0.67} & { 0.66} & { 0.65} & { 0.72} \\ \midrule
{ } & { GK} & { 0.27} & { 0.53} & { 0.46} & { 0.44} & { 0.46} & { 0.39} & { \textbf{0.46}} & { 0.41} & { 0.37} & { 0.44} & { 0.41} & { 0.38} & { 0.45} & { 0.47} & { 0.41} \\
{ } & { MathQA} & { \textbf{0.91}} & { 0.52} & { 0.95} & { 0.28} & { 0.49} & { 0.94} & 0.60& { 0.44} & { 0.93} & { 0.27} & { 0.43} & { 0.93} & { 0.14} & { 0.51} & { 0.94} \\
{ } & { MathQSA} & { \textbf{0.86}} & { 0.53} & { 0.91} & { 0.29} & { 0.49} & 0.90& { 0.66} & { 0.47} & { 0.89} & { 0.39} & { 0.39} & { 0.88} & { 0.27} & { 0.49} & 0.90\\
{ } & { SciQ} & { 0.42} & { 0.51} & { 0.57} & { 0.34} & { 0.48} & { 0.54} & { \textbf{0.50}} & { 0.49} & { 0.55} & { 0.42} & { 0.49} & { 0.55} & { 0.39} & { 0.51} & { 0.57} \\ \cmidrule{2-17}
\multirow{-5}{*}{{ \textbf{MST-7B-Gen}}} & { \textbf{Average}} & { \textbf{0.62}} & { 0.52} & { 0.72} & { 0.34} & { 0.48} & { 0.69} & { 0.56} & { 0.45} & { 0.69} & { 0.38} & { 0.43} & { 0.69} & { 0.31} & 0.50& { 0.71} \\ \midrule
{ } & { GK} & { 0.42} & { 0.46} & { 0.46} & { \textbf{0.63}} & { 0.66} & { 0.61} & { 0.58} & { 0.65} & { 0.58} & { 0.61} & { 0.66} & { 0.63} & { 0.53} & { 0.65} & { 0.54} \\
{ } & { MathQA} & { \textbf{0.68}} & { 0.48} & 0.90& 0.60& { 0.56} & { 0.92} & 0.60& { 0.55} & { 0.92} & { 0.64} & { 0.52} & { 0.91} & 0.60& { 0.55} & { 0.91} \\
{ } & { MathQSA} & { \textbf{0.64}} & { 0.51} & { 0.84} & { 0.58} & { 0.64} & { 0.88} & { 0.59} & { 0.63} & { 0.88} & 0.60& { 0.55} & { 0.85} & { 0.57} & { 0.62} & { 0.87} \\
{ } & { SciQ} & 0.50& { 0.49} & { 0.53} & { \textbf{0.56}} & { 0.65} & { 0.67} & { 0.54} & 0.60& { 0.61} & { 0.51} & { 0.55} & { 0.57} & { 0.52} & { 0.62} & { 0.62} \\ \cmidrule{2-17}
\multirow{-5}{*}{{ \textbf{GM-7B-Gen}}} & { \textbf{Average}} & { 0.56} & { 0.49} & { 0.68} & { \textbf{0.59}} & { 0.63} & { 0.77} & { 0.58} & { 0.61} & { 0.75} & { \textbf{0.59}} & { 0.57} & { 0.74} & { 0.56} & { 0.61} & { 0.74} \\ \midrule
{ } & { GK} & { 0.47} & { 0.52} & { 0.31} & {\textbf{0.72}} & 0.70& { 0.51} & { \textbf{0.72}} & 0.70& { 0.49} & { 0.58} & { 0.61} & { 0.44} & { 0.69} & { 0.66} & { 0.48} \\
{ } & { MathQA} & { 0.66} & { 0.51} & { 0.89} & { 0.69} & { 0.65} & { 0.93} & \textbf{0.70} & { 0.67} & { 0.93} & { 0.66} & { 0.54} & 0.90& { 0.69} & { 0.65} & { 0.93} \\
{ } & { MathQSA} & { 0.58} & { 0.49} & { 0.75} & { \textbf{0.65}} & { 0.71} & { 0.87} & { 0.63} & { 0.67} & { 0.85} & { 0.59} & { 0.56} & { 0.79} & { 0.63} & { 0.68} & { 0.86} \\
{ } & { SciQ} & { 0.49} & { 0.48} & { 0.44} & { \textbf{0.62}} & { 0.68} & { 0.63} & 0.60& { 0.65} & { 0.61} & { 0.53} & { 0.58} & { 0.53} & { 0.59} & { 0.63} & { 0.59} \\ \cmidrule{2-17}
\multirow{-5}{*}{{ \textbf{LL-70B-Gen}}} & { \textbf{Average}} & { 0.55} & 0.50& 0.60& { \textbf{0.67}} & { 0.69} & { 0.74} & { 0.66} & { 0.67} & { 0.72} & { 0.59} & { 0.57} & { 0.67} & { 0.65} & { 0.66} & { 0.72} \\ \midrule
{ } & { GK} & { 0.53} & { 0.52} & { 0.46} & { \textbf{0.63}} & { 0.71} & { 0.68} & { \textbf{0.63}} & { 0.73} & { 0.69} & { 0.58} & { 0.64} & { 0.62} & { 0.58} & { 0.64} & { 0.62} \\
{ } & { MathQA} & { 0.72} & { 0.54} & { 0.94} & { 0.71} & { 0.55} & { 0.94} & { \textbf{0.76}} & { 0.57} & { 0.95} & { 0.72} & { 0.56} & { 0.94} & { 0.72} & { 0.56} & { 0.94} \\
{ } & { MathQSA} & { 0.67} & { 0.54} & 0.90& { 0.68} & { 0.52} & 0.90& \textbf{0.70} & { 0.55} & { 0.91} & { 0.66} & { 0.53} & 0.90& { 0.66} & { 0.53} & 0.90\\
{ } & { SciQ} & { 0.49} & { 0.47} & { 0.54} & { \textbf{0.58}} & { 0.68} & { 0.71} & { 0.57} & { 0.65} & { 0.69} & { 0.52} & { 0.55} & { 0.59} & { 0.52} & { 0.55} & { 0.59} \\ \cmidrule{2-17}
\multirow{-5}{*}{{ \textbf{ENSB-Gen}}} & { \textbf{Average}} & 0.60& { 0.52} & { 0.71} & { 0.65} & { 0.62} & { 0.81} & { \textbf{0.67}} & { 0.63} & { 0.81} & { 0.62} & { 0.57} & { 0.76} & { 0.62} & { 0.57} & { 0.76} \\ \midrule
{ } & { GK} & { 0.57} & { 0.52} & { 0.36} & { 0.65} & { 0.72} & { 0.59} & \textbf{0.70} & { 0.75} & { 0.61} & { 0.58} & { 0.64} & { 0.53} & \textbf{0.70} & { 0.72} & 0.60\\
{ } & { MathQA} & { 0.76} & { 0.54} & { 0.98} & { 0.75} & { 0.59} & { 0.98} & { 0.78} & { 0.62} & { 0.98} & { 0.75} & { 0.62} & { 0.98} & { \textbf{0.80}} & { 0.66} & { 0.98} \\
{ } & { MathQSA} & { 0.72} & { 0.61} & { 0.97} & { 0.73} & { 0.58} & { 0.97} & { \textbf{0.76}} & { 0.63} & { 0.97} & { 0.71} & { 0.56} & { 0.96} & { 0.75} & { 0.63} & { 0.97} \\
{ } & { SciQ} & 0.50& { 0.48} & { 0.47} & 0.60& 0.70& { 0.66} & { \textbf{0.61}} & { 0.68} & { 0.64} & { 0.53} & { 0.55} & { 0.51} & { \textbf{0.61}} & 0.70& { 0.65} \\ \cmidrule{2-17}
\multirow{-5}{*}{{ \begin{tabular}[c]{@{}l@{}}\textbf{HA-Test}\end{tabular}}} & { \textbf{Average}} & { 0.64} & { 0.54} & 0.70& { 0.68} & { 0.65} & 0.80& { 0.71} & { 0.67} & 0.80& { 0.64} & { 0.59} & { 0.75} & { \textbf{0.72}} & { 0.68} & 0.80\\ \toprule
\end{tabular}}
\caption{Hallucination detection with statistical classifier results for various models trained on labels obtained from Exact-match based approach on Kaggle test sets. The best result highlighted in \textbf{bold}.}
\label{tab:string_ml_kaggle}
\end{table*}

%% file: appendix_tables/table_qwen_jeopardy_ML_appendix.tex
\begin{table*}
\centering\footnotesize\
\resizebox{\textwidth}{!}{%
\begin{tabular}{@{}ll|ccc|ccc|ccc|ccc|ccc @{}} \\ \toprule
& & \multicolumn{3}{c}{\textbf{QR}} & \multicolumn{3}{c}{\textbf{RR}} & \multicolumn{3}{c}{\textbf{EC-EC}} & \multicolumn{3}{c}{\textbf{C-C}} & \multicolumn{3}{c}{\textbf{QR+RR}} \\ \cmidrule{3-17}
\multirow{-2}{*}{\textbf{Test set}} & \multirow{-2}{*}{\textbf{Sub-category}} & \textbf{F1} & \textbf{AUC} &\textbf{ B-ACC} & \textbf{F1} & \textbf{AUC} & \textbf{B-ACC} & \textbf{F1} & \textbf{AUC} & \textbf{B-ACC} & \textbf{F1} & \textbf{AUC} & \textbf{B-ACC} & \textbf{F1} & \textbf{AUC} & \textbf{B-ACC} \\  \midrule
&Arts and humanity & 0.67 & 0.66 & 0.92 & 0.76 & 0.67 & 0.92 & \textbf{0.78} & 0.74 & 0.94 & 0.67 & 0.63 & 0.92 & \textbf{0.78} & 0.74 & 0.94 \\
& Geography and travel & 0.63 & 0.68 & 0.86 & 0.68 & 0.70& 0.86 & \textbf{0.71} & 0.76 & 0.90& 0.62 & 0.64 & 0.85 & \textbf{0.71} & 0.76 & 0.90\\
& Language and communication & 0.66 & 0.65 & 0.93 & 0.73 & 0.64 & 0.91 & \textbf{0.74} & 0.71 & 0.94 & 0.64 & 0.60& 0.91 & \textbf{0.74} & 0.70& 0.94 \\
& Sciences & 0.64 & 0.65 & 0.89 & 0.72 & 0.69 & 0.89 & \textbf{0.74} & 0.75 & 0.93 & 0.63 & 0.62 & 0.88 & 0.73 & 0.75 & 0.92 \\
& Social sciences & 0.66 & 0.68 & 0.90& 0.72 & 0.67 & 0.89 & 0.75 & 0.76 & 0.92 & 0.64 & 0.62 & 0.88 & \textbf{0.76} & 0.76 & 0.92 \\ 
\multirow{-6}{*}{\textbf{TL-1.1B-Gen}} & Sports and recreation & 0.66 & 0.67 & 0.91 & 0.76 & 0.71 & 0.92 & 0.78 & 0.76 & 0.94 & 0.64 & 0.58 & 0.89 & \textbf{0.79} & 0.77 & 0.94 \\ \cmidrule{2-17}
 & \textbf{Average} & 0.65 & 0.67 & 0.90& 0.73 & 0.68 & 0.90& \textbf{0.75} & 0.75 & 0.93 & 0.64 & 0.62 & 0.89 & \textbf{0.75} & 0.75 & 0.93 \\ \midrule
&Arts and humanity & 0.54 & 0.56 & 0.71 & 0.69 & 0.77 & 0.86 & \textbf{0.70} & 0.78 & 0.86 & 0.60& 0.65 & 0.78 & \textbf{0.70} & 0.78 & 0.86 \\
& Geography and travel & 0.54 & 0.59 & 0.40& 0.73 & 0.78 & 0.60& \textbf{0.75} & 0.81 & 0.64 & 0.58 & 0.64 & 0.46 & 0.74 & 0.81 & 0.64 \\
& Language and communication & 0.54 & 0.58 & 0.63 & 0.62 & 0.74 & 0.76 & \textbf{0.63} & 0.74 & 0.76 & 0.52 & 0.56 & 0.61 & \textbf{0.63} & 0.75 & 0.76 \\
& Sciences & 0.50 & 0.57 & 0.43 & 0.74 & 0.81 & 0.69 & \textbf{0.76} & 0.82 & 0.71 & 0.61 & 0.67 & 0.52 & 0.74 & 0.82 & 0.69 \\
& Social sciences & 0.57 & 0.63 & 0.52 & 0.73 & 0.79 & 0.72 & \textbf{0.75} & 0.81 & 0.74 & 0.58 & 0.64 & 0.58 & \textbf{0.75} & 0.81 & 0.74 \\ 
\multirow{-6}{*}{\textbf{PHI-3.5B-Gen}} & Sports and recreation & 0.51 & 0.55 & 0.65 & 0.63 & 0.70& 0.80& 0.66 & 0.76 & 0.82 & 0.60& 0.62 & 0.72 & \textbf{0.68} & 0.75 & 0.82 \\ \cmidrule{2-17}
 & \textbf{Average} & 0.53 & 0.58 & 0.56 & 0.69 & 0.77 & 0.74 & \textbf{0.71} & 0.79 & 0.76 & 0.58 & 0.63 & 0.61 & \textbf{0.71} & 0.79 & 0.75 \\ \midrule
&Arts and humanity & 0.53 & 0.62 & 0.52 & 0.80& 0.88 & 0.83 & \textbf{0.81} & 0.88 & 0.84 & 0.72 & 0.79 & 0.74 & 0.80& 0.88 & 0.84 \\
& Geography and travel & 0.61 & 0.66 & 0.32 & \textbf{0.89} & 0.93 & 0.73 & 0.88 & 0.91 & 0.70& 0.84 & 0.83 & 0.60& 0.88 & 0.92 & 0.72 \\
& Language and communication & 0.54 & 0.63 & 0.60& 0.72 & 0.83 & 0.80& \textbf{0.73} & 0.84 & 0.81 & 0.67 & 0.74 & 0.71 & 0.71 & 0.83 & 0.80\\
& Sciences & 0.54 & 0.63 & 0.41 & 0.82 & 0.87 & 0.74 & \textbf{0.83} & 0.89 & 0.77 & 0.74 & 0.79 & 0.64 & \textbf{0.83} & 0.88 & 0.75 \\
& Social sciences & 0.52 & 0.63 & 0.42 & 0.83 & 0.90& 0.77 & \textbf{0.85} & 0.90& 0.77 & 0.78 & 0.82 & 0.66 & \textbf{0.85} & 0.91 & 0.79 \\ 
\multirow{-6}{*}{\textbf{LL-8B-Gen}} & Sports and recreation & 0.57 & 0.66 & 0.56 & 0.79 & 0.87 & 0.81 & \textbf{0.80} & 0.88 & 0.81 & 0.73 & 0.81 & 0.74 & \textbf{0.80} & 0.88 & 0.81 \\ \cmidrule{2-17}
 & \textbf{Average} & 0.55 & 0.64 & 0.47 & 0.81 & 0.88 & 0.78 & \textbf{0.82} & 0.88 & 0.78 & 0.75 & 0.80& 0.68 & 0.81 & 0.88 & 0.79 \\ \midrule
&Arts and humanity & 0.57 & 0.59 & 0.73 & 0.73 & 0.81 & 0.87 & \textbf{0.75} & 0.83 & 0.89 & 0.63 & 0.69 & 0.80& 0.74 & 0.82 & 0.88 \\
& Geography and travel & 0.58 & 0.67 & 0.62 & 0.76 & 0.84 & 0.75 & \textbf{0.77} & 0.86 & 0.79 & 0.68 & 0.74 & 0.67 & 0.76 & 0.85 & 0.78 \\
& Language and communication & 0.56 & 0.61 & 0.74 & \textbf{0.65} & 0.76 & 0.83 & \textbf{0.65} & 0.78 & 0.85 & 0.59 & 0.64 & 0.76 & 0.66 & 0.78 & 0.85 \\
& Sciences & 0.55 & 0.64 & 0.64 & 0.73 & 0.84 & 0.81 & 0.74 & 0.85 & 0.83 & 0.61 & 0.69 & 0.69 & \textbf{0.75} & 0.85 & 0.83 \\
& Social sciences & 0.56 & 0.64 & 0.68 & 0.74 & 0.83 & 0.82 & \textbf{0.76} & 0.85 & 0.84 & 0.63 & 0.71 & 0.72 & \textbf{0.76} & 0.85 & 0.84 \\ 
\multirow{-6}{*}{\textbf{ENSB-Gen}} & Sports and recreation & 0.57 & 0.64 & 0.75 & 0.69 & 0.79 & 0.85 & \textbf{0.73} & 0.82 & 0.87 & 0.61 & 0.68 & 0.78 & 0.72 & 0.81 & 0.87 \\ \cmidrule{2-17}
 & \textbf{Average} & 0.57 & 0.63 & 0.69 & 0.72 & 0.81 & 0.82 & \textbf{0.73} & 0.83 & 0.85 & 0.63 & 0.69 & 0.74 & \textbf{0.73} & 0.83 & 0.84 \\ \midrule
&Arts and humanity & 0.58 & 0.61 & 0.72 & 0.59 & 0.69 & 0.79 & \textbf{0.60} & 0.70& 0.79 & \textbf{0.60} & 0.66 & 0.77 & \textbf{0.60} & 0.70& 0.79 \\
& Geography and travel & 0.55 & 0.57 & 0.43 & \textbf{0.66} & 0.67 & 0.53 & \textbf{0.66} & 0.65 & 0.52 & 0.65 & 0.65 & 0.49 & \textbf{0.66} & 0.68 & 0.55 \\
& Language and communication & 0.55 & 0.59 & 0.70& 0.56 & 0.71 & 0.77 & 0.56 & 0.72 & 0.78 & \textbf{0.59} & 0.69 & 0.76 & 0.56 & 0.70& 0.77 \\
& Sciences & 0.53 & 0.60& 0.46 & \textbf{0.66} & 0.70& 0.57 & \textbf{0.66} & 0.71 & 0.57 & \textbf{0.66} & 0.67 & 0.54 & \textbf{0.66} & 0.71 & 0.57 \\
& Social sciences & 0.51 & 0.58 & 0.52 & \textbf{0.65} & 0.69 & 0.64 & \textbf{0.65} & 0.70& 0.65 & \textbf{0.65} & 0.66 & 0.60& \textbf{0.65} & 0.69 & 0.64 \\
\multirow{-6}{*}{\textbf{GM-7B-Gen}} & Sports and recreation & 0.50 & 0.56 & 0.60& 0.59 & 0.69 & 0.68 & \textbf{0.62} & 0.70& 0.69 & 0.61 & 0.65 & 0.66 & 0.61 & 0.71 & 0.69 \\  \cmidrule{2-17}
 & \textbf{Average} & 0.54 & 0.59 & 0.57 & 0.62 & 0.69 & 0.66 & \textbf{0.63} & 0.70& 0.67 & 0.63 & 0.66 & 0.64 & 0.62 & 0.70& 0.67 \\ \midrule
&Arts and humanity & 0.60& 0.61 & 0.79 & 0.67 & 0.71 & 0.84 & \textbf{0.70} & 0.74 & 0.86 & 0.59 & 0.62 & 0.80& 0.69 & 0.74 & 0.86 \\
& Geography and travel & 0.58 & 0.67 & 0.67 & 0.65 & 0.74 & 0.70& 0.70& 0.79 & 0.77 & 0.55 & 0.63 & 0.65 & \textbf{0.71} & 0.79 & 0.77 \\
& Language and communication & 0.57 & 0.60& 0.76 & 0.63 & 0.71 & 0.81 & \textbf{0.65} & 0.72 & 0.83 & 0.56 & 0.58 & 0.76 & 0.63 & 0.71 & 0.83 \\
& Sciences & 0.58 & 0.64 & 0.70& 0.66 & 0.74 & 0.76 & 0.68 & 0.78 & 0.80& 0.56 & 0.61 & 0.69 & \textbf{0.69} & 0.78 & 0.81 \\
& Social sciences & 0.57 & 0.64 & 0.72 & 0.65 & 0.72 & 0.77 & \textbf{0.69} & 0.77 & 0.81 & 0.57 & 0.64 & 0.73 & 0.68 & 0.77 & 0.81 \\
\multirow{-6}{*}{\textbf{MST-7B-Gen}} & Sports and recreation & 0.58 & 0.63 & 0.77 & 0.67 & 0.74 & 0.82 & \textbf{0.69} & 0.77 & 0.86 & 0.58 & 0.62 & 0.77 & 0.68 & 0.76 & 0.85 \\ \cmidrule{2-17}
 & \textbf{Average} & 0.58 & 0.63 & 0.74 & 0.66 & 0.73 & 0.78 & \textbf{0.69} & 0.76 & 0.82 & 0.57 & 0.62 & 0.73 & 0.68 & 0.76 & 0.82 \\ \midrule
&Arts and humanity & 0.52 & 0.59 & 0.57 & 0.38 & 0.54 & 0.54 & 0.37 & 0.57 & 0.55 & \textbf{0.54} & 0.59 & 0.56 & 0.40& 0.60& 0.58 \\
& Geography and travel & 0.52 & 0.66 & 0.42 & 0.68 & 0.56 & 0.26 & \textbf{0.69} & 0.67 & 0.34 & \textbf{0.69} & 0.59 & 0.36 & \textbf{0.69} & 0.68 & 0.36 \\
& Language and communication & 0.53 & 0.57 & 0.49 & 0.44 & 0.49 & 0.41 & 0.44 & 0.55 & 0.45 & \textbf{0.55} & 0.51 & 0.49 & 0.47 & 0.55 & 0.47 \\
& Sciences & 0.49 & 0.62 & 0.46 & 0.60& 0.52 & 0.32 & 0.59 & 0.60& 0.35 & \textbf{0.64} & 0.54 & 0.38 & 0.60& 0.57 & 0.34 \\
& Social sciences & 0.56 & 0.63 & 0.44 & 0.62 & 0.59 & 0.35 & 0.62 & 0.59 & 0.36 & \textbf{0.65} & 0.60& 0.39 & 0.63 & 0.60& 0.40\\
\multirow{-6}{*}{\textbf{LL-70B-Gen}} & Sports and recreation & 0.51 & 0.60& 0.55 & 0.44 & 0.52 & 0.41 & 0.44 & 0.64 & 0.52 & \textbf{0.67} & 0.62 & 0.62 & 0.44 & 0.61 & 0.51 \\ \cmidrule{2-17}
 & \textbf{Average} & 0.52 & 0.61 & 0.49 & 0.53 & 0.54 & 0.38 & 0.53 & 0.60& 0.43 & \textbf{0.62} & 0.58 & 0.47 & 0.54 & 0.60& 0.44 \\ \midrule
&Arts and humanity & 0.57 & 0.58 & 0.73 & 0.73 & 0.82 & 0.88 & \textbf{0.75} & 0.83 & 0.89 & 0.65 & 0.70& 0.81 & \textbf{0.75} & 0.82 & 0.88 \\
& Geography and travel & 0.56 & 0.65 & 0.59 & \textbf{0.76} & 0.84 & 0.75 & 0.75 & 0.85 & 0.75 & 0.70& 0.77 & 0.67 & 0.75 & 0.85 & 0.75 \\
& Language and communication & 0.57 & 0.62 & 0.75 & \textbf{0.70} & 0.81 & 0.87 & 0.69 & 0.82 & 0.87 & 0.57 & 0.65 & 0.76 & 0.69 & 0.81 & 0.87 \\
& Sciences & 0.53 & 0.63 & 0.68 & 0.74 & 0.86 & 0.83 & \textbf{0.76} & 0.87 & 0.84 & 0.65 & 0.74 & 0.74 & 0.75 & 0.86 & 0.84 \\
& Social sciences & 0.58 & 0.66 & 0.69 & 0.75 & 0.83 & 0.81 & 0.76 & 0.85 & 0.84 & 0.61 & 0.69 & 0.72 & \textbf{0.77} & 0.86 & 0.84 \\
\multirow{-6}{*}{\begin{tabular}[c]{@{}l@{}}\textbf{HA-Test}\end{tabular}} & Sports and recreation & 0.57 & 0.65 & 0.72 & 0.67 & 0.79 & 0.82 & 0.72 & 0.82 & 0.85 & 0.63 & 0.72 & 0.78 & \textbf{0.73} & 0.82 & 0.85 \\ \cmidrule{2-17}
 & \textbf{Average} & 0.56 & 0.63 & 0.69 & 0.73 & 0.83 & 0.83 & \textbf{0.74} & 0.84 & 0.84 & 0.64 & 0.71 & 0.75 & \textbf{0.74} & 0.84 & 0.84 \\ \bottomrule
\end{tabular}} 
\caption{Hallucination detection with statistical classifier results for various models trained on labels obtained from LLM-based approach on Jeopardy test sets. The best result highlighted in \textbf{bold}.}
\label{tab:qwen_ml_jeopardy}
\end{table*}

%% file: appendix_tables/table_qwen_kaggle_ML_appendix.tex
\begin{table*}
\centering\footnotesize
\resizebox{\textwidth}{!}{%
\begin{tabular}{@{}ll|ccc|ccc|ccc|ccc|ccc @{}} \\ \toprule
 &  & \multicolumn{3}{c}{\textbf{QR}} & \multicolumn{3}{c}{\textbf{RR}} & \multicolumn{3}{c}{\textbf{EC-EC}} & \multicolumn{3}{c}{\textbf{C-C}} & \multicolumn{3}{c}{\textbf{QR+RR}} \\ \cmidrule{3-17}
\multirow{-2}{*}{\textbf{Test set}} & \multirow{-2}{*}{\textbf{Sub-category}} & \textbf{F1} & \textbf{AUC} & \textbf{B-ACC} & \textbf{F1} & \textbf{AUC} & \textbf{B-ACC} & \textbf{F1} & \textbf{AUC} & \textbf{B-ACC} & \textbf{F1} & \textbf{AUC} & \textbf{B-ACC} & \textbf{F1} & \textbf{AUC} & \textbf{B-ACC} \\ \midrule
 & GK & 0.55 & 0.58 & 0.72 & 0.70& 0.79 & 0.84 & 0.73 & 0.81 & 0.87 & 0.62 & 0.67 & 0.79 & \textbf{0.77} & 0.82 & 0.87 \\
 & MathQA & 0.87 & 0.65 & 1\phantom{0}\phantom{0} & 0.91 & 0.48 & 0.99 & \textbf{0.95} & 0.63 & 1\phantom{0}\phantom{0} & 0.88 & 0.65 & 1\phantom{0}\phantom{0} & \textbf{0.95} & 0.66 & 1\phantom{0}\phantom{0} \\
 & MathQSA & 0.83 & 0.57 & 0.99 & 0.90& 0.49 & 0.99 & \textbf{0.92} & 0.59 & 0.99 & 0.84 & 0.60& 0.99 & \textbf{0.92} & 0.55 & 0.99 \\
\multirow{-4}{*}{\textbf{TL-1.1B-Gen}} & SciQ & 0.57 & 0.58 & 0.79 & \textbf{0.70} & 0.68 & 0.84 & 0.69 & 0.70& 0.85 & 0.57 & 0.57 & 0.80& 0.69 & 0.69 & 0.85 \\ \cmidrule{2-17}
 & \textbf{Average} & 0.71 & 0.60& 0.88 & 0.80& 0.61 & 0.92 & 0.82 & 0.68 & 0.93 & 0.73 & 0.62 & 0.90& \textbf{0.83} & 0.68 & 0.93 \\ \midrule
 & GK & 0.53 & 0.41 & 0.13 & 0.72 & 0.65 & 0.35 & 0.74 & 0.59 & 0.32 & 0.62 & 0.53 & 0.18 & \textbf{0.76} & 0.64 & 0.34 \\
 & MathQA & \textbf{0.67} & 0.56 & 0.83 & 0.60& 0.58 & 0.84 & 0.65 & 0.61 & 0.85 & 0.66 & 0.56 & 0.84 & 0.66 & 0.62 & 0.85 \\
 & MathQSA & 0.60& 0.52 & 0.76 & 0.63 & 0.63 & 0.82 & 0.64 & 0.63 & 0.81 & 0.62 & 0.55 & 0.77 & \textbf{0.65} & 0.64 & 0.82 \\
\multirow{-4}{*}{\textbf{PHI-3.5B-Gen}} & SciQ & 0.53 & 0.51 & 0.23 & 0.67 & 0.67 & 0.39 & 0.70& 0.65 & 0.37 & 0.54 & 0.53 & 0.25 & \textbf{0.71} & 0.67 & 0.38 \\ \cmidrule{2-17}
 & \textbf{Average} & 0.58 & 0.50& 0.49 & 0.66 & 0.63 & 0.60& 0.68 & 0.62 & 0.59 & 0.61 & 0.54 & 0.51 & \textbf{0.70} & 0.64 & 0.60\\ \midrule
 & GK & 0.48 & 0.52 & 0.18 & 0.78 & 0.67 & 0.33 & 0.81 & 0.71 & 0.37 & 0.62 & 0.58 & 0.25 & \textbf{0.82} & 0.71 & 0.38 \\
 & MathQA & 0.67 & 0.54 & 0.83 & 0.74 & 0.66 & 0.88 & \textbf{0.76} & 0.67 & 0.88 & 0.69 & 0.59 & 0.85 & \textbf{0.76} & 0.67 & 0.88 \\
 & MathQSA & 0.60& 0.54 & 0.73 & \textbf{0.70} & 0.71 & 0.83 & \textbf{0.70} & 0.70& 0.82 & 0.61 & 0.58 & 0.75 & \textbf{0.70} & 0.69 & 0.82 \\
\multirow{-4}{*}{\textbf{LL-8B-Gen}} & SciQ & 0.49 & 0.53 & 0.30 & 0.71 & 0.73 & 0.51 & \textbf{0.72} & 0.73 & 0.51 & 0.60& 0.63 & 0.40& \textbf{0.72} & 0.72 & 0.51 \\ \cmidrule{2-17}
 & \textbf{Average} & 0.56 & 0.53 & 0.51 & 0.73 & 0.69 & 0.64 & \textbf{0.75} & 0.70& 0.65 & 0.63 & 0.60& 0.56 & \textbf{0.75} & 0.70& 0.65 \\ \midrule
 & GK & 0.47 & 0.50& 0.38 & 0.70& 0.76 & 0.66 & \textbf{0.76} & 0.76 & 0.69 & 0.65 & 0.68 & 0.60& 0.72 & 0.76 & 0.69 \\
 & MathQA & 0.73 & 0.56 & 0.91 & 0.78 & 0.67 & 0.93 & \textbf{0.80} & 0.70& 0.94 & 0.74 & 0.60& 0.92 & \textbf{0.80} & 0.70& 0.94 \\
 & MathQSA & 0.70& 0.53 & 0.89 & 0.77 & 0.67 & 0.92 & 0.78 & 0.68 & 0.92 & 0.71 & 0.56 & 0.89 & \textbf{0.79} & 0.68 & 0.92 \\
\multirow{-4}{*}{\textbf{ENSB-Gen}} & SciQ & 0.50& 0.51 & 0.42 & 0.68 & 0.77 & 0.68 & 0.69 & 0.75 & 0.66 & 0.54 & 0.57 & 0.47 & \textbf{0.70} & 0.76 & 0.67 \\ \cmidrule{2-17}
 & \textbf{Average} & 0.60& 0.53 & 0.65 & 0.73 & 0.72 & 0.80& \textbf{0.76} & 0.72 & 0.80& 0.66 & 0.60& 0.72 & 0.75 & 0.73 & 0.81 \\ \midrule
 & GK & 0.52 & 0.56 & 0.37 & \textbf{0.68} & 0.65 & 0.52 & 0.64 & 0.62 & 0.48 & 0.60& 0.65 & 0.46 & 0.66 & 0.64 & 0.51 \\
 & MathQA & \textbf{0.72} & 0.51 & 0.90& 0.69 & 0.64 & 0.93 & 0.69 & 0.64 & 0.93 & \textbf{0.72} & 0.58 & 0.92 & 0.69 & 0.65 & 0.93 \\
 & MathQSA & \textbf{0.68} & 0.52 & 0.83 & 0.64 & 0.70& 0.90& 0.66 & 0.72 & 0.91 & 0.65 & 0.57 & 0.86 & 0.65 & 0.71 & 0.91 \\
\multirow{-4}{*}{\textbf{GM-7B-Gen}} & SciQ & 0.47 & 0.53 & 0.36 & \textbf{0.66} & 0.68 & 0.53 & \textbf{0.66} & 0.65 & 0.50& 0.53 & 0.58 & 0.42 & \textbf{0.66} & 0.66 & 0.51 \\ \cmidrule{2-17}
 & \textbf{Average} & 0.60& 0.53 & 0.62 & \textbf{0.67} & 0.67 & 0.72 & 0.66 & 0.66 & 0.71 & 0.63 & 0.60& 0.67 & \textbf{0.67} & 0.67 & 0.72 \\ \midrule
 & GK & 0.16 & 0.56 & 0.38 & \textbf{0.49} & 0.46 & 0.32 & 0.16 & 0.40& 0.28 & 0.17 & 0.60& 0.38 & 0.16 & 0.59 & 0.35 \\
 & MathQA & \textbf{0.90} & 0.52 & 0.94 & 0.61 & 0.48 & 0.93 & \textbf{0.90} & 0.49 & 0.93 & \textbf{0.90} & 0.53 & 0.94 & \textbf{0.90} & 0.55 & 0.94 \\
 & MathQSA & \textbf{0.87} & 0.53 & 0.92 & 0.63 & 0.52 & 0.92 & \textbf{0.87} & 0.49 & 0.91 & \textbf{0.87} & 0.54 & 0.92 & \textbf{0.87} & 0.55 & 0.93 \\
\multirow{-4}{*}{\textbf{MST-7B-Gen}} & SciQ & 0.20 & 0.51 & 0.38 & \textbf{0.51} & 0.49 & 0.37 & 0.21 & 0.49 & 0.37 & 0.20 & 0.52 & 0.39 & 0.21 & 0.52 & 0.38 \\ \cmidrule{2-17}
 & \textbf{Average} & 0.53 & 0.53 & 0.66 & \textbf{0.56} & 0.49 & 0.64 & 0.54 & 0.47 & 0.62 & 0.54 & 0.55 & 0.66 & 0.54 & 0.55 & 0.65 \\ \midrule
 & GK & 0.46 & 0.41 & 0.11 & 0.84 & 0.81 & 0.52 & \textbf{0.88} & 0.84 & 0.57 & 0.58 & 0.69 & 0.36 & 0.83 & 0.79 & 0.42 \\
 & MathQA & 0.67 & 0.52 & 0.83 & \textbf{0.81} & 0.73 & 0.91 & \textbf{0.81} & 0.74 & 0.91 & 0.69 & 0.57 & 0.85 & \textbf{0.81} & 0.73 & 0.91 \\
 & MathQSA & 0.60& 0.51 & 0.72 & \textbf{0.76} & 0.78 & 0.88 & \textbf{0.76} & 0.79 & 0.88 & 0.62 & 0.56 & 0.76 & \textbf{0.76} & 0.78 & 0.88 \\
\multirow{-4}{*}{\textbf{LL-70B-Gen}} & SciQ & 0.47 & 0.52 & 0.26 & \textbf{0.75} & 0.76 & 0.51 & 0.74 & 0.73 & 0.50& 0.53 & 0.59 & 0.34 & 0.74 & 0.74 & 0.50\\ \cmidrule{2-17}
 & \textbf{Average} & 0.55 & 0.49 & 0.48 & 0.79 & 0.77 & 0.71 & \textbf{0.80} & 0.78 & 0.72 & 0.61 & 0.60& 0.58 & 0.79 & 0.76 & 0.68 \\ \midrule
 & GK & 0.49 & 0.51 & 0.36 & 0.71 & 0.77 & 0.64 & \textbf{0.77} & 0.78 & 0.65 & 0.67 & 0.72 & 0.61 & 0.73 & 0.78 & 0.65 \\
 & MathQA & 0.81 & 0.50& 0.97 & 0.83 & 0.58 & 0.98 & 0.86 & 0.67 & 0.98 & 0.82 & 0.63 & 0.98 & \textbf{0.87} & 0.67 & 0.98 \\
 & MathQSA & 0.78 & 0.56 & 0.97 & 0.83 & 0.56 & 0.96 & \textbf{0.84} & 0.61 & 0.97 & 0.78 & 0.47 & 0.95 & \textbf{0.84} & 0.62 & 0.97 \\
\multirow{-4}{*}{\begin{tabular}[c]{@{}l@{}}\textbf{HA-Test}\end{tabular}} & SciQ & 0.50& 0.48 & 0.48 & \textbf{0.65} & 0.74 & 0.70& \textbf{0.65} & 0.71 & 0.67 & 0.51 & 0.55 & 0.53 & \textbf{0.65} & 0.71 & 0.67 \\ \cmidrule{2-17}
 & \textbf{Average} & 0.65 & 0.51 & 0.70& 0.76 & 0.66 & 0.82 & \textbf{0.78} & 0.69 & 0.82 & 0.70& 0.59 & 0.77 & 0.77 & 0.70& 0.82 \\ \bottomrule
\end{tabular}}
\caption{Hallucination detection with statistical classifier results for various models trained on labels obtained from LLM-based approach on Kaggle test sets. The best result highlighted in \textbf{bold}.}
\label{tab:qwen_kaggle_ml}
\end{table*}

%% file: appendix_tables/table_string_jeopardy_BERT_appendix.tex
\begin{table*}
\centering\footnotesize
\resizebox{\textwidth}{!}{%
\begin{tabular}{ll|ccc|ccc|ccc|ccc|ccc|ccc} \\ \toprule
\multirow{2}{*}{\textbf{Test set}} & \multirow{2}{*}{\textbf{Sub-category}} & \multicolumn{3}{c}{\textbf{QR}} & \multicolumn{3}{c}{\textbf{RR}} & \multicolumn{3}{c}{\textbf{EC-EC}} & \multicolumn{3}{c}{\textbf{CC}} & \multicolumn{3}{c}{\textbf{QR-RR}} & \multicolumn{3}{c}{\textbf{q-r+Q-R+R-R}} \\ \cmidrule{3-20}
 &  & \textbf{F1} & \textbf{AUC} & \textbf{B-ACC} & \textbf{F1} & \textbf{AUC} & \textbf{B-ACC} & \textbf{F1} & \textbf{AUC} & \textbf{B-ACC} & \textbf{F1} & \textbf{AUC} & \textbf{B-ACC} & \textbf{F1} & \textbf{AUC} & \textbf{B-ACC} & \textbf{F1} & \textbf{AUC} & \textbf{B-ACC} \\ \midrule
\multirow{6}{*}{\textbf{TL-1.1B-Test}} & Arts and humanity & 0.67 & 0.61 & 0.94 & 0.85 & 0.58 & 0.93 & 0.85 & 0.66 & 0.94 & 0.65 & 0.62 & 0.94 & 0.85 & 0.64 & 0.94 & \textbf{0.90} & 0.89 & 0.99 \\
 & Geography and travel & 0.62 & 0.64 & 0.86 & 0.70& 0.61 & 0.84 & 0.75 & 0.70& 0.86 & 0.63 & 0.62 & 0.85 & 0.72 & 0.68 & 0.86 & \textbf{0.84} & 0.93 & 0.98 \\
 & Language and communication & 0.62 & 0.59 & 0.94 & 0.83 & 0.56 & 0.93 & 0.82 & 0.63 & 0.94 & 0.62 & 0.60& 0.94 & 0.81 & 0.63 & 0.94 & \textbf{0.90} & 0.87 & 0.98 \\
 & Sciences & 0.64 & 0.61 & 0.91 & 0.81 & 0.61 & 0.90& 0.81 & 0.68 & 0.92 & 0.65 & 0.60& 0.91 & 0.80& 0.67 & 0.92 & \textbf{0.86} & 0.89 & 0.98 \\
 & Social sciences & 0.65 & 0.62 & 0.92 & 0.81 & 0.59 & 0.91 & 0.82 & 0.69 & 0.93 & 0.64 & 0.61 & 0.91 & 0.81 & 0.66 & 0.92 & \textbf{0.88} & 0.92 & 0.99 \\
 & Sports and recreation & 0.66 & 0.63 & 0.93 & 0.84 & 0.59 & 0.91 & 0.84 & 0.68 & 0.93 & 0.64 & 0.60& 0.92 & 0.84 & 0.65 & 0.92 & \textbf{0.90} & 0.92 & 0.99 \\ \midrule
\multirow{6}{*}{\textbf{PHI-3.5B-Gen}} & Arts and humanity & 0.60& 0.63 & 0.84 & 0.69 & 0.73 & 0.89 & 0.74 & 0.78 & 0.91 & 0.65 & 0.70& 0.87 & 0.73 & 0.78 & 0.91 & \textbf{0.81} & 0.88 & 0.96 \\
 & Geography and travel & 0.56 & 0.61 & 0.57 & 0.64 & 0.69 & 0.68 & 0.65 & 0.72 & 0.69 & 0.61 & 0.62 & 0.59 & 0.66 & 0.72 & 0.69 & \textbf{0.75} & 0.86 & 0.86 \\
 & Language and communication & 0.53 & 0.63 & 0.77 & 0.57 & 0.72 & 0.83 & 0.61 & 0.74 & 0.84 & 0.59 & 0.65 & 0.79 & 0.59 & 0.73 & 0.84 & \textbf{0.74} & 0.83 & 0.91 \\
 & Sciences & 0.56 & 0.61 & 0.62 & 0.66 & 0.73 & 0.75 & 0.67 & 0.75 & 0.76 & 0.63 & 0.67 & 0.68 & 0.67 & 0.75 & 0.76 & \textbf{0.72} & 0.81 & 0.81 \\
 & Social sciences & 0.54 & 0.62 & 0.69 & 0.63 & 0.71 & 0.78 & 0.67 & 0.74 & 0.81 & 0.61 & 0.65 & 0.74 & 0.66 & 0.73 & 0.80& \textbf{0.75} & 0.85 & 0.89 \\
 & Sports and recreation & 0.58 & 0.65 & 0.82 & 0.64 & 0.71 & 0.86 & 0.70& 0.76 & 0.88 & 0.61 & 0.67 & 0.85 & 0.69 & 0.76 & 0.88 & \textbf{0.81} & 0.91 & 0.96 \\ \midrule
\multirow{6}{*}{\textbf{LL-8B-Gen}} & Arts and humanity & 0.57 & 0.60& 0.62 & 0.74 & 0.80& 0.84 & 0.74 & 0.81 & 0.84 & 0.69 & 0.76 & 0.79 & 0.74 & 0.81 & 0.84 & \textbf{0.75} & 0.82 & 0.84 \\
 & Geography and travel & 0.62 & 0.62 & 0.46 & 0.71 & 0.76 & 0.67 & 0.69 & 0.77 & 0.68 & 0.72 & 0.73 & 0.63 & 0.70& 0.77 & 0.68 & \textbf{0.78} & 0.86 & 0.80\\
 & Language and communication & 0.55 & 0.63 & 0.70& 0.65 & 0.80& 0.85 & 0.65 & 0.81 & 0.86 & 0.62 & 0.75 & 0.81 & 0.66 & 0.81 & 0.86 & \textbf{0.72} & 0.82 & 0.86 \\
 & Sciences & 0.59 & 0.63 & 0.58 & 0.67 & 0.75 & 0.76 & 0.67 & 0.77 & 0.78 & 0.66 & 0.74 & 0.72 & 0.67 & 0.76 & 0.77 & \textbf{0.69} & 0.79 & 0.77 \\
 & Social sciences & 0.56 & 0.59 & 0.59 & 0.65 & 0.77 & 0.77 & 0.66 & 0.76 & 0.76 & 0.67 & 0.73 & 0.73 & 0.66 & 0.76 & 0.76 & \textbf{0.70} & 0.81 & 0.80\\
 & Sports and recreation & 0.59 & 0.65 & 0.69 & 0.69 & 0.79 & 0.82 & 0.68 & 0.81 & 0.83 & 0.66 & 0.74 & 0.78 & 0.69 & 0.80& 0.83 & \textbf{0.74} & 0.84 & 0.86 \\ \midrule
\multirow{6}{*}{\textbf{MST-7B-Gen}} & Arts and humanity & 0.59 & 0.57 & 0.82 & 0.75 & 0.69 & 0.87 & 0.75 & 0.72 & 0.89 & 0.67 & 0.62 & 0.85 & 0.74 & 0.71 & 0.88 & \textbf{0.84} & 0.90& 0.97 \\
 & Geography and travel & 0.57 & 0.58 & 0.65 & 0.64 & 0.68 & 0.69 & 0.67 & 0.72 & 0.72 & 0.58 & 0.63 & 0.68 & 0.66 & 0.70& 0.71 & \textbf{0.79} & 0.91 & 0.94 \\
 & Language and communication & 0.56 & 0.57 & 0.80& 0.70& 0.68 & 0.86 & 0.68 & 0.68 & 0.85 & 0.68 & 0.60& 0.82 & 0.66 & 0.67 & 0.85 & \textbf{0.80} & 0.87 & 0.95 \\
 & Sciences & 0.57 & 0.59 & 0.72 & 0.71 & 0.72 & 0.80& 0.70& 0.74 & 0.82 & 0.62 & 0.62 & 0.76 & 0.70& 0.73 & 0.81 & \textbf{0.79} & 0.89 & 0.94 \\
 & Social sciences & 0.57 & 0.58 & 0.76 & 0.70& 0.68 & 0.81 & 0.70& 0.71 & 0.83 & 0.66 & 0.63 & 0.79 & 0.69 & 0.69 & 0.82 & \textbf{0.80} & 0.89 & 0.95 \\
 & Sports and recreation & 0.58 & 0.61 & 0.83 & 0.72 & 0.68 & 0.86 & 0.71 & 0.73 & 0.88 & 0.67 & 0.64 & 0.85 & 0.71 & 0.72 & 0.87 & \textbf{0.82} & 0.90& 0.97 \\ \midrule
\multirow{6}{*}{\textbf{GM-7B-Gen}} & Arts and humanity & 0.58 & 0.60& 0.62 & 0.73 & 0.80& 0.84 & 0.73 & 0.81 & 0.84 & 0.69 & 0.76 & 0.79 & \textbf{0.74} & 0.81 & 0.84 & 0.67 & 0.67 & 0.62 \\
 & Geography and travel & 0.62 & 0.63 & 0.47 & 0.71 & 0.76 & 0.67 & 0.70& 0.77 & 0.68 & \textbf{0.72} & 0.73 & 0.63 & 0.70& 0.77 & 0.68 & 0.69 & 0.75 & 0.53 \\
 & Language and communication & 0.56 & 0.63 & 0.70& 0.65 & 0.81 & 0.85 & 0.64 & 0.81 & 0.86 & 0.63 & 0.75 & 0.81 & 0.66 & 0.81 & 0.85 & \textbf{0.71} & 0.73 & 0.74 \\
 & Sciences & 0.60& 0.63 & 0.59 & 0.67 & 0.76 & 0.77 & 0.67 & 0.76 & 0.78 & 0.66 & 0.74 & 0.72 & \textbf{0.68} & 0.76 & 0.77 & 0.67 & 0.72 & 0.67 \\
 & Social sciences & 0.56 & 0.59 & 0.59 & 0.65 & 0.76 & 0.76 & 0.66 & 0.76 & 0.76 & \textbf{0.67} & 0.73 & 0.73 & 0.66 & 0.76 & 0.76 & 0.66 & 0.72 & 0.65 \\
 & Sports and recreation & 0.60& 0.64 & 0.68 & \textbf{0.69} & 0.79 & 0.83 & 0.68 & 0.81 & 0.83 & 0.67 & 0.74 & 0.78 & 0.68 & 0.80& 0.83 & 0.68 & 0.72 & 0.69 \\ \midrule
\multirow{6}{*}{\textbf{LL-70B-Gen}} & Arts and humanity & 0.58 & 0.59 & 0.69 & 0.24 & 0.52 & 0.59 & 0.30 & 0.58 & 0.68 & 0.48 & 0.57 & 0.66 & 0.27 & 0.57 & 0.67 & \textbf{0.70} & 0.81 & 0.88 \\
 & Geography and travel & 0.62 & 0.64 & 0.54 & 0.45 & 0.61 & 0.55 & 0.46 & 0.61 & 0.54 & 0.60& 0.51 & 0.45 & 0.46 & 0.59 & 0.51 & \textbf{0.75} & 0.85 & 0.83 \\
 & Language and communication & 0.56 & 0.62 & 0.68 & 0.26 & 0.51 & 0.59 & 0.30 & 0.53 & 0.63 & 0.46 & 0.53 & 0.64 & 0.28 & 0.51 & 0.61 & \textbf{0.69} & 0.75 & 0.82 \\
 & Sciences & 0.62 & 0.65 & 0.63 & 0.38 & 0.44 & 0.42 & 0.42 & 0.57 & 0.53 & 0.60& 0.58 & 0.58 & 0.39 & 0.54 & 0.50& \textbf{0.73} & 0.80& 0.81 \\
 & Social sciences & 0.59 & 0.59 & 0.55 & 0.40& 0.51 & 0.50& 0.42 & 0.54 & 0.51 & 0.56 & 0.56 & 0.53 & 0.41 & 0.53 & 0.50& \textbf{0.71} & 0.81 & 0.81 \\
 & Sports and recreation & 0.59 & 0.63 & 0.68 & 0.32 & 0.52 & 0.54 & 0.34 & 0.58 & 0.61 & 0.53 & 0.57 & 0.63 & 0.32 & 0.56 & 0.61 & \textbf{0.69} & 0.80& 0.83 \\ \midrule
\multirow{6}{*}{\textbf{ENSB-Gen}} & Arts and humanity & 0.60& 0.61 & 0.80& 0.76 & 0.80& 0.90& 0.78 & 0.82 & 0.91 & 0.66 & 0.71 & 0.86 & 0.78 & 0.81 & 0.91 & \textbf{0.84} & 0.90& 0.96 \\
 & Geography and travel & 0.59 & 0.64 & 0.67 & 0.69 & 0.76 & 0.76 & 0.70& 0.79 & 0.78 & 0.65 & 0.72 & 0.73 & 0.71 & 0.78 & 0.78 & \textbf{0.78} & 0.91 & 0.92 \\
 & Language and communication & 0.57 & 0.61 & 0.81 & 0.68 & 0.75 & 0.88 & 0.67 & 0.77 & 0.89 & 0.62 & 0.68 & 0.84 & 0.69 & 0.76 & 0.89 & \textbf{0.79} & 0.86 & 0.94 \\
 & Sciences & 0.60& 0.64 & 0.73 & 0.75 & 0.80& 0.85 & 0.75 & 0.82 & 0.86 & 0.65 & 0.73 & 0.80& \textbf{0.76} & 0.81 & 0.85 & \textbf{0.76} & 0.87 & 0.92 \\
 & Social sciences & 0.56 & 0.62 & 0.75 & 0.73 & 0.78 & 0.85 & 0.73 & 0.79 & 0.86 & 0.64 & 0.71 & 0.81 & 0.73 & 0.79 & 0.86 & \textbf{0.78} & 0.89 & 0.94 \\
 & Sports and recreation & 0.62 & 0.70& 0.86 & 0.74 & 0.79 & 0.89 & 0.74 & 0.81 & 0.90& 0.68 & 0.75 & 0.88 & 0.76 & 0.81 & 0.91 & \textbf{0.81} & 0.90& 0.96 \\ \midrule
\multirow{6}{*}{\textbf{HA-Test}} & Arts and humanity & 0.52 & 0.50& 0.66 & 0.17 & 0.50& 0.66 & 0.52 & 0.50& 0.66 & 0.66 & 0.50& 0.66 & 0.52 & 0.50& 0.66 & \textbf{0.68} & 0.79 & 0.90\\
 & Geography and travel & 0.23 & 0.50& 0.40& 0.45 & 0.50& 0.40& 0.23 & 0.50& 0.40& 0.40& 0.50& 0.40& 0.23 & 0.50& 0.40& \textbf{0.76} & 0.84 & 0.82 \\
 & Language and communication & 0.50& 0.50& 0.64 & 0.19 & 0.50& 0.64 & 0.50& 0.50& 0.64 & 0.64 & 0.50& 0.64 & 0.50& 0.50& 0.64 & \textbf{0.73} & 0.80& 0.87 \\
 & Sciences & 0.35 & 0.50& 0.52 & 0.32 & 0.50& 0.52 & 0.35 & 0.50& 0.52 & 0.52 & 0.50& 0.52 & 0.35 & 0.50& 0.52 & \textbf{0.72} & 0.81 & 0.83 \\
 & Social sciences & 0.35 & 0.50& 0.51 & 0.32 & 0.50& 0.51 & 0.35 & 0.50& 0.51 & 0.51 & 0.50& 0.51 & 0.35 & 0.50& 0.51 & \textbf{0.75} & 0.83 & 0.85 \\
 & Sports and recreation & 0.42 & 0.50& 0.58 & 0.25 & 0.50& 0.58 & 0.42 & 0.50& 0.58 & 0.58 & 0.50& 0.58 & 0.42 & 0.50& 0.58 & \textbf{0.71} & 0.83 & 0.87 \\ \toprule
\end{tabular}}
\caption{Hallucination detection with BERT classifier results for various models trained on labels obtained from Exact-match based approach on Jeopardy test sets. The best result is highlighted in \textbf{bold}.}
\label{tab:string_jeopardy_bert}
\end{table*}

%% file: appendix_tables/table_string_kaggle_BERT_appendix.tex
\begin{table*}
\centering\footnotesize
\resizebox{\textwidth}{!}{%
\begin{tabular}{ll|ccc|ccc|ccc|ccc|ccc|ccc} \\ \toprule
\multirow{2}{*}{\textbf{Test set}} & \multirow{2}{*}{\textbf{Sub-category}} & \multicolumn{3}{c}{\textbf{QR}} & \multicolumn{3}{c}{\textbf{RR}} & \multicolumn{3}{c}{\textbf{EC-EC}} & \multicolumn{3}{c}{\textbf{CC}} & \multicolumn{3}{c}{\textbf{QR-RR}} & \multicolumn{3}{c}{\textbf{q-r+Q-R+R-R}} \\ \cmidrule{3-20}
 &  & \textbf{F1} &\textbf{ AUC} & \textbf{B-ACC} & \textbf{F1} & \textbf{AUC} &\textbf{ B-ACC} & \textbf{F1} & \textbf{AUC} &\textbf{ B-ACC} & \textbf{F1} &\textbf{ AUC} &\textbf{ B-ACC} &\textbf{ F1} & \textbf{AUC} &\textbf{ B-ACC} & \textbf{F1} & \textbf{AUC} & \textbf{B-ACC} \\ \midrule
\multirow{4}{*}{\textbf{TL-1.1B-Gen}} & SciQ & 0.33 & 0.55 & 0.85 & \textbf{0.74} & 0.65 & 0.87 & 0.68 & 0.68 & 0.89 & 0.48 & 0.58 & 0.86 & 0.63 & 0.66 & 0.88 & 0.54 & 0.73 & 0.91 \\
 & MathQA & 0.73 & 0.54 & 0.95 & 0.85 & 0.51 & 0.95 & \textbf{0.87} & 0.55 & 0.95 & 0.79 & 0.55 & 0.95 & \textbf{0.87} & 0.54 & 0.95 & 0.90& 0.83 & 0.99 \\
 & MathQSA & 0.66 & 0.54 & 0.92 & 0.80& 0.52 & 0.93 & \textbf{0.82} & 0.56 & 0.93 & 0.72 & 0.55 & 0.93 & \textbf{0.82} & 0.54 & 0.93 & 0.80& 0.72 & 0.97 \\ 
 & GK & 0.33 & 0.58 & 0.76 & \textbf{0.73} & 0.74 & 0.83 & \textbf{0.73} & 0.77 & 0.86 & 0.64 & 0.68 & 0.81 & 0.68 & 0.77 & 0.85 & 0.55 & 0.77 & 0.86 \\ \midrule
\multirow{4}{*}{\textbf{PHI-3.5B-Gen}} & SciQ & 0.50& 0.46 & 0.41 & \textbf{0.59} & 0.64 & 0.54 & 0.56 & 0.59 & 0.50& 0.52 & 0.52 & 0.44 & 0.55 & 0.58 & 0.49 & 0.50& 0.71 & 0.65 \\
 & MathQA & 0.69 & 0.54 & 0.95 & 0.54 & 0.55 & 0.95 & 0.64 & 0.57 & 0.95 & 0.70& 0.57 & 0.95 & 0.68 & 0.58 & 0.95 & \textbf{0.91} & 0.89 & 0.99 \\
 & MathQSA & 0.63 & 0.47 & 0.90& 0.62 & 0.52 & 0.91 & 0.64 & 0.52 & 0.92 & 0.67 & 0.53 & 0.92 & 0.66 & 0.52 & 0.92 & \textbf{0.78} & 0.70& 0.96 \\
 & GK & 0.65 & 0.58 & 0.38 & 0.68 & 0.62 & 0.44 & \textbf{0.70} & 0.66 & 0.50& 0.68 & 0.63 & 0.46 & 0.69 & 0.70& 0.51 & 0.64 & 0.66 & 0.50\\   \midrule
\multirow{4}{*}{\textbf{LL-8B-Gen}} & SciQ & 0.48 & 0.46 & 0.46 & \textbf{0.58} & 0.74 & 0.70& 0.55 & 0.71 & 0.67 & 0.55 & 0.68 & 0.62 & 0.55 & 0.68 & 0.65 & 0.44 & 0.75 & 0.72 \\
 & MathQA & 0.72 & 0.54 & 0.91 & 0.79 & 0.64 & 0.93 & 0.80& 0.65 & 0.94 & 0.74 & 0.61 & 0.93 & \textbf{0.82} & 0.64 & 0.93 & \textbf{0.82} & 0.81 & 0.97 \\
 & MathQSA & 0.62 & 0.54 & 0.79 & 0.71 & 0.70& 0.87 & 0.71 & 0.71 & 0.86 & 0.62 & 0.60& 0.81 & \textbf{0.74} & 0.70& 0.86 & 0.58 & 0.68 & 0.85 \\
 & GK & 0.58 & 0.60& 0.36 & 0.64 & 0.74 & 0.51 & 0.66 & 0.72 & 0.51 & \textbf{0.68} & 0.70& 0.49 & 0.67 & 0.68 & 0.50& 0.60& 0.66 & 0.55 \\  \midrule
\multirow{4}{*}{\textbf{MST-7B-Gen}} & SciQ & 0.40& 0.50& 0.56 & 0.42 & 0.50& 0.56 & 0.41 & 0.50& 0.55 & 0.40& 0.49 & 0.54 & 0.40& 0.50& 0.56 & \textbf{0.62} & 0.70& 0.74 \\
 & MathQA & \textbf{0.91} & 0.45 & 0.93 & \textbf{0.91} & 0.54 & 0.95 & \textbf{0.91} & 0.52 & 0.94 & \textbf{0.91} & 0.48 & 0.94 & \textbf{0.91} & 0.49 & 0.94 & \textbf{0.91} & 0.86 & 0.99 \\
 & MathQSA & \textbf{0.86} & 0.43 & 0.88 & 0.85 & 0.58 & 0.92 & 0.85 & 0.56 & 0.92 & \textbf{0.86} & 0.45 & 0.89 & \textbf{0.86} & 0.48 & 0.90& 0.84 & 0.75 & 0.96 \\
 & GK & 0.25 & 0.45 & 0.37 & 0.32 & 0.56 & 0.48 & 0.28 & 0.50& 0.41 & 0.26 & 0.46 & 0.39 & 0.25 & 0.45 & 0.38 & \textbf{0.61} & 0.62 & 0.56 \\  \midrule
\multirow{4}{*}{\textbf{GM-7B-Gen}} & SciQ & \textbf{0.49} & 0.50& 0.57 & 0.47 & 0.67 & 0.70& 0.45 & 0.65 & 0.67 & 0.46 & 0.60& 0.63 & 0.48 & 0.62 & 0.66 & 0.35 & 0.71 & 0.74 \\
 & MathQA & 0.64 & 0.45 & 0.89 & 0.60& 0.58 & 0.92 & 0.57 & 0.56 & 0.91 & 0.59 & 0.55 & 0.92 & 0.65 & 0.56 & 0.92 & \textbf{0.76} & 0.72 & 0.96 \\
 & MathQSA & 0.63 & 0.53 & 0.83 & 0.51 & 0.68 & 0.88 & 0.52 & 0.67 & 0.88 & 0.57 & 0.63 & 0.87 & \textbf{0.64} & 0.67 & 0.89 & 0.53 & 0.65 & 0.87 \\
 & GK & 0.54 & 0.59 & 0.52 & 0.53 & 0.68 & 0.68 & 0.48 & 0.71 & 0.66 & \textbf{0.59} & 0.68 & 0.66 & 0.48 & 0.70& 0.65 & 0.44 & 0.74 & 0.72 \\  \midrule
\multirow{4}{*}{\textbf{LL-70B-Gen}} & SciQ & 0.50& 0.46 & 0.45 & 0.61 & 0.75 & 0.71 & 0.58 & 0.71 & 0.67 & 0.56 & 0.64 & 0.58 & \textbf{0.59} & 0.67 & 0.65 & 0.46 & 0.74 & 0.72 \\
 & MathQA & 0.60& 0.50& 0.89 & \textbf{0.85} & 0.73 & 0.94 & 0.82 & 0.72 & 0.94 & 0.66 & 0.58 & 0.91 & 0.82 & 0.72 & 0.94 & 0.82 & 0.74 & 0.95 \\
 & MathQSA & 0.61 & 0.53 & 0.78 & \textbf{0.78} & 0.77 & 0.89 & 0.76 & 0.76 & 0.89 & 0.61 & 0.60& 0.81 & \textbf{0.78} & 0.76 & 0.89 & 0.64 & 0.69 & 0.85 \\
 & GK & 0.59 & 0.60& 0.36 & \textbf{0.68} & 0.76 & 0.58 & 0.66 & 0.72 & 0.53 & \textbf{0.68} & 0.73 & 0.55 & 0.66 & 0.75 & 0.54 & 0.58 & 0.73 & 0.60\\  \midrule
\multirow{4}{*}{\textbf{ENSB-Gen}} & SciQ & 0.41 & 0.44 & 0.53 & \textbf{0.65} & 0.74 & 0.75 & 0.60& 0.71 & 0.73 & 0.48 & 0.59 & 0.62 & 0.59 & 0.71 & 0.72 & 0.52 & 0.79 & 0.81 \\
 & MathQA & 0.70& 0.54 & 0.94 & 0.78 & 0.55 & 0.94 & 0.80& 0.59 & 0.95 & 0.74 & 0.58 & 0.95 & 0.81 & 0.59 & 0.95 & \textbf{0.88} & 0.85 & 0.99 \\
 & MathQSA & 0.64 & 0.54 & 0.90& 0.73 & 0.52 & 0.90& \textbf{0.75} & 0.55 & 0.91 & 0.66 & 0.52 & 0.90& 0.74 & 0.55 & 0.91 & \textbf{0.75} & 0.72 & 0.95 \\
 & GK & 0.49 & 0.52 & 0.47 & 0.70& 0.77 & 0.72 & \textbf{0.71} & 0.77 & 0.74 & 0.64 & 0.71 & 0.66 & 0.69 & 0.76 & 0.72 & 0.61 & 0.75 & 0.76 \\  \midrule
\multirow{4}{*}{\textbf{HA-Test}} & SciQ & 0.30 & 0.50& 0.47 & 0.37 & 0.50& 0.47 & 0.37 & 0.50& 0.47 & 0.30 & 0.50& 0.47 & 0.37 & 0.50& 0.47 & \textbf{0.43} & 0.74 & 0.72 \\
 & MathQA & \textbf{0.95} & 0.50& 0.97 &0\phantom{0}\phantom{0}& 0.50& 0.97 &0\phantom{0}\phantom{0}& 0.50& 0.97 & \textbf{0.95} & 0.50& 0.97 &0\phantom{0}\phantom{0}& 0.50& 0.97 & 0.86 & 0.80& 0.99 \\
 & MathQSA & \textbf{0.93} & 0.50& 0.96 &0\phantom{0}\phantom{0}& 0.50& 0.96 &0\phantom{0}\phantom{0}& 0.50& 0.96 & \textbf{0.93} & 0.50& 0.96 &0\phantom{0}\phantom{0}& 0.50& 0.96 & 0.75 & 0.75 & 0.98 \\
 & GK & 0.16 & 0.49 & 0.32 & 0.55 & 0.49 & 0.32 & 0.55 & 0.49 & 0.32 & 0.16 & 0.50& 0.32 & 0.55 & 0.49 & 0.32 & \textbf{0.59} & 0.74 & 0.60\\ \toprule
\end{tabular}}
\caption{Hallucination detection with BERT classifier results for various models trained on labels obtained from Exact-match based approach on Kaggle test sets. The best result highlighted in \textbf{bold}.}
\label{tab:string_kaggle_bert}
\end{table*}

%% file: appendix_tables/table_qwen_jeopardy_BERT_appendix.tex
\begin{table*}
\centering\footnotesize
\resizebox{\textwidth}{!}{%
\begin{tabular}{@{}ll|ccc|ccc|ccc|ccc|ccc|ccc@{}}
\toprule
\multirow{2}{*}{\textbf{Test set}} & \multirow{2}{*}{\textbf{Sub-category}}  & \multicolumn{3}{c}{\textbf{QR}} & \multicolumn{3}{c}{\textbf{RR}} & \multicolumn{3}{c}{\textbf{EC-EC}} & \multicolumn{3}{c}{\textbf{CC}} & \multicolumn{3}{c}{\textbf{QR-RR}} & \multicolumn{3}{c}{\textbf{q-r+Q-R+R-R}} \\ \cmidrule{3-20}
 & & \textbf{F1} &\textbf{ AUC }& \textbf{B-ACC} & \textbf{F1} & \textbf{AUC} & \textbf{B-ACC} & \textbf{F1} & \textbf{AUC} & \textbf{B-ACC} & \textbf{F1} & \textbf{AUC} & \textbf{B-ACC} & \textbf{F1} & \textbf{AUC} & \textbf{B-ACC} & \textbf{F1} & \textbf{AUC} & \textbf{B-ACC} \\ \midrule
 & Arts and humanity & 0.57 & 0.66 & 0.92 &0.80& 0.69 & 0.92 & 0.81 & 0.77 & 0.95 & 0.57 & 0.63 & 0.92 &0.80& 0.77 & 0.95 & \textbf{0.85} & 0.85 & 0.97 \\
& Geography and travel & 0.56 &0.70& 0.87 & 0.73 & 0.72 & 0.87 & 0.77 & 0.81 & 0.92 & 0.54 & 0.64 & 0.85 & 0.76 &0.80& 0.92 & \textbf{0.79} & 0.84 & 0.94 \\
& Language and communication & 0.51 & 0.65 & 0.93 & 0.77 & 0.65 & 0.92 & 0.76 & 0.74 & 0.95 & 0.47 & 0.59 & 0.91 & 0.75 & 0.74 & 0.95 & \textbf{0.85} & 0.82 & 0.97 \\
& Sciences & 0.55 & 0.67 &0.90& 0.77 & 0.71 &0.90& 0.78 & 0.79 & 0.93 & 0.49 &0.60& 0.88 & 0.77 & 0.79 & 0.93 & \textbf{0.79} & 0.83 & 0.95 \\
& Social sciences & 0.55 & 0.69 &0.90& 0.75 &0.70&0.90& 0.79 &0.80& 0.94 & 0.56 & 0.62 & 0.88 & 0.79 &0.80& 0.94 & \textbf{0.82} & 0.86 & 0.96 \\
\multirow{-6}{*}{\textbf{TL-1.1B-Gen}} & Sports and recreation & 0.55 & 0.65 & 0.91 & 0.81 & 0.73 & 0.92 & 0.82 & 0.79 & 0.95 & 0.57 & 0.59 &0.90& 0.81 &0.80& 0.95 & \textbf{0.84} & 0.84 & 0.96 \\ \midrule
& Arts and humanity &0.50&0.60& 0.73 & 0.71 &0.80& 0.87 & 0.71 & 0.81 & 0.88 &0.60& 0.69 & 0.81 & 0.71 & 0.81 & 0.88 & \textbf{0.80} & 0.88 & 0.92 \\
& Geography and travel & 0.66 & 0.67 & 0.45 & 0.75 & 0.81 & 0.65 & 0.75 & 0.83 & 0.67 & 0.73 & 0.71 & 0.53 & 0.75 & 0.82 & 0.67 & \textbf{0.82} & 0.87 & 0.76 \\
& Language and communication & 0.49 & 0.62 & 0.65 & 0.62 & 0.76 & 0.78 & 0.61 & 0.77 & 0.79 & 0.49 & 0.63 & 0.66 & 0.61 & 0.76 & 0.78 & \textbf{0.73} & 0.82 & 0.83 \\
& Sciences & 0.66 & 0.62 & 0.47 & 0.79 & 0.85 & 0.74 & 0.79 & 0.85 & 0.75 &0.70&0.70& 0.56 & 0.79 & 0.84 & 0.74 & \textbf{0.83} & 0.89 & 0.81 \\
& Social sciences &0.60& 0.64 & 0.53 & 0.75 & 0.82 & 0.76 & 0.75 & 0.83 & 0.77 & 0.68 & 0.69 & 0.63 & 0.76 & 0.83 & 0.77 & \textbf{0.80} & 0.89 & 0.83 \\
\multirow{-6}{*}{\textbf{PHI-3.5B-Gen}} & Sports and recreation & 0.45 &0.60& 0.67 & 0.67 & 0.75 & 0.82 & 0.66 & 0.77 & 0.84 & 0.58 & 0.65 & 0.76 & 0.66 & 0.77 & 0.84 & \textbf{0.75} & 0.83 & 0.87 \\ \midrule
& Arts and humanity & 0.61 & 0.65 & 0.54 & 0.82 & 0.89 & 0.86 & 0.82 & 0.89 & 0.86 & 0.75 & 0.83 & 0.78 & 0.82 & 0.89 & 0.85 & \textbf{0.84} &0.90& 0.88 \\
& Geography and travel & 0.79 &0.70& 0.35 & 0.88 & 0.93 & 0.74 & 0.88 & 0.93 & 0.74 & 0.88 & 0.88 & 0.68 & 0.88 & 0.93 & 0.75 & \textbf{0.90} & 0.92 & 0.78 \\
& Language and communication &0.60& 0.67 & 0.63 & 0.73 & 0.86 & 0.83 & 0.73 & 0.86 & 0.84 & 0.66 & 0.78 & 0.74 & 0.73 & 0.86 & 0.83 & \textbf{0.76} & 0.85 & 0.84 \\
& Sciences &0.70& 0.66 & 0.44 & 0.84 &0.90& 0.79 & 0.84 & 0.89 &0.80& 0.79 & 0.84 &0.70& 0.84 & 0.89 & 0.79 & \textbf{0.85} & 0.89 & 0.81 \\
& Social sciences & 0.74 & 0.72 &0.50& 0.85 & 0.92 &0.80& 0.86 & 0.92 & 0.81 & 0.82 & 0.86 & 0.71 & 0.86 & 0.93 & 0.81 & \textbf{0.87} & 0.92 & 0.83 \\
\multirow{-6}{*}{\textbf{LL-8B-Gen}} & Sports and recreation & 0.68 & 0.71 & 0.61 &0.80& 0.89 & 0.84 & 0.81 & 0.89 & 0.83 & 0.76 & 0.84 & 0.77 & 0.81 & 0.89 & 0.82 & \textbf{0.83} &0.90& 0.86 \\ \midrule
& Arts and humanity & 0.53 & 0.62 & 0.79 & 0.72 & 0.74 & 0.85 & 0.73 & 0.77 & 0.87 & 0.51 & 0.63 & 0.81 & 0.72 & 0.77 & 0.87 & \textbf{0.82} &0.90& 0.95 \\
& Geography and travel & 0.63 &0.70&0.70& 0.73 & 0.79 & 0.73 & 0.76 & 0.83 &0.80&0.60& 0.67 & 0.69 & 0.74 & 0.82 &0.80& \textbf{0.80} & 0.89 & 0.88 \\
& Language and communication & 0.49 & 0.61 & 0.76 & 0.66 & 0.74 & 0.83 & 0.66 & 0.76 & 0.85 & 0.44 & 0.62 & 0.77 & 0.64 & 0.75 & 0.84 & \textbf{0.79} & 0.87 & 0.93 \\
& Sciences & 0.59 & 0.66 & 0.71 & 0.71 & 0.78 & 0.78 & 0.74 & 0.82 & 0.83 & 0.54 & 0.62 &0.70& 0.74 & 0.82 & 0.83 & \textbf{0.82} &0.90& 0.92 \\
& Social sciences & 0.58 & 0.66 & 0.74 & 0.69 & 0.75 & 0.77 & 0.73 &0.80& 0.83 & 0.56 & 0.66 & 0.74 & 0.72 &0.80& 0.82 & \textbf{0.82} & 0.91 & 0.93 \\
\multirow{-6}{*}{\textbf{MST-7B-Gen}} & Sports and recreation & 0.54 & 0.64 & 0.78 & 0.71 & 0.75 & 0.83 & 0.72 & 0.79 & 0.87 & 0.51 & 0.64 & 0.79 & 0.71 & 0.79 & 0.87 & \textbf{0.81} & 0.89 & 0.94 \\ \midrule
& Arts and humanity & 0.53 & 0.64 & 0.73 & 0.58 & 0.69 & 0.79 & 0.59 &0.70& 0.79 & 0.59 & 0.69 & 0.79 & 0.58 & 0.71 & 0.79 & \textbf{0.70} & 0.77 & 0.86 \\
& Geography and travel &0.60& 0.58 & 0.45 & 0.65 & 0.68 & 0.55 & 0.64 & 0.69 & 0.56 & 0.64 & 0.65 & 0.52 & 0.64 &0.70& 0.56 & \textbf{0.70} & 0.76 & 0.66 \\
& Language and communication & 0.49 & 0.62 & 0.71 & 0.55 & 0.73 & 0.79 & 0.54 & 0.74 & 0.79 & 0.52 & 0.71 & 0.77 & 0.54 & 0.74 & 0.79 & \textbf{0.67} & 0.76 & 0.83 \\
& Sciences &0.60& 0.62 & 0.49 & 0.65 & 0.71 & 0.58 & 0.65 & 0.71 & 0.58 & 0.65 &0.70& 0.56 & 0.65 & 0.71 & 0.58 & \textbf{0.71} & 0.79 &0.70\\
& Social sciences & 0.59 & 0.63 & 0.55 & 0.65 &0.70& 0.65 & 0.65 &0.70& 0.65 & 0.64 & 0.67 & 0.61 & 0.65 &0.70& 0.65 & \textbf{0.72} & 0.79 & 0.75 \\
\multirow{-6}{*}{\textbf{GM-7B-Gen}} & Sports and recreation & 0.54 & 0.64 & 0.66 & 0.59 & 0.71 & 0.68 & 0.61 & 0.72 &0.70& 0.57 & 0.68 & 0.67 &0.60& 0.72 &0.70& \textbf{0.68} & 0.78 & 0.78 \\ \midrule
& Arts and humanity & 0.58 & 0.62 &0.60& 0.36 & 0.47 & 0.47 & 0.36 & 0.57 & 0.55 & 0.44 & 0.57 & 0.56 & 0.36 & 0.54 & 0.54 & \textbf{0.68} & 0.81 & 0.83 \\
& Geography and travel & 0.75 & 0.76 & 0.47 & 0.68 & 0.49 & 0.23 & 0.68 & 0.58 & 0.28 & 0.74 & 0.57 & 0.36 & 0.68 & 0.54 & 0.26 & \textbf{0.76} & 0.72 & 0.54 \\
& Language and communication & 0.56 & 0.61 & 0.54 & 0.43 & 0.42 & 0.37 & 0.43 & 0.46 & 0.39 &0.50& 0.51 & 0.48 & 0.43 & 0.46 & 0.39 & \textbf{0.67} & 0.72 & 0.68 \\
& Sciences & 0.69 & 0.69 & 0.52 & 0.59 & 0.46 & 0.26 & 0.59 & 0.59 & 0.33 & 0.66 & 0.61 & 0.44 & 0.59 & 0.51 & 0.29 & \textbf{0.75} &0.80& 0.69 \\
& Social sciences & 0.69 & 0.66 & 0.44 & 0.61 & 0.53 & 0.32 & 0.61 & 0.56 & 0.35 & 0.66 & 0.62 & 0.42 & 0.61 & 0.55 & 0.34 & \textbf{0.77} & 0.85 & 0.76 \\
\multirow{-6}{*}{\textbf{LL-70B-Gen}} & Sports and recreation & 0.66 & 0.68 & 0.64 & 0.44 &0.50&0.40& 0.44 & 0.58 & 0.47 & 0.58 & 0.62 & 0.62 & 0.44 & 0.57 & 0.45 & \textbf{0.69} & 0.78 & 0.77 \\ \midrule
& Arts and humanity & 0.54 & 0.62 & 0.74 & 0.78 & 0.84 & 0.89 & 0.77 & 0.85 &0.90&0.60& 0.73 & 0.82 & 0.77 & 0.85 &0.90& \textbf{0.83} &0.90& 0.94 \\
& Geography and travel & 0.66 & 0.69 & 0.64 & 0.78 & 0.87 & 0.79 & 0.79 & 0.88 & 0.82 & 0.69 & 0.77 &0.70& 0.79 & 0.88 & 0.81 & \textbf{0.82} & 0.92 & 0.88 \\
& Language and communication & 0.51 & 0.63 & 0.75 & 0.68 & 0.79 & 0.85 & 0.67 & 0.81 & 0.87 &0.50& 0.67 & 0.77 & 0.66 &0.80& 0.86 & \textbf{0.77} & 0.86 & 0.91 \\
& Sciences & 0.61 & 0.67 & 0.66 & 0.79 & 0.87 & 0.84 &0.80& 0.88 & 0.86 & 0.63 & 0.75 & 0.73 & 0.79 & 0.88 & 0.85 & \textbf{0.83} & 0.92 & 0.91 \\
& Social sciences & 0.61 & 0.68 &0.70& 0.78 & 0.86 & 0.84 & 0.79 & 0.88 & 0.87 & 0.62 & 0.75 & 0.76 & 0.79 & 0.87 & 0.86 & \textbf{0.84} & 0.93 & 0.93 \\
\multirow{-6}{*}{\textbf{ENSB-Gen}} & Sports and recreation & 0.56 & 0.68 & 0.77 & 0.75 & 0.82 & 0.87 & 0.74 & 0.84 & 0.89 & 0.63 & 0.74 & 0.82 & 0.74 & 0.84 & 0.89 & \textbf{0.81} & 0.88 & 0.93 \\ \midrule
& Arts and humanity & 0.52 &0.50& 0.66 & 0.17 &0.50& 0.66 & 0.17 &0.50& 0.66 & 0.52 &0.50& 0.66 & 0.52 &0.50& 0.66 & \textbf{0.61} & 0.81 &0.90\\
& Geography and travel & 0.23 &0.50&0.40& 0.45 &0.50&0.40& 0.45 &0.50&0.40& 0.23 &0.50&0.40& 0.23 &0.50&0.40& \textbf{0.70} & 0.85 & 0.82 \\
& Language and communication &0.50&0.50& 0.64 & 0.19 &0.50& 0.64 & 0.19 &0.50& 0.64 &0.50&0.50& 0.64 &0.50&0.50& 0.64 & \textbf{0.65} & 0.76 & 0.85 \\
& Sciences & 0.35 &0.50& 0.52 & 0.32 &0.50& 0.52 & 0.32 &0.50& 0.52 & 0.35 &0.50& 0.52 & 0.35 &0.50& 0.52 & \textbf{0.64} & 0.82 & 0.83 \\
& Social sciences & 0.35 &0.50& 0.51 & 0.32 &0.50& 0.51 & 0.32 &0.50& 0.51 & 0.35 &0.50& 0.51 & 0.35 &0.50& 0.51 & \textbf{0.69} & 0.87 & 0.89 \\
\multirow{-6}{*}{\textbf{HA-Test}}& Sports and recreation & 0.42 &0.50& 0.58 & 0.25 &0.50& 0.58 & 0.25 &0.50& 0.58 & 0.42 &0.50& 0.58 & 0.42 &0.50& 0.58 & \textbf{0.62} & 0.76 & 0.83 \\ \bottomrule
\end{tabular}}
\caption{Hallucination detection with BERT classifier results for various models trained on labels obtained from the LLM-based approach on the Jeopardy test sets.}
\label{tab:qwen_jeopardy_bert}
\end{table*}

%% file: appendix_tables/table_qwen_kaggle_BERT_appendix.tex
\begin{table*}[b]
\centering\footnotesize
\resizebox{\textwidth}{!}{%
\begin{tabular}{@{}ll|ccc|ccc|ccc|ccc|ccc|ccc@{}} \\ \toprule
\multirow{2}{*}{\textbf{Test set}} & \multirow{2}{*}{\textbf{Sub-category}}  & \multicolumn{3}{c}{\textbf{QR}} & \multicolumn{3}{c}{\textbf{RR}} & \multicolumn{3}{c}{\textbf{EC-EC}} & \multicolumn{3}{c}{\textbf{CC}} & \multicolumn{3}{c}{\textbf{QR-RR}} & \multicolumn{3}{c}{\textbf{q-r+Q-R+R-R}} \\ \cmidrule{3-20}
&  & \textbf{F1} & \textbf{AUC} & \textbf{B-ACC} & \textbf{F1} & \textbf{AUC} & \textbf{B-ACC} & \textbf{F1} & \textbf{AUC} & \textbf{B-ACC} & \textbf{F1} & \textbf{AUC} & \textbf{B-ACC} & \textbf{F1} & \textbf{AUC} & \textbf{B-ACC} & \textbf{F1} & \textbf{AUC} & \textbf{B-ACC} \\ \midrule
\multirow{4}{*}{\textbf{TL-1.1B-Gen}} & SciQ & 0.45 & 0.60& 0.80& \textbf{0.72} & 0.68 & 0.82 & \textbf{0.72} & 0.73 & 0.87 & 0.54 & 0.59 & 0.81 & \textbf{0.72} & 0.73 & 0.86 & 0.66 & 0.75 & 0.89 \\
 & MathQA & 0.84 & 0.67 &1\phantom{0}\phantom{0}& 0.93 & 0.47 & 0.99 & 0.95 & 0.62 &1\phantom{0}\phantom{0}& 0.89 & 0.65 &1\phantom{0}\phantom{0}& 0.97 & 0.62 &1\phantom{0}\phantom{0}& \textbf{0.99} & 0.74 &1\phantom{0}\phantom{0}\\
 & MathQSA & 0.79 & 0.58 & 0.99 & 0.92 & 0.54 & 0.99 & 0.94 & 0.59 & 0.99 & 0.85 & 0.59 & 0.99 & 0.95 & 0.58 & 0.99 & \textbf{0.97} & 0.69 & 0.99 \\
 & GK & 0.45 & 0.59 & 0.72 & 0.72 & 0.72 & 0.72 & 0.77 & 0.86 & 0.90& 0.68 & 0.72 & 0.82 & 0.77 & 0.85 & 0.89 & \textbf{0.80} & 0.86 & 0.89 \\ \midrule
\multirow{4}{*}{\textbf{PHI-3.5B-Gen}} & SciQ & 0.65 & 0.52 & 0.24 & 0.74 & 0.71 & 0.41 & \textbf{0.75} & 0.68 & 0.39 & 0.64 & 0.55 & 0.25 & 0.73 & 0.68 & 0.39 & 0.74 & 0.72 & 0.45 \\
 & MathQA & 0.66 & 0.56 & 0.83 & 0.55 & 0.60& 0.85 & 0.61 & 0.62 & 0.86 & 0.68 & 0.58 & 0.84 & 0.65 & 0.62 & 0.86 & \textbf{0.80} & 0.75 & 0.91 \\
 & MathQSA & 0.60& 0.54 & 0.77 & 0.65 & 0.65 & 0.83 & 0.65 & 0.66 & 0.83 & 0.62 & 0.55 & 0.77 & 0.67 & 0.65 & 0.82 & \textbf{0.73} & 0.72 & 0.86 \\
 & GK & 0.72 & 0.46 & 0.15 & 0.77 & 0.64 & 0.40& 0.80& 0.64 & 0.34 & 0.70& 0.53 & 0.18 & 0.80& 0.63 & 0.39 & \textbf{0.82} & 0.64 & 0.37 \\ \midrule
\multirow{4}{*}{\textbf{LL-8B-Gen}} & SciQ & 0.56 & 0.54 & 0.31 & \textbf{0.75} & 0.76 & 0.54 & \textbf{0.75} & 0.76 & 0.55 & 0.70& 0.70& 0.44 & \textbf{0.75} & 0.75 & 0.54 & 0.69 & 0.76 & 0.57 \\
 & MathQA & 0.71 & 0.56 & 0.84 & 0.77 & 0.67 & 0.87 & 0.77 & 0.69 & 0.89 & 0.72 & 0.60& 0.86 & 0.78 & 0.69 & 0.89 & \textbf{0.81} & 0.78 & 0.92 \\
 & MathQSA & 0.61 & 0.54 & 0.73 & 0.72 & 0.71 & 0.81 & 0.73 & 0.74 & 0.84 & 0.64 & 0.60& 0.76 & 0.73 & 0.73 & 0.84 & \textbf{0.76} & 0.77 & 0.86 \\
 & GK & 0.70& 0.58 & 0.21 & 0.81 & 0.72 & 0.40& 0.81 & 0.72 & 0.40& \textbf{0.82} & 0.68 & 0.38 & 0.81 & 0.72 & 0.37 & 0.81 & 0.68 & 0.41 \\ \midrule
\multirow{4}{*}{\textbf{MST-7B-Gen}} & SciQ & 0.20 & 0.50& 0.37 & 0.22 & 0.51 & 0.38 & 0.20 & 0.51 & 0.37 & 0.20 & 0.48 & 0.36 & 0.20 & 0.51 & 0.37 & \textbf{0.25} & 0.62 & 0.50\\
 & MathQA & \textbf{0.90} & 0.50& 0.93 & \textbf{0.90} & 0.52 & 0.94 & \textbf{0.90} & 0.52 & 0.93 & \textbf{0.90} & 0.47 & 0.92 & \textbf{0.90} & 0.51 & 0.93 & \textbf{0.90} & 0.76 & 0.98 \\
 & MathQSA & \textbf{0.87} & 0.52 & 0.92 & \textbf{0.87} & 0.51 & 0.91 & \textbf{0.87} & 0.52 & 0.92 & \textbf{0.87} & 0.47 & 0.90& \textbf{0.87} & 0.49 & 0.91 & \textbf{0.87} & 0.68 & 0.95 \\ 
 & GK & 0.16 & 0.53 & 0.33 & 0.19 & 0.54 & 0.34 & 0.16 & 0.60& 0.37 & 0.16 & 0.38 & 0.26 & 0.16 & 0.52 & 0.32 & \textbf{0.21} & 0.66 & 0.52 \\ \midrule
\multirow{4}{*}{\textbf{GM-7B-Gen}} & SciQ & 0.55 & 0.55 & 0.39 & \textbf{0.68} & 0.68 & 0.53 & 0.65 & 0.68 & 0.53 & 0.65 & 0.62 & 0.46 & 0.66 & 0.67 & 0.51 & 0.57 & 0.69 & 0.56 \\
 & MathQA & \textbf{0.77} & 0.51 & 0.89 & 0.69 & 0.66 & 0.93 & 0.66 & 0.66 & 0.93 & 0.70& 0.61 & 0.92 & 0.68 & 0.66 & 0.93 & 0.74 & 0.71 & 0.94 \\
 & MathQSA & \textbf{0.73} & 0.54 & 0.83 & 0.64 & 0.72 & 0.91 & 0.61 & 0.73 & 0.91 & 0.65 & 0.64 & 0.88 & 0.64 & 0.71 & 0.91 & 0.65 & 0.71 & 0.90\\
 & GK & \textbf{0.66} & 0.60& 0.40& \textbf{0.66} & 0.65 & 0.54 & 0.64 & 0.66 & 0.53 & \textbf{0.66} & 0.61 & 0.46 & 0.64 & 0.64 & 0.50& 0.62 & 0.68 & 0.58 \\ \midrule
\multirow{4}{*}{\textbf{LL-70B-Gen}} & SciQ & 0.55 & 0.53 & 0.27 & \textbf{0.79} & 0.78 & 0.55 & \textbf{0.79} & 0.77 & 0.56 & 0.70& 0.65 & 0.37 & 0.78 & 0.76 & 0.54 & 0.71 & 0.77 & 0.57 \\
 & MathQA & 0.70& 0.53 & 0.83 & \textbf{0.85} & 0.75 & 0.90& \textbf{0.85} & 0.78 & 0.92 & 0.70& 0.59 & 0.86 & 0.84 & 0.77 & 0.92 & 0.84 & 0.78 & 0.92 \\
 & MathQSA & 0.62 & 0.52 & 0.73 & 0.80& 0.81 & 0.88 & \textbf{0.82} & 0.84 & 0.90& 0.65 & 0.61 & 0.77 & 0.80& 0.82 & 0.89 & 0.81 & 0.81 & 0.89 \\
 & GK & 0.72 & 0.62 & 0.23 & 0.87 & 0.84 & 0.60& \textbf{0.88} & 0.83 & 0.56 & 0.82 & 0.76 & 0.39 & 0.86 & 0.84 & 0.54 & 0.85 & 0.84 & 0.55 \\ \midrule
\multirow{4}{*}{\textbf{ENSB-Gen}} & SciQ & 0.53 & 0.50& 0.42 & \textbf{0.74} & 0.80& 0.68 & 0.73 & 0.79 & 0.70& 0.58 & 0.60& 0.49 & 0.73 & 0.79 & 0.70& 0.70& 0.82 & 0.76 \\
 & MathQA & 0.72 & 0.57 & 0.91 & 0.79 & 0.70& 0.93 & 0.81 & 0.72 & 0.94 & 0.76 & 0.62 & 0.92 & 0.82 & 0.72 & 0.94 & \textbf{0.88} & 0.84 & 0.97 \\
 & MathQSA & 0.69 & 0.54 & 0.89 & 0.79 & 0.67 & 0.91 & 0.79 & 0.71 & 0.93 & 0.72 & 0.57 & 0.90& 0.81 & 0.72 & 0.93 & \textbf{0.85} & 0.81 & 0.96 \\
 & GK & 0.62 & 0.53 & 0.39 & 0.75 & 0.78 & 0.61 & 0.78 & 0.80& 0.72 & 0.71 & 0.71 & 0.61 & \textbf{0.79} & 0.78 & 0.71 & 0.77 & 0.83 & 0.74 \\ \midrule
\multirow{4}{*}{\textbf{HA-Test}} & SciQ & 0.30 & 0.50& 0.47 & 0.37 & 0.50& 0.47 & 0.37 & 0.50& 0.47 & 0.30 & 0.50& 0.47 & 0.37 & 0.50& 0.47 & \textbf{0.44} & 0.71 & 0.70\\
 & MathQA & \textbf{0.95} & 0.50& 0.97 &0\phantom{0}\phantom{0}& 0.50& 0.97 &0\phantom{0}\phantom{0}& 0.50& 0.97 & \textbf{0.95} & 0.50& 0.97 &0\phantom{0}\phantom{0}& 0.50& 0.97 & 0.88 & 0.82 & 0.99 \\
 & MathQSA & \textbf{0.93} & 0.50& 0.96 &0\phantom{0}\phantom{0}& 0.50& 0.96 &0\phantom{0}\phantom{0}& 0.50& 0.96 & \textbf{0.93} & 0.50& 0.96 &0\phantom{0}\phantom{0}& 0.50& 0.96 & 0.75 & 0.79 & 0.98 \\
 & GK & 0.16 & 0.49 & 0.32 & 0.55 & 0.50& 0.32 & 0.55 & 0.51 & 0.33 & 0.16 & 0.49 & 0.32 & 0.55 & 0.49 & 0.32 & \textbf{0.63} & 0.71 & 0.55 \\ \toprule
\end{tabular}}
\caption{Hallucination detection with BERT classifier results for various models trained on labels obtained from LLM-based approach on Kaggle test sets. The best result highlighted in \textbf{bold}.}
\label{tab:qwen_kaggle_bert}
\end{table*}

%% file: datasheet.tex
\subsection{Motivation}

\noindent\textbf{Q:} For what purpose was the dataset created? (Was there a specific task in mind? Was there a specific gap that needed to be filled? Please provide a description.)

\noindent\textbf{A:} This dataset is developed to facilitate research on reference-free hallucination detection in Large Language Models (LLMs). We observe a significant lack of suitable and sufficiently large benchmark datasets spanning multiple domains for reference-free hallucination detection. It will benefit the research community by enabling the development of hallucination detection pipelines and evaluating their robustness using this dataset.
\\
\\
\noindent\textbf{Q:} Who created the dataset (e.g., which team, research group) and on behalf of which entity (e.g., company, institution, organization)?

\noindent\textbf{A:} The authors of this research paper created both the synthetic and human-annotated datasets.
\\
\\
\noindent\textbf{Q:} Who funded the creation of the dataset?
\noindent\textbf{A:} NA.
\\
\\
\noindent\textbf{Q:} Any other comments?
\noindent\textbf{A:} No.

\subsection{Composition}
\noindent\textbf{Q:} What do the instances that comprise the dataset represent (e.g., documents, photos, people, countries)? (Are there multiple types of instances (e.g., movies, users, and ratings; people and interactions between them; nodes and edges)? Please provide a description.)

\noindent\textbf{A:} Each instance in the dataset contains a question, an actual answer, responses generated by an LLM, and a label for each response indicating hallucination (1) or not hallucination (0).
\\
\\
\noindent\textbf{Q:} How many instances are there in total (of each type, if appropriate)?

\noindent\textbf{A:} The synthetic datasets contain 27,406 instances from the Jeopardy dataset and 56,328 instances from the Kaggle dataset. Refer to Tables \ref{tab:halugurad_data} and \ref{tab:kaggle_data_stats} for more information. Meanwhile, the human-annotated test set consists of a total of 19,560 instances, out of which 9,560 are from the Jeopardy dataset and 10,000 are from the Kaggle dataset, for more details refer to Section \ref{sec:Test dataset creation}.
\\
\\
\noindent\textbf{Q:} Does the dataset contain all possible instances or is it a sample (not necessarily random) of instances from a larger set?

\noindent\textbf{A:} The dataset consists of all instances derived from the raw data that we gathered and processed.
\\
\\
\noindent\textbf{Q:} Is any information missing from individual instances? If so, please provide a description, explaining why this information is missing (e.g., because it was unavailable). This does not include intentionally removed information but might include, e.g., redacted text. 

\noindent\textbf{A:} No.
\\
\\
\noindent\textbf{Q:} Are relationships between individual instances made explicit (e.g., users' movie ratings, social network links)? If so, please describe how these relationships are made explicit.)

\noindent\textbf{A:} No
\\
\\
\noindent\textbf{Q:} Are there recommended data splits (e.g., training, development/validation, testing)? If so, please provide a description of these splits, explaining the rationale behind them.

\noindent\textbf{A:} Yes. Refer to Appendix Section~\ref{sec:train_test} for an explanation. The split information is presented in Tables~\ref{tab:halugurad_data} and \ref{tab:kaggle_data_stats}.
\\
\\
\noindent\textbf{Q:} Are there any errors, sources of noise, or redundancies in the dataset? 

\noindent\textbf{A:} We perform rule-based filtration to remove noisy samples present in the dataset; however, it is not feasible to manually inspect all data instances.
\\
\\
\noindent\textbf{Q:} Is the dataset self-contained, or does it link to or otherwise rely on external resources (e.g., websites, tweets, other datasets)? (If it links to or relies on external resources, a) are there guarantees that they will exist, and remain constant, over time; b) are there official archival versions of the complete dataset (i.e., including the external resources as they existed at the time the dataset was created); c) are there any restrictions (e.g., licenses, fees) associated with any of the external resources that might apply to a dataset consumer? Please provide descriptions of all external resources and any restrictions associated with them, as well as links or other access points, as appropriate.)

\noindent\textbf{A:} The dataset is self-contained and can be downloaded, used, adapted, and redistributed without restrictions.
\\
\\
\noindent\textbf{Q:} Does the dataset contain data that might be considered confidential (e.g., data that is protected by legal privilege or by doctor-patient confidentiality, data that includes the content of individuals' non-public communications)? If so, please provide a description.

\noindent\textbf{A:} No, as all samples in the dataset are publicly available.
\\
\\
\noindent\textbf{Q:} Does the dataset contain data that, if viewed directly, might be offensive, insulting, threatening, or might otherwise cause anxiety? If so, please describe why. 

\noindent\textbf{A:} No.
\\
\\
\noindent\textbf{Q:} Does the dataset relate to people? (If not, you may skip the remaining questions in this section.)

\noindent\textbf{A:} No. 
\\
\\
\noindent\textbf{Q:} Does the dataset identify any subpopulations (e.g., by age, gender)? If so, please describe how these subpopulations are identified and provide a description of their respective distributions within the dataset.

\noindent\textbf{A:} No.
\\
\\
\noindent\textbf{Q:} Is it possible to identify individuals (i.e., one or more natural persons), either directly or indirectly (i.e., in combination with other data) from the dataset? If so, please describe how.

\noindent\textbf{A:} No.
\\
\\
\noindent\textbf{Q:} Does the dataset contain data that might be considered sensitive in any way (e.g., data that reveals race or ethnic origins, sexual orientations, religious beliefs, political opinions or union memberships, or locations; financial or health data; biometric or genetic data; forms of government identification, such as social security numbers; criminal history)? If so, please provide a description.

\noindent\textbf{A:} No.
\\
\\
\noindent\textbf{Q:} Any other comments?

\noindent\textbf{A:} No.

\subsection{Collection process}
\noindent\textbf{Q:} How was the data associated with each instance acquired? (Was the data directly observable (e.g., raw text, movie ratings), reported by subjects (e.g., survey responses), or indirectly inferred/derived from other data (e.g., part-of-speech tags, model-based guesses for age or language)? If data was reported by subjects or indirectly inferred/derived from
other data, was the data validated/verified? If so, please describe how.)

\noindent\textbf{A:} The data is obtained from Jeopardy \cite{Jeopardy} and various Kaggle websites.
\\
\\
\noindent\textbf{Q:} What mechanisms or procedures were used to collect the data (e.g., hardware apparatus or sensor, manual human curation, software program, software API)? (How were these mechanisms or procedures validated?)

\noindent\textbf{A:} We manually downloaded the data.
\\
\\
\noindent\textbf{Q:} If the dataset is a sample from a larger set, what was the sampling strategy (e.g., deterministic, probabilistic with specific sampling probabilities)? 

\noindent\textbf{A:}  The dataset is not sampled from a larger corpus.
\\
\\
\noindent\textbf{Q:} Who was involved in the data collection process (e.g., students, crowd workers, contractors) and how were they compensated (e.g., how much were crowd workers paid)?

\noindent\textbf{A:} The dataset was collected from open-source websites, and we will make the processing scripts open-source.
\\
\\
\noindent\textbf{Q:} Over what timeframe was the data collected? (Does this timeframe match the creation timeframe of the data associated with the instances (e.g., a recent crawl of old news articles)? If not, please describe the timeframe in which the data associated with the instances was created.)

\noindent\textbf{A:} The data was collected in late 2024. 
\\
\\
\noindent\textbf{Q:} Were any ethical review processes conducted (e.g., by an institutional review board)? If so, please provide a description of these review processes, including the outcomes, as well as a link or other access point to any supporting documentation.

\noindent\textbf{A:} No.
\\
\\
\noindent\textbf{Q:} Did you collect the data from the individuals in question directly, or obtain it via third parties or other sources (e.g., websites)?

\noindent\textbf{A:} The dataset was obtained by downloading it from open-source websites. See Section~\ref{sec:data_creation} for more details.
\\
\\
\noindent\textbf{Q:} Were the individuals in question notified about the data collection? (If so, please describe (or show with screenshots or other information) how notice was provided, and provide a link or other access point to, or otherwise reproduce, the exact language of the notification itself.)

\noindent\textbf{A:} No. All datasets used to create HaluCounterEval are open source.
\\
\\
\noindent\textbf{Q:} Did the individuals in question consent to the collection and use of their data? (If so, please describe (or show with screenshots or other information) how consent was requested and provided, and provide a link or other access point to, or otherwise reproduce, the exact language to which the individuals consented.)

\noindent\textbf{A:} No. All the datasets present in the HaluCounterEval are open-source.
\\
\\
\noindent\textbf{Q:} If consent was obtained, were the consenting individuals provided with a mechanism to revoke their consent in the future or for certain uses? (If so, please provide a description, as well as a link or other access point to the mechanism (if appropriate).)

\noindent\textbf{A:} N/A.
\\
\\
\noindent\textbf{Q:} Has an analysis of the potential impact of the dataset and its use on data subjects (e.g., a data protection impact analysis) been conducted? (If so, please provide a description of this analysis, including the outcomes, as well as a link or other access point to any supporting documentation.)

\noindent\textbf{A:} No.
\\
\\
\noindent\textbf{Q:} Any other comments?

\noindent\textbf{A:} No.

\subsection{Preprocessing, cleaning, labeling}

\noindent\textbf{Q:} Was any preprocessing/cleaning/labeling of the data done (e.g., discretization or bucketing, tokenization, part-of-speech tagging, SIFT feature extraction, removal of instances, processing of missing values)? (If so, please provide a description. If not, you may skip the remainder of the questions in this section.)

\noindent\textbf{A:} Yes, detailed in Section~\ref{sec:data_creation}.
\\
\\
\noindent\textbf{Q:} Was the ``raw'' data saved in addition to the preprocessed/cleaned/labeled data (e.g., to support unanticipated future uses)? (If so, please provide a link or other access point to the "raw" data.)

\noindent\textbf{A:} The ``raw'' data is saved and we plan to release it shortly.
\\
\\
\noindent\textbf{Q:} Is the software used to preprocess/clean/label the instances available? (If so, please provide a link or other access point.)

\noindent\textbf{A:} Yes, in the GitHub repository footnoted in the main content.
\\
\\
\noindent\textbf{Q:} Any other comments?

\noindent\textbf{A:} No.

\subsection{Uses}

\noindent\textbf{Q:} Has the dataset been used for any tasks already? (If so, please provide a description.)

\noindent\textbf{A:} We have used the dataset for training and testing purposes to perform reference-free hallucination detection. For more details, please refer to Section~\ref{sec:experiment_results}.
\\
\\
\noindent\textbf{Q:} Is there a repository that links to any or all papers or systems that use the dataset? (If so, please provide a link or other access point.)

\noindent\textbf{A:} No.
\\
\\
\noindent\textbf{Q:} What (other) tasks could the dataset be used for?

\noindent\textbf{A:} The dataset can be utilized for a wide range of NLP tasks concerning factual question-answering, and hallucination mitigation.
\\
\\
\noindent\textbf{Q:} Is there anything about the composition of the dataset or the way it was collected and preprocessed/cleaned/labeled that might impact future uses? (For example, is there anything that a future user might need to know to avoid uses that could result in unfair treatment of individuals or groups (e.g., stereotyping, quality of service issues) or other undesirable harms (e.g., financial harms, legal risks) If so, please provide a description. Is there anything a future user could do to mitigate these undesirable harms?)

\noindent\textbf{A:} Yes, we applied rule-based filtration to remove noisy samples from the raw dataset, as detailed in Appendix Section~\ref{sec:jeopardy_filter}.
\\
\\
\noindent\textbf{Q:} Are there tasks for which the dataset should not be used? (If so, please provide a description.)

\noindent\textbf{A:} Our dataset may include misleading responses, as the sample responses are sourced from various large language models (LLMs). Therefore, it should not be used for any purposes that could result in discrimination or harm.  
\\
\\
\noindent\textbf{Q:} Any other comments?

\noindent\textbf{A:} No.

\subsection{Distribution}

\noindent\textbf{Q:} Will the dataset be distributed to third parties outside of the entity (e.g., company, institution, organization) on behalf of which the dataset was created? (If so, please provide a description.)

\noindent\textbf{A:} Yes, the data will be free to the public to download, use, modify, and re-distribute.
\\
\\
\noindent\textbf{Q:} How will the dataset be distributed (e.g., tarball on the website, API, GitHub)? (Does the dataset have a digital object identifier (DOI)?)

\noindent\textbf{A:} The dataset will be hosted in Huggingface.
\\
\\
\noindent\textbf{Q:} When will the dataset be distributed?

\noindent\textbf{A:} The dataset is available now.
\\
\\
\noindent\textbf{Q:} Will the dataset be distributed under a copyright or other intellectual property (IP) license, and/or under applicable terms of use (ToU)? (If so, please describe this license and/or ToU, and provide a link or other access point to, or otherwise reproduce, any relevant licensing terms or ToU, as well as any fees associated with these restrictions.)

\noindent\textbf{A:} Yes, the dataset is distributed under the CC BY 4.0 license.
\\
\\
\noindent\textbf{Q:} Have any third parties imposed IP-based or other restrictions on the data associated with the instances? (If so, please describe these restrictions, and provide a link or other access point to, or otherwise reproduce, any relevant licensing terms, as well as any fees associated with these restrictions.).

\noindent\textbf{A:} The datasets used in this paper are open-sourced, such that there are no restrictions associated with the data.
\\
\\
\noindent\textbf{Q:} Do any export controls or other regulatory restrictions apply to the dataset or individual instances? (If so, please describe these restrictions, and provide a link or other access point to, or otherwise reproduce, any supporting documentation.)

\noindent\textbf{A:} No.
\\
\\
\noindent\textbf{Q:} Any other comments?

\noindent\textbf{A:} No.

\subsection{Maintenance}

\noindent\textbf{Q:} Who is supporting/hosting/maintaining the dataset?

\noindent\textbf{A:} Authors of this paper.
\\
\\
\noindent\textbf{Q:} How can the owner/curator/manager of the dataset be contacted (e.g., email address)?

\noindent\textbf{A:} Via email or issues in the Hugging Face or GitHub repositories.
\\
\\
\noindent\textbf{Q:} Is there an erratum? (If so, please provide a link or other access point.)

\noindent\textbf{A:} No.
\\
\\
\noindent\textbf{Q:} Will the dataset be updated (e.g., to correct labeling errors, add new instances, delete instances)? (If so, please describe how often, by whom, and how updates will be communicated to users (e.g., mailing list, GitHub)?)

\noindent\textbf{A:} Currently there is no plan to update the dataset.
\\
\\
\noindent\textbf{Q:} If the dataset relates to people, are there applicable limits on the retention of the data associated with the instances (e.g., were individuals in question told that their data would be retained for a fixed period of time and then deleted)? (If so, please describe these limits and explain how they will be enforced.)

\noindent\textbf{A:} No.
\\
\\
\noindent\textbf{Q:} Will older versions of the dataset continue to be supported/hosted/maintained? (If so, please describe how. If not, please describe how its obsolescence will be communicated to users.)

\noindent\textbf{A:} There is no older version of the dataset.
\\
\\
\noindent\textbf{Q:} If others want to extend/augment/build on/contribute to the dataset, is there a mechanism for them to do so? (If so, please provide a description. Will these contributions be validated/verified? If so, please describe how. If not, why not? Is there a process for communicating/distributing these contributions to other users? If so, please provide a description.)

\noindent\textbf{A:} Yes, they can freely extend this dataset by downloading it from GitHub.
\\
\\
\noindent\textbf{Q:} Any other comments?

\noindent\textbf{A:} No.

%% file: 0_main.bbl
\begin{thebibliography}{47}
\providecommand{\natexlab}[1]{#1}

\bibitem[{Abdin et~al.(2024)Abdin, Aneja, Awadalla, Awadallah, Awan, Bach, Bahree, Bakhtiari, Bao, Behl et~al.}]{abdin2024phi}
Marah Abdin, Jyoti Aneja, Hany Awadalla, Ahmed Awadallah, Ammar~Ahmad Awan, Nguyen Bach, Amit Bahree, Arash Bakhtiari, Jianmin Bao, Harkirat Behl, et~al. 2024.
\newblock \href {https://arxiv.org/abs/2404.14219v4} {Phi-3 technical report: A highly capable language model locally on your phone}.
\newblock \emph{arXiv preprint arXiv:2404.14219}.

\bibitem[{Achiam et~al.(2023)Achiam, Adler, Agarwal, Ahmad, Akkaya, Aleman, Almeida, Altenschmidt, Altman, Anadkat et~al.}]{achiam2023gpt}
Josh Achiam, Steven Adler, Sandhini Agarwal, Lama Ahmad, Ilge Akkaya, Florencia~Leoni Aleman, Diogo Almeida, Janko Altenschmidt, Sam Altman, Shyamal Anadkat, et~al. 2023.
\newblock \href {https://arxiv.org/abs/2303.08774} {Gpt-4 technical report}.
\newblock \emph{arXiv preprint arXiv:2303.08774}.

\bibitem[{Ahn et~al.(2024)Ahn, Verma, Lou, Liu, Zhang, and Yin}]{ahn-etal-2024-large}
Janice Ahn, Rishu Verma, Renze Lou, Di~Liu, Rui Zhang, and Wenpeng Yin. 2024.
\newblock \href {https://aclanthology.org/2024.eacl-srw.17/} {Large language models for mathematical reasoning: Progresses and challenges}.
\newblock In \emph{Proceedings of the 18th Conference of the European Chapter of the Association for Computational Linguistics: Student Research Workshop}, pages 225--237, St. Julian{'}s, Malta. Association for Computational Linguistics.

\bibitem[{Azaria and Mitchell(2023)}]{azaria-mitchell-2023-internal}
Amos Azaria and Tom Mitchell. 2023.
\newblock \href {https://doi.org/10.18653/v1/2023.findings-emnlp.68} {The internal state of an {LLM} knows when it`s lying}.
\newblock In \emph{Findings of the Association for Computational Linguistics: EMNLP 2023}, pages 967--976, Singapore. Association for Computational Linguistics.

\bibitem[{Chen et~al.(2024)Chen, Liu, Chen, Gu, Wu, Tao, Fu, and Ye}]{cheninside}
Chao Chen, Kai Liu, Ze~Chen, Yi~Gu, Yue Wu, Mingyuan Tao, Zhihang Fu, and Jieping Ye. 2024.
\newblock \href {https://openreview.net/forum?id=Zj12nzlQbz} {Inside: Llms' internal states retain the power of hallucination detection}.
\newblock In \emph{The Twelfth International Conference on Learning Representations}.

\bibitem[{Deutsch et~al.(2022)Deutsch, Dror, and Roth}]{deutsch-etal-2022-limitations}
Daniel Deutsch, Rotem Dror, and Dan Roth. 2022.
\newblock \href {https://doi.org/10.18653/v1/2022.emnlp-main.753} {On the limitations of reference-free evaluations of generated text}.
\newblock In \emph{Proceedings of the 2022 Conference on Empirical Methods in Natural Language Processing}, pages 10960--10977, Abu Dhabi, United Arab Emirates. Association for Computational Linguistics.

\bibitem[{Devlin et~al.(2019)Devlin, Chang, Lee, and Toutanova}]{devlin-etal-2019-bert}
Jacob Devlin, Ming-Wei Chang, Kenton Lee, and Kristina Toutanova. 2019.
\newblock \href {https://doi.org/10.18653/v1/N19-1423} {{BERT}: Pre-training of deep bidirectional transformers for language understanding}.
\newblock In \emph{Proceedings of the 2019 Conference of the North {A}merican Chapter of the Association for Computational Linguistics: Human Language Technologies, Volume 1 (Long and Short Papers)}, pages 4171--4186, Minneapolis, Minnesota. Association for Computational Linguistics.

\bibitem[{Du et~al.(2024)Du, Xiao, and Li}]{du2024haloscope}
Xuefeng Du, Chaowei Xiao, and Sharon Li. 2024.
\newblock \href {https://proceedings.neurips.cc/paper_files/paper/2024/hash/ba92705991cfbbcedc26e27e833ebbae-Abstract-Conference.html} {Haloscope: Harnessing unlabeled llm generations for hallucination detection}.
\newblock \emph{Advances in Neural Information Processing Systems}, 37:102948--102972.

\bibitem[{Dubey et~al.(2024)Dubey, Jauhri, Pandey, Kadian, Al-Dahle, Letman, Mathur, Schelten, Yang, Fan et~al.}]{dubey2024llama}
Abhimanyu Dubey, Abhinav Jauhri, Abhinav Pandey, Abhishek Kadian, Ahmad Al-Dahle, Aiesha Letman, Akhil Mathur, Alan Schelten, Amy Yang, Angela Fan, et~al. 2024.
\newblock \href {https://arxiv.org/abs/2407.21783} {The llama 3 herd of models}.
\newblock \emph{arXiv preprint arXiv:2407.21783}.

\bibitem[{Farquhar et~al.(2024)Farquhar, Kossen, Kuhn, and Gal}]{farquhar2024detecting}
Sebastian Farquhar, Jannik Kossen, Lorenz Kuhn, and Yarin Gal. 2024.
\newblock \href {https://www.nature.com/articles/s41586-024-07421-0} {Detecting hallucinations in large language models using semantic entropy}.
\newblock \emph{Nature}, 630(8017):625--630.

\bibitem[{Fortier-Dubois and Rosati(2023)}]{fortier-dubois-rosati-2023-using}
Etienne Fortier-Dubois and Domenic Rosati. 2023.
\newblock \href {https://doi.org/10.18653/v1/2023.acl-short.72} {Using contradictions improves question answering systems}.
\newblock In \emph{Proceedings of the 61st Annual Meeting of the Association for Computational Linguistics (Volume 2: Short Papers)}, pages 827--840, Toronto, Canada. Association for Computational Linguistics.

\bibitem[{Gebru et~al.(2021)Gebru, Morgenstern, Vecchione, Vaughan, Wallach, III, and Crawford}]{datasheet}
Timnit Gebru, Jamie Morgenstern, Briana Vecchione, Jennifer~Wortman Vaughan, Hanna Wallach, Hal~Daum\'{e} III, and Kate Crawford. 2021.
\newblock \href {https://doi.org/10.1145/3458723} {Datasheets for datasets}.
\newblock \emph{Commun. ACM}, 64(12).

\bibitem[{GK()}]{GK}
GK.
\newblock \url{https://www.kaggle.com/datasets/ilyaryabov/general-knowledge-qa}.
\newblock Online; accessed 1-November-2024.

\bibitem[{He et~al.(2021)He, Liu, Gao, and Chen}]{hedeberta}
Pengcheng He, Xiaodong Liu, Jianfeng Gao, and Weizhu Chen. 2021.
\newblock \href {https://openreview.net/forum?id=XPZIaotutsD} {Deberta: Decoding-enhanced bert with disentangled attention}.
\newblock In \emph{International Conference on Learning Representations}.

\bibitem[{Hu et~al.(2024)Hu, Ru, Qiu, Guo, Zhang, Xu, Luo, Liu, Zhang, and Zhang}]{hu2024refchecker}
Xiangkun Hu, Dongyu Ru, Lin Qiu, Qipeng Guo, Tianhang Zhang, Yang Xu, Yun Luo, Pengfei Liu, Yue Zhang, and Zheng Zhang. 2024.
\newblock \href {https://arxiv.org/abs/2405.14486} {Refchecker: Reference-based fine-grained hallucination checker and benchmark for large language models}.
\newblock \emph{arXiv preprint arXiv:2405.14486}.

\bibitem[{Hurst et~al.(2024)Hurst, Lerer, Goucher, Perelman, Ramesh, Clark, Ostrow, Welihinda, Hayes, Radford et~al.}]{hurst2024gpt}
Aaron Hurst, Adam Lerer, Adam~P Goucher, Adam Perelman, Aditya Ramesh, Aidan Clark, AJ~Ostrow, Akila Welihinda, Alan Hayes, Alec Radford, et~al. 2024.
\newblock \href {https://arxiv.org/abs/2410.21276} {Gpt-4o system card}.
\newblock \emph{arXiv preprint arXiv:2410.21276}.

\bibitem[{Jeopardy()}]{Jeopardy}
Jeopardy.
\newblock \url{https://www.reddit.com/r/datasets/comments/1uyd0t/200000_jeopardy_questions_in_a_json_file/?rdt=35719}.
\newblock Online; accessed 1-December-2024.

\bibitem[{Ji et~al.(2024)Ji, Chen, Ishii, Cahyawijaya, Bang, Wilie, and Fung}]{ji2024llm}
Ziwei Ji, Delong Chen, Etsuko Ishii, Samuel Cahyawijaya, Yejin Bang, Bryan Wilie, and Pascale Fung. 2024.
\newblock \href {https://aclanthology.org/2024.blackboxnlp-1.6/} {Llm internal states reveal hallucination risk faced with a query}.
\newblock In \emph{Proceedings of the 7th BlackboxNLP Workshop: Analyzing and Interpreting Neural Networks for NLP}, pages 88--104.

\bibitem[{Jiang et~al.(2023)Jiang, Sablayrolles, Mensch, Bamford, Chaplot, Casas, Bressand, Lengyel, Lample, Saulnier et~al.}]{jiang2023mistral}
Albert~Q Jiang, Alexandre Sablayrolles, Arthur Mensch, Chris Bamford, Devendra~Singh Chaplot, Diego de~las Casas, Florian Bressand, Gianna Lengyel, Guillaume Lample, Lucile Saulnier, et~al. 2023.
\newblock \href {https://arxiv.org/abs/2310.06825} {Mistral 7b}.
\newblock \emph{arXiv preprint arXiv:2310.06825}.

\bibitem[{Li et~al.(2023)Li, Cheng, Zhao, Nie, and Wen}]{li-etal-2023-halueval}
Junyi Li, Xiaoxue Cheng, Xin Zhao, Jian-Yun Nie, and Ji-Rong Wen. 2023.
\newblock \href {https://doi.org/10.18653/v1/2023.emnlp-main.397} {{H}alu{E}val: A large-scale hallucination evaluation benchmark for large language models}.
\newblock In \emph{Proceedings of the 2023 Conference on Empirical Methods in Natural Language Processing}, pages 6449--6464, Singapore. Association for Computational Linguistics.

\bibitem[{Li et~al.(2024)Li, Geng, Lyu, Zhu, Panov, and Karray}]{li-etal-2024-reference}
Qing Li, Jiahui Geng, Chenyang Lyu, Derui Zhu, Maxim Panov, and Fakhri Karray. 2024.
\newblock \href {https://doi.org/10.18653/v1/2024.findings-emnlp.262} {Reference-free hallucination detection for large vision-language models}.
\newblock In \emph{Findings of the Association for Computational Linguistics: EMNLP 2024}, pages 4542--4551, Miami, Florida, USA. Association for Computational Linguistics.

\bibitem[{Lin et~al.(2022)Lin, Liu, and Shang}]{lin-etal-2022-towards}
Zi~Lin, Jeremiah~Zhe Liu, and Jingbo Shang. 2022.
\newblock \href {https://doi.org/10.18653/v1/2022.findings-acl.328} {Towards collaborative neural-symbolic graph semantic parsing via uncertainty}.
\newblock In \emph{Findings of the Association for Computational Linguistics: ACL 2022}, pages 4160--4173, Dublin, Ireland. Association for Computational Linguistics.

\bibitem[{Liu et~al.(2022)Liu, Zhang, Brockett, Mao, Sui, Chen, and Dolan}]{liu-etal-2022-token}
Tianyu Liu, Yizhe Zhang, Chris Brockett, Yi~Mao, Zhifang Sui, Weizhu Chen, and Bill Dolan. 2022.
\newblock \href {https://doi.org/10.18653/v1/2022.acl-long.464} {A token-level reference-free hallucination detection benchmark for free-form text generation}.
\newblock In \emph{Proceedings of the 60th Annual Meeting of the Association for Computational Linguistics (Volume 1: Long Papers)}, pages 6723--6737, Dublin, Ireland. Association for Computational Linguistics.

\bibitem[{Liu et~al.(2024)Liu, Khalifa, and Wang}]{liu2024litcab}
Xin Liu, Muhammad Khalifa, and Lu~Wang. 2024.
\newblock \href {https://openreview.net/pdf?id=jH67LHVOIO} {Litcab: Lightweight language model calibration over short-and long-form responses}.
\newblock In \emph{The Twelfth International Conference on Learning Representations}.

\bibitem[{Luo et~al.(2024)Luo, Xiao, and Ma}]{luo-etal-2024-zero-resource}
Junyu Luo, Cao Xiao, and Fenglong Ma. 2024.
\newblock \href {https://doi.org/10.18653/v1/2024.findings-emnlp.204} {Zero-resource hallucination prevention for large language models}.
\newblock In \emph{Findings of the Association for Computational Linguistics: EMNLP 2024}, pages 3586--3602, Miami, Florida, USA. Association for Computational Linguistics.

\bibitem[{Malinin and Gales(2021)}]{malininuncertainty}
Andrey Malinin and Mark Gales. 2021.
\newblock \href {https://openreview.net/forum?id=jN5y-zb5Q7m} {Uncertainty estimation in autoregressive structured prediction}.
\newblock In \emph{International Conference on Learning Representations}.

\bibitem[{Manakul et~al.(2023)Manakul, Liusie, and Gales}]{manakul-etal-2023-selfcheckgpt}
Potsawee Manakul, Adian Liusie, and Mark Gales. 2023.
\newblock \href {https://doi.org/10.18653/v1/2023.emnlp-main.557} {{S}elf{C}heck{GPT}: Zero-resource black-box hallucination detection for generative large language models}.
\newblock In \emph{Proceedings of the 2023 Conference on Empirical Methods in Natural Language Processing}, pages 9004--9017, Singapore. Association for Computational Linguistics.

\bibitem[{MathQA()}]{MathQA}
MathQA.
\newblock \url{https://www.kaggle.com/datasets/thedevastator/dataset-for-solving-math-word-problems}.
\newblock Online; accessed 1-November-2024.

\bibitem[{MathQSA()}]{MathQSA}
MathQSA.
\newblock \url{https://www.kaggle.com/datasets/awsaf49/math-qsa-dataset}.
\newblock Online; accessed 1-November-2024.

\bibitem[{Park and Kim(2025)}]{park-kim-2025-llms}
Hancheol Park and Geonmin Kim. 2025.
\newblock \href {https://aclanthology.org/2025.coling-industry.38/} {Where do {LLM}s encode the knowledge to assess the ambiguity?}
\newblock In \emph{Proceedings of the 31st International Conference on Computational Linguistics: Industry Track}, pages 445--452, Abu Dhabi, UAE. Association for Computational Linguistics.

\bibitem[{Ren et~al.()Ren, Luo, Zhao, Krishna, Saleh, Lakshminarayanan, and Liu}]{renout}
Jie Ren, Jiaming Luo, Yao Zhao, Kundan Krishna, Mohammad Saleh, Balaji Lakshminarayanan, and Peter~J Liu.
\newblock \href {https://openreview.net/pdf?id=kJUS5nD0vPB} {Out-of-distribution detection and selective generation for conditional language models}.
\newblock In \emph{The Eleventh International Conference on Learning Representations}.

\bibitem[{Sahoo et~al.(2024{\natexlab{a}})Sahoo, Meharia, Ghosh, Saha, Jain, and Chadha}]{sahoo-etal-2024-comprehensive}
Pranab Sahoo, Prabhash Meharia, Akash Ghosh, Sriparna Saha, Vinija Jain, and Aman Chadha. 2024{\natexlab{a}}.
\newblock \href {https://doi.org/10.18653/v1/2024.findings-emnlp.685} {A comprehensive survey of hallucination in large language, image, video and audio foundation models}.
\newblock In \emph{Findings of the Association for Computational Linguistics: EMNLP 2024}, pages 11709--11724, Miami, Florida, USA. Association for Computational Linguistics.

\bibitem[{Sahoo et~al.(2024{\natexlab{b}})Sahoo, Meharia, Ghosh, Saha, Jain, and Chadha}]{sahoo2024unveiling}
Pranab Sahoo, Prabhash Meharia, Akash Ghosh, Sriparna Saha, Vinija Jain, and Aman Chadha. 2024{\natexlab{b}}.
\newblock \href {https://aclanthology.org/2024.findings-emnlp.685} {A comprehensive survey of hallucination in large language, image, video and audio foundation models}.
\newblock In \emph{Findings of the Association for Computational Linguistics: EMNLP 2024}, pages 11709--11724, Miami, Florida, USA. Association for Computational Linguistics.

\bibitem[{ScientificQA()}]{ScientificQA}
ScientificQA.
\newblock Sciq: A dataset for science question answering.
\newblock \url{https://www.kaggle.com/datasets/thedevastator/sciq-a-dataset-for-science-question-\\answering}.
\newblock Kaggle; accessed 1 November 2024.

\bibitem[{Snyder et~al.(2024)Snyder, Moisescu, and Zafar}]{snyder2024early}
Ben Snyder, Marius Moisescu, and Muhammad~Bilal Zafar. 2024.
\newblock \href {https://dl.acm.org/doi/abs/10.1145/3637528.3671796} {On early detection of hallucinations in factual question answering}.
\newblock In \emph{Proceedings of the 30th ACM SIGKDD Conference on Knowledge Discovery and Data Mining}, pages 2721--2732.

\bibitem[{Srivatsa and Kochmar(2024)}]{srivatsa-kochmar-2024-makes}
Kv~Aditya Srivatsa and Ekaterina Kochmar. 2024.
\newblock \href {https://doi.org/10.18653/v1/2024.findings-naacl.72} {What makes math word problems challenging for {LLM}s?}
\newblock In \emph{Findings of the Association for Computational Linguistics: NAACL 2024}, pages 1138--1148, Mexico City, Mexico. Association for Computational Linguistics.

\bibitem[{Team et~al.(2024)Team, Mesnard, Hardin, Dadashi, Bhupatiraju, Pathak, Sifre, Rivi{\`e}re, Kale, Love et~al.}]{team2024gemma}
Gemma Team, Thomas Mesnard, Cassidy Hardin, Robert Dadashi, Surya Bhupatiraju, Shreya Pathak, Laurent Sifre, Morgane Rivi{\`e}re, Mihir~Sanjay Kale, Juliette Love, et~al. 2024.
\newblock \href {https://arxiv.org/abs/2403.08295} {Gemma: Open models based on gemini research and technology}.
\newblock \emph{arXiv preprint arXiv:2403.08295}.

\bibitem[{Wei et~al.(2024)Wei, Yao, Ton, Guo, Estornell, and Liu}]{wei2024measuring}
Jiaheng Wei, Yuanshun Yao, Jean-Francois Ton, Hongyi Guo, Andrew Estornell, and Yang Liu. 2024.
\newblock \href {https://arxiv.org/abs/2402.10412} {Measuring and reducing llm hallucination without gold-standard answers via expertise-weighting}.
\newblock \emph{arXiv preprint arXiv:2402.10412}.

\bibitem[{Williams et~al.(2018)Williams, Nangia, and Bowman}]{williams-etal-2018-broad}
Adina Williams, Nikita Nangia, and Samuel Bowman. 2018.
\newblock \href {https://doi.org/10.18653/v1/N18-1101} {A broad-coverage challenge corpus for sentence understanding through inference}.
\newblock In \emph{Proceedings of the 2018 Conference of the North {A}merican Chapter of the Association for Computational Linguistics: Human Language Technologies, Volume 1 (Long Papers)}, pages 1112--1122, New Orleans, Louisiana. Association for Computational Linguistics.

\bibitem[{Wu et~al.(2024)Wu, Liu, Wang, Zhang, Wu, Wang, and Tan}]{wu-etal-2024-logical}
Junfei Wu, Qiang Liu, Ding Wang, Jinghao Zhang, Shu Wu, Liang Wang, and Tieniu Tan. 2024.
\newblock \href {https://doi.org/10.18653/v1/2024.findings-acl.414} {Logical closed loop: Uncovering object hallucinations in large vision-language models}.
\newblock In \emph{Findings of the Association for Computational Linguistics: ACL 2024}, pages 6944--6962, Bangkok, Thailand. Association for Computational Linguistics.

\bibitem[{Xiao and Wang(2021)}]{xiao-wang-2021-hallucination}
Yijun Xiao and William~Yang Wang. 2021.
\newblock \href {https://doi.org/10.18653/v1/2021.eacl-main.236} {On hallucination and predictive uncertainty in conditional language generation}.
\newblock In \emph{Proceedings of the 16th Conference of the European Chapter of the Association for Computational Linguistics: Main Volume}, pages 2734--2744, Online. Association for Computational Linguistics.

\bibitem[{Yang et~al.(2024)Yang, Yang, Zhang, Hui, Zheng, Yu, Li, Liu, Huang, Wei et~al.}]{yang2024qwen2}
An~Yang, Baosong Yang, Beichen Zhang, Binyuan Hui, Bo~Zheng, Bowen Yu, Chengyuan Li, Dayiheng Liu, Fei Huang, Haoran Wei, et~al. 2024.
\newblock \href {https://arxiv.org/abs/2412.15115} {Qwen2. 5 technical report}.
\newblock \emph{arXiv preprint arXiv:2412.15115}.

\bibitem[{Yang and Zhao(2024)}]{yang-zhao-2024-llms}
Dongjie Yang and Hai Zhao. 2024.
\newblock \href {https://doi.org/10.18653/v1/2024.findings-emnlp.117} {Are {LLM}s aware that some questions are not open-ended?}
\newblock In \emph{Findings of the Association for Computational Linguistics: EMNLP 2024}, pages 2142--2152, Miami, Florida, USA. Association for Computational Linguistics.

\bibitem[{Yang(2024)}]{yang2024improving}
Zijiang Yang. 2024.
\newblock \href {https://arxiv.org/abs/2412.07108} {Improving the natural language inference robustness to hard dataset by data augmentation and preprocessing}.
\newblock \emph{arXiv preprint arXiv:2412.07108}.

\bibitem[{Yehuda et~al.(2024)Yehuda, Malkiel, Barkan, Weill, Ronen, and Koenigstein}]{yehuda-etal-2024-interrogatellm}
Yakir Yehuda, Itzik Malkiel, Oren Barkan, Jonathan Weill, Royi Ronen, and Noam Koenigstein. 2024.
\newblock \href {https://doi.org/10.18653/v1/2024.acl-long.506} {{I}nterrogate{LLM}: Zero-resource hallucination detection in {LLM}-generated answers}.
\newblock In \emph{Proceedings of the 62nd Annual Meeting of the Association for Computational Linguistics (Volume 1: Long Papers)}, pages 9333--9347, Bangkok, Thailand. Association for Computational Linguistics.

\bibitem[{Zhang et~al.(2023)Zhang, Li, Das, Malin, and Kumar}]{zhang-etal-2023-sac3}
Jiaxin Zhang, Zhuohang Li, Kamalika Das, Bradley Malin, and Sricharan Kumar. 2023.
\newblock \href {https://doi.org/10.18653/v1/2023.findings-emnlp.1032} {{SAC}$^3$: Reliable hallucination detection in black-box language models via semantic-aware cross-check consistency}.
\newblock In \emph{Findings of the Association for Computational Linguistics: EMNLP 2023}, pages 15445--15458, Singapore. Association for Computational Linguistics.

\bibitem[{Zhang et~al.(2024)Zhang, Zeng, Wang, and Lu}]{zhang2024tinyllama}
Peiyuan Zhang, Guangtao Zeng, Tianduo Wang, and Wei Lu. 2024.
\newblock \href {https://arxiv.org/abs/2401.02385} {Tinyllama: An open-source small language model}.
\newblock \emph{arXiv preprint arXiv:2401.02385}.

\end{thebibliography}
